\documentclass[11pt]{article}

\usepackage[preprint]{acl}

\usepackage{times}
\usepackage{latexsym}

\usepackage[T1]{fontenc}
\usepackage{amssymb}   
\usepackage{pifont}    

\usepackage[utf8]{inputenc}

\usepackage{microtype}

\usepackage{inconsolata}

\usepackage{amsmath}
\usepackage{enumitem}
\usepackage{graphicx}
\usepackage[table]{xcolor}
\newcommand{\rankcell}[2]{\cellcolor{green!#2!red!35}#1}

\usepackage{booktabs}
\definecolor{spadered!10}{HTML}{ffc9c9}
\title{SPADE: Soft Prompts for Ai-text DEtection}

\author{Alex Jiang\textsuperscript{1,2}, Didier Schwab\textsuperscript{2}, Chuyuan Li\textsuperscript{2} \\
  \textsuperscript{1}ChapsVision, France \\
  \textsuperscript{2}Univ. Grenoble Alpes, CNRS, Grenoble INP*, LIG, 38000 Grenoble, France \\
  \texttt{ajiang@chapsvision.com}}

\begin{document}
\maketitle
\begin{abstract}
The easy access to Large Language Models (LLMs) has made AI-generated texts on the web increasingly ubiquitous, making AI-text detection a crucial task. Prompts greatly influence the generation process, but existing detectors focus solely on the generated texts --- supervised methods learn from labeled texts only, and zero-shot methods rely on the text's probability distributions --- ignoring the underlying generation process, which limits their robustness in open-world scenarios. We introduce \textbf{SPADE}: Soft Prompts for Ai-text DEtection, a novel framework for detecting AI-generated text that transfers better across datasets. Specifically, we measure the prompt's influence and use it as a discriminative signal. We characterize the prompt's influence by introducing a Pointwise Mutual Information (PMI) term that captures how much a prompt explains a text. Based on this quantity, SPADE trains one soft prompt per class to amplify the separation, and predicts the class whose prompt best explains the candidate text. Extensive experiments on several benchmarks against various detectors demonstrate the robustness and effectiveness of our method, achieving state-of-the-art (SOTA) performance on out-of-distribution (OOD) datasets.
\end{abstract}

\section{Introduction}
Large Language Models (LLMs) such as GPT \citep{openai2024gpt4technicalreport}, LLaMA \citep{grattafiori2024llama3herdmodels}, or Gemini \citep{geminiteam2025geminifamilyhighlycapable} have transformed text generation by producing artificial content that is reasonable, fluent, and contextually relevant. This breakthrough has dramatically lowered technical barriers to produce automatically artificial contents, enabling large-scale automated disinformation and phishing campaigns, and academic dishonesty. The proportion of synthetic content has increased in some media, from 1\% in 2022 to almost 20\% in 2024 \citep{brooks-etal-2024-rise}, and this trend continues. In practice, prior works focus on two possible formulations of the detection task: model attribution, i.e., identifying the source model \citep{pmlr-v267-sun25z}, and binary classification, i.e, deciding whether a text is human-written or AI-generated \citep{solaiman2019releasestrategiessocialimpacts, NEURIPS2019_3e9f0fc9}. 
In this paper, we follow the predominant formulation and treat AI-text detection as a binary classification task.

\begin{figure}[t]
  \includegraphics[width=\columnwidth]{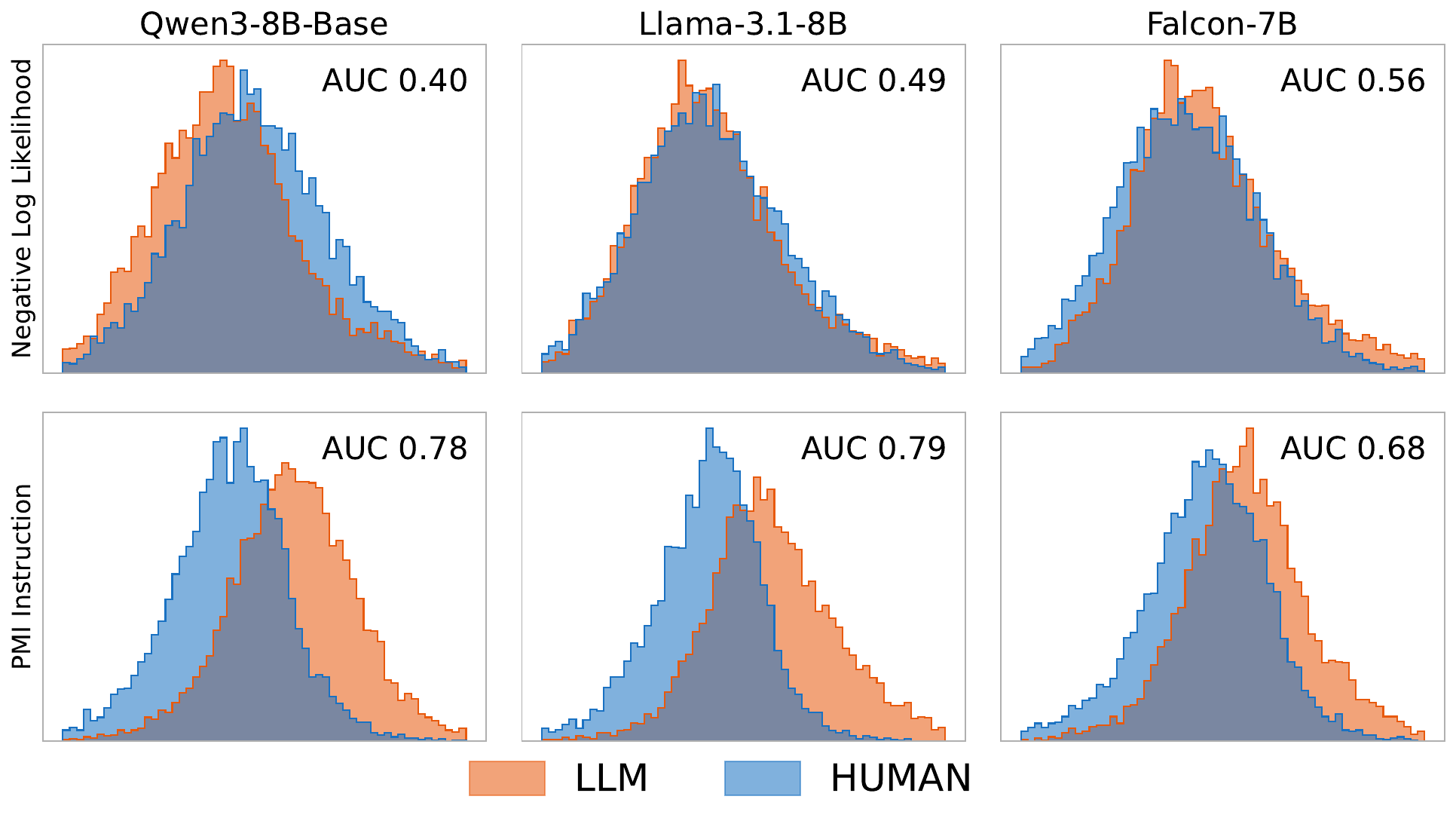}
  \caption{Histograms of the negative log-likelihood (NLL) (above) and $\text{PMI}$ (below) using a generic instruction prompt "Make this text more formal and professional:" across three backbones (Qwen, LLaMA, Falcon) on MIRAGE-DIG \citep{fu2025detectanyllm} test split. \textbf{The prompt's influence PMI is a better discriminative signal than NLL.}}
  \label{fig:motivation}
\end{figure}

LLM generation is inherently conditioned on the input prompt, and different prompts can induce substantially different output distributions and textual characteristics \citep{jiang-etal-2020-know}. In practice, recovering the exact generative context of either class is infeasible, as it remains unknown at test time. As a result, existing detectors often conflate prompt-induced characteristics with those intrinsic to LLM generation, leading to inconsistent performance across prompts \citep{koike-etal-2024-prompt}.

Most classical detectors do not explicitly account for this conditioning variable. Supervised detectors learn from labeled data using pre-trained models or hand-crafted features \citep{guo2023closechatgpthumanexperts, verma-etal-2024-ghostbuster}. 
They perform well in-domain \citep{macko-etal-2025-multisocial}, but are typically trained on AI-generated texts elicited by a limited set of hard prompts, potentially capturing prompt-specific patterns that hinder generalization to unseen prompts.
On the other hand, zero-shot detectors \citep{bao2024fastdetectgpt, pmlr-v235-hans24a} use statistical measures computed from the logits of some scoring models. 
They generalize better but rely on scoring models that closely resemble the source models \citep{pmlr-v306-sun26b}, and overlook the influence of prompts on generation distributions by deriving detection signals solely from the generated text.

Instead, we find that conditioning text likelihood (PMI) on even a \textit{generic instruction} can substantially improve the separation between AI-generated and human-written texts, and this holds across different language models (see Figure~\ref{fig:motivation}).
This suggests that the generative context provides discriminative information beyond text likelihood alone. 
Based on this, we hypothesize that 
each class is characterized by a distinct latent context: \textbf{AI-generated texts are conditioned on instructions, whereas human-written texts reflect a different, implicit writing context.}

In this line, regeneration-based methods---both supervised \citep{ICLR2024_1888b9df, NEURIPS2024_1d35af80, hao-etal-2025-learning} and zero-shot methods \citep{ICLR2024_d4ce6738}---try to bring the conditioning variable to the detection task. They prompt an LLM to regenerate a given text, and then compare the regenerated texts and the given text through features. All of them focus on using a hard prompt, and they use it only to reproduce the text rather than to measure how well it explains the text.

Since the generative context is unknown at test time, and human-written texts have no explicit prompt. We instead address this challenge by learning two class-specific soft prompts that approximate the latent contexts underlying AI-generated and human-written texts, respectively. We term this approach \textbf{SPADE: Soft Prompts for Ai-text DEtection}. We further use the log-likelihood ratio under these two learned contexts as a detection score, measuring which class-specific context better explains a given text.

\begin{figure*}[t]
  \centering
  \includegraphics[width=\textwidth]{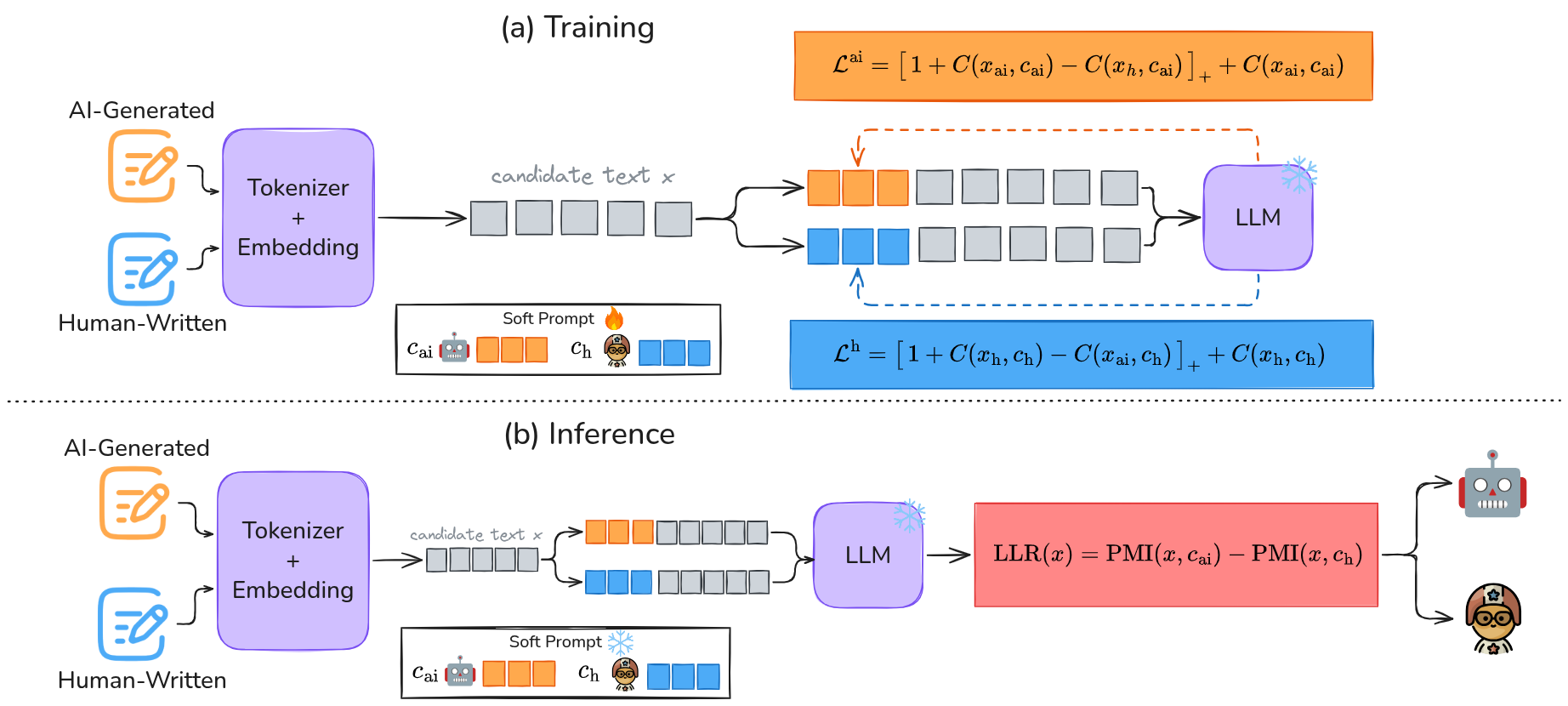}
  \caption{\textbf{Overview of SPADE}: (a) Training. We propose a contrastive loss to learn two discriminative soft prompts to model each class writing process. (b) Inference. We employ a log-likelihood ratio test between the two trained soft prompts for Human $c_h$ and AI $c_\text{ai}$. SPADE classifies by selecting the soft prompt whose $\text{PMI}$ best explains the candidate text. 
}
  \label{fig:overview}
\end{figure*}

Extensive experiments on two benchmarks (EvoBench and RAID) confirm the effectiveness, robustness, and transferability of SPADE. Additionally, we evaluate the robustness of our method across different LLM backbones and text lengths, achieving state-of-the-art (SOTA) performance against several supervised and zero-shot baselines.
In summary, the main contributions are as follows:

\begin{itemize}[noitemsep, topsep=0pt]
    \item We introduce a new formulation of AI-text detection based on learning class-specific soft prompts.
    \item State-of-the-art performance across two benchmarks, particularly at low false positive rates.
    \item Better generalizability in cross-dataset and cross-family while remaining inference-efficient and storage-light, requiring only two learned soft prompts and a frozen backbone. 
\end{itemize}

\section{Related Work}
\subsection{Existing detection methods}
\textbf{Supervised methods} train a classifier to distinguish AI-generated text from Human-written text. Some methods train a BERT-like model on labeled texts \cite{guo2023closechatgpthumanexperts, macko-etal-2025-multisocial}, fine-tune encoders to learn representations that separate the two classes via contrastive learning \cite{ICLR2024_7e54a259, NEURIPS2024_a117a3cd, ta-etal-2026-faid}, or train adversarially to improve robustness against perturbation attacks \cite{NEURIPS2023_30e15e59, li-etal-2025-iron}. Another line of works trains ensemble-based classifiers using linguistic features \cite{kumarage2023stylometricdetectionaigeneratedtext} and statistical features \cite{verma-etal-2024-ghostbuster, basani2026diversityboostsaigeneratedtext}. Although these methods achieve strong performance in-distribution, they often lack robustness against unseen domains and models, and tend to overfit easily \cite{pmlr-v202-mitchell23a}.

\textbf{Zero-shot methods} often use statistical cues to make their decisions. These cues are, for instance, probability-based such as Log-Likelihood \citep{solaiman2019releasestrategiessocialimpacts}, Entropy \citep{gehrmann-etal-2019-gltr}, Negative log-curvature \citep{pmlr-v202-mitchell23a, bao2024fastdetectgpt}, log-likelihood log-rank ratio \cite{su-etal-2023-detectllm}, Cross-Perplexity \citep{pmlr-v235-hans24a}, normalized temperature sensitivity \citep{ma-etal-2026-zero} computed from the logits of one or several scoring models. In a real-world scenario, the LLM that generated a given text is unknown and often comes from black-box model, making the extraction of the true logits impossible. Therefore, zero-shot methods need to use some surrogate models to approximate black-box distributions, intuitively lacking robustness to distributional shift between the source and scoring models. 

Concurrently, EchoPrompt \citep{bao2026responseresponsedetectingllmgenerated} proposed an implicit reward modeling approach \citep{NEURIPS2025_f50258b3, CHEN2026131212}, amplified by the $\text{PMI}$. Unlike EchoPrompt, SPADE explicitly trains two soft prompts and derives its signal entirely from the prompt by comparing the $\text{PMI}$ of the text under two learned soft prompts.

\subsection{Parameter-Efficient Fine-tuning}
\textbf{Soft prompting} offers an efficient alternative to full fine-tuning by adapting frozen LLMs to downstream tasks, such as summarization \citep{liu-etal-2022-psp} and sentiment classification \citep{wu-shi-2022-adversarial}, without requiring manually crafted prompts or updates to the model parameters. 
Unlike conventional soft prompting, which steers generation toward downstream tasks, SPADE learns class-specific soft prompts to enhance the discriminative signal shown in Figure \ref{fig:motivation} for AI-text detection.

\textbf{LoRA} \citep{hu2022lora} is another alternative to full fine-tuning, alongside soft prompts. ImBD \citep{10.1609/aaai.v39i22.34525}  uses LoRA to tune the model for the detection task. However, LoRA modifies the model’s internal behavior, whereas our method keeps the pre-trained model untouched to study how it responds to a prompt.

\section{Method}
\label{sec:methods}
We propose SPADE (Soft Prompts for Ai-text DEtection), a detection framework that learns class-specific soft prompts to distinguish AI-generated from human-written texts. As illustrated in Figure \ref{fig:overview}, we train two soft prompts to learn both contexts and perform a log-likelihood ratio to test which of the two soft prompts best explains a given text.

\subsection{Notation}
\label{sec:notation}
We denote any conditioning information for a text as a context $c$. A context can be a hard prompt, a learned soft prompt, or an implicit conditioning underlying the text's creation (e.g., the writer's background, task, or register). Given a context $c$, a tokenized input text $x$, and a backbone LLM $q$, the pointwise mutual information (PMI) is:
\begin{equation}
    \text{PMI}(x, c) = \log q(x|c) - \log q(x),
    \label{eq:pmi_equation}
\end{equation}
where $C(x, c) = - \log q(x|c)$ is the conditional negative log-likelihood.
The $\text{PMI}$ is computed by subtracting the conditional negative log-likelihood (CNLL) from the negative log-likelihood (NLL) of the input text. It measures how much the context helps the model generate a specific text.

In the following section, we consider two classes $y \in \{\text{ai}, \text{h}\}$ as AI-generated and Human-written. We denote $x_y$ an input text of class $y$, $c_y$ the soft prompt associated to class $y$, and $x_{\bar{y}}$ an input text from the opposite class.

\subsection{Training the Soft Prompt}
\label{sec:training}
A natural loss to train the soft prompt $c_y$ is to minimize $C(x_y, c_y)$ of the texts of class $y$. However, simply minimizing $C$ may increase the likelihood of both classes, leading to a soft prompt that explains both classes similarly. Therefore, we propose a hinge loss on the CNLL to learn a more discriminative signal. 

However, a pure hinge loss is invariant to a global shift since it constrains only the relative gap between the two conditional scores. To ensure that the class-specific prompt also assigns high likelihood to its own class, we also minimize its in-class CNLL. The final loss thus becomes:
\begin{equation}
\resizebox{\columnwidth}{!}{$
    \mathcal{L}_{y} =
    \underbrace{C(x_y, c_y)}_{\text{CNLL}}
    +
    \underbrace{\big[\, m + C(x_y, c_y) - C(x_{\bar{y}}, c_y) \,\big]_+}_{\text{Hinge}} ,
$}
\label{eq:loss}
\end{equation}
where $[z]_+ = \max(0, z)$, $m > 0$ is the margin (we use $m = 1$ in all experiments), and $C$ is the CNLL.

Intuitively, the hinge loss forces the soft prompt $c_y$ to explain $x_y$ better than $x_{\bar{y}}$ by at least a margin $m$. In practice, since the two soft prompts share only a frozen backbone $q$, they can be trained in parallel by batching both prompts through a single forward pass.

\subsection{Detection using a log-likelihood ratio test}
\label{subsec:method-detection}
Given a candidate text $x$ and the two soft prompts  $c_\text{ai}$ and $c_h$, the log-likelihood ratio (LLR) test between the two $\text{PMIs}$ is defined as:
\begin{equation}
    \text{LLR}(x) = \text{PMI}(x, c_\text{ai}) - \text{PMI}(x, c_h).
    \label{eq:llr}
\end{equation}

The ratio aims to test which context best explains the candidate text $x$: we classify $x$ as AI-generated if $\text{LLR}(x) > \tau$, and human-written otherwise.

The detection signal focuses solely on a comparison between $\text{PMI}(x, c_\text{ai})$ and $\text{PMI}(x, c_h)$.

By Equation \ref{eq:pmi_equation}, we can write $\text{LLR}$ as the difference between the two CNLLs:
\begin{equation}
    \text{LLR}(x) = C(x, c_h) - C(x, c_\text{ai}).
    \label{eq:llr_cancel}
\end{equation}

This simplification of the LLR speeds up the inference time, since we only need to compute a single forward with a batch of 2 and can skip the NLL computation.

\section{Experimental Setup}
\label{sec:experimental-setup}
\subsection{Datasets and Evaluation Metrics}
\paragraph{Training dataset.} We train and validate on MIRAGE \citep{fu2025detectanyllm}, containing both human-written texts and AI-generated texts from 9 families of instruct LLMs across 5 domains and 3 prompt types. MIRAGE proposes two datasets: (1)~Disjoint-Input Generation (DIG), each human-written text is paired with a unique AI-generated text from one of the 9 families. (2)~Shared-Input Generation (SIG), each 2,384 human-written texts is paired with one AI-generated text per family. For each of 3 random seeds, we draw a train/validation/test split of MIRAGE-DIG, and one split per family for MIRAGE-SIG.

\begin{table*}[t]
  \centering
  \resizebox{\textwidth}{!}{%
  \begin{tabular}{l cc cc cc cc cc}
    \hline
    & \multicolumn{2}{c}{\textbf{harmful}} & \multicolumn{2}{c}{\textbf{peerread}} & \multicolumn{2}{c}{\textbf{pubmed}} & \multicolumn{2}{c}{\textbf{writing}} & \multicolumn{2}{c}{\textbf{Avg.}} \\
    \cmidrule(lr){2-3} \cmidrule(lr){4-5} \cmidrule(lr){6-7} \cmidrule(lr){8-9} \cmidrule(lr){10-11}
    \textbf{Detector} & AUC & TPR & AUC & TPR & AUC & TPR & AUC & TPR & AUC & TPR \\
    \hline
        \rowcolor{gray!20}
    \multicolumn{11}{c}{\textbf{Zero-Shot Detectors}} \\
    Likelihood & $57.1$ & $8.45$ & $82.6$ & $37.6$ & $90.8$ & $62.1$ & $97.4$ & $89.1$ & $82.0$ & $49.3$ \\
    Entropy & $55.6$ & $6.16$ & $59.1$ & $10.5$ & $90.7$ & $68.3$ & $96.7$ & $86.0$ & $75.5$ & $42.8$ \\
    DNA-GPT \citep{ICLR2024_d4ce6738} & $54.7$ & $5.49$ & $89.0$ & $52.2$ & $87.5$ & $48.7$ & $90.4$ & $59.3$ & $80.4$ & $41.4$ \\
    Fast-DetectGPT \citep{bao2024fastdetectgpt} & $56.0$ & $6.32$ & $86.6$ & $50.0$ & $71.3$ & $22.3$ & $80.7$ & $41.8$ & $73.6$ & $30.1$ \\
    Binoculars \citep{pmlr-v235-hans24a} & $54.4$ & $6.96$ & $86.6$ & $51.5$ & $69.1$ & $24.0$ & $82.8$ & $51.0$ & $73.2$ & $33.4$ \\
    NTS \citep{ma-etal-2026-zero} & $48.9$ & $6.54$ & $73.3$ & $28.0$ & $46.2$ & $1.41$ & $59.8$ & $16.8$ & $57.1$ & $13.2$ \\
    \hline
    \rowcolor{gray!20}
    \multicolumn{11}{c}{\textbf{Supervised Detectors}} \\
    RoBERTa        & $95.3_{1.7}$ & $75.9_{8.2}$ & $89.4_{0.8}$ & $63.5_{4.8}$ & $74.1_{1.3}$ & $27.4_{5.2}$ & $96.1_{1.2}$ & $86.4_{3.9}$ & $88.7_{0.4}$ & $63.3_{2.4}$ \\
    RAIDAR \citep{ICLR2024_1888b9df}          & $76.6_{2.3}$ & $26.4_{3.2}$ & $67.3_{0.6}$ & $20.0_{1.6}$ & $64.5_{3.2}$ & $10.7_{3.8}$ & $81.8_{0.8}$ & $35.5_{1.0}$ & $72.6_{1.2}$ & $23.2_{1.2}$ \\
    DPIC \citep{NEURIPS2024_1d35af80}          & $80.5_{0.7}$ & $46.1_{2.9}$ & $76.1_{1.9}$ & $41.9_{2.9}$ & $66.0_{1.4}$ & $26.2_{1.7}$ & $92.0_{0.1}$ & $72.8_{0.8}$ & $78.6_{0.8}$ & $46.8_{1.8}$ \\
    ImBD \citep{10.1609/aaai.v39i22.34525} & $92.3_{1.9}$ & $63.7_{10.0}$ & $93.2_{1.2}$ & $67.0_{5.2}$ & $87.3_{4.0}$ & $49.4_{7.9}$ & $98.7_{0.5}$ & $94.5_{4.5}$ & $92.9_{1.2}$ & $68.7_{4.0}$ \\
  \rowcolor{spadered!10}  
    \textbf{SPADE (ours)} & $\mathbf{96.7}_{0.4}$ & $\mathbf{88.4}_{1.4}$ & $\mathbf{95.8}_{0.9}$ & $\mathbf{89.1}_{1.8}$ & $\mathbf{98.3}_{0.3}$ & $\mathbf{95.1}_{0.5}$ & $\mathbf{99.8}_{0.0}$ & $\mathbf{98.9}_{0.2}$ & $\mathbf{97.7}_{0.3}$ & $\mathbf{92.9}_{0.7}$ \\
    \hline
  \end{tabular}%
  }
  \caption{AUC and TPR on EvoBench. All supervised baselines are trained on MIRAGE-DIG and the reported number corresponds to mean$_{\text{std}}$ in defined in \S\ref{sec:experimental-setup} over three seeds. \textbf{Bold} indicates the best method per column.}
  \label{tab:evobench_dig}
\end{table*}

\begin{table*}[t]
    \centering
    \resizebox{\textwidth}{!}{%
    \begin{tabular}{l*{18}{c}}
      \hline
      & \multicolumn{2}{c}{\textbf{Abstracts}} & \multicolumn{2}{c}{\textbf{Books}} & \multicolumn{2}{c}{\textbf{News}} & \multicolumn{2}{c}{\textbf{Poetry}} & \multicolumn{2}{c}{\textbf{Recipes}} & \multicolumn{2}{c}{\textbf{Reddit}} & \multicolumn{2}{c}{\textbf{Reviews}} & \multicolumn{2}{c}{\textbf{Wiki}} & \multicolumn{2}{c}{\textbf{Avg.}} \\
      \cmidrule(lr){2-3} \cmidrule(lr){4-5} \cmidrule(lr){6-7} \cmidrule(lr){8-9} \cmidrule(lr){10-11} \cmidrule(lr){12-13} \cmidrule(lr){14-15} \cmidrule(lr){16-17} \cmidrule(lr){18-19}
        \textbf{Detector} & AUC & TPR & AUC & TPR & AUC & TPR & AUC & TPR & AUC & TPR & AUC & TPR & AUC & TPR & AUC & TPR & AUC & TPR\\
      \hline
        \rowcolor{gray!20}
    \multicolumn{19}{c}{\textbf{Zero-Shot Detectors}} \\
    Likelihood & $75.5$ & $10.0$ & $82.9$ & $11.2$ & $75.6$ & $6.67$ & $75.3$ & $4.00$ & $77.1$ & $14.2$ & $78.1$ & $4.70$ & $81.3$ & $9.59$ & $78.6$ & $4.86$ & $78.1$ & $8.16$ \\
    Entropy & $68.6$ & $10.0$ & $80.8$ & $11.8$ & $71.2$ & $6.00$ & $75.3$ & $4.00$ & $77.7$ & $16.3$ & $76.1$ & $4.70$ & $78.6$ & $9.59$ & $75.2$ & $4.86$ & $75.4$ & $8.41$ \\
    DNA-GPT \citep{ICLR2024_d4ce6738} & $71.0$ & $20.7$ & $77.5$ & $24.9$ & $60.1$ & $10.7$ & $68.3$ & $10.7$ & $67.4$ & $14.2$ & $76.2$ & $15.4$ & $77.8$ & $24.0$ & $75.3$ & $14.6$ & $71.7$ & $16.9$ \\
    Fast-DetectGPT \citep{bao2024fastdetectgpt} & $76.6$ & $32.3$ & $66.4$ & $18.9$ & $73.0$ & $18.7$ & $59.4$ & $19.0$ & $49.6$ & $15.6$ & $72.2$ & $25.0$ & $66.5$ & $28.1$ & $72.5$ & $41.7$ & $67.0$ & $24.9$ \\
    Binoculars \citep{pmlr-v235-hans24a} & $80.7$ & $50.7$ & $75.1$ & $40.8$ & $76.1$ & $38.7$ & $64.9$ & $27.3$ & $53.8$ & $24.1$ & $76.7$ & $36.2$ & $72.3$ & $34.9$ & $75.6$ & $43.8$ & $71.9$ & $37.1$ \\
    NTS \citep{ma-etal-2026-zero} & $76.3$ & $33.3$ & $67.8$ & $20.1$ & $69.1$ & $16.0$ & $63.6$ & $15.3$ & $74.4$ & $21.3$ & $72.9$ & $20.1$ & $63.1$ & $10.3$ & $63.9$ & $18.8$ & $68.9$ & $19.4$ \\
    \hline
    \rowcolor{gray!20}
    \multicolumn{19}{c}{\textbf{Supervised Detectors}} \\
    RoBERTa & $87.9$ & $45.1$ & $83.8$ & $46.2$ & $79.4$ & $43.2$ & $89.4$ & $65.6$ & $78.7$ & $36.4$ & $86.7$ & $50.2$ & $90.5$ & $64.2$ & $80.0$ & $41.8$ & $84.6$ & $49.1$ \\
    RAIDAR \citep{ICLR2024_1888b9df} & $59.9$ & $23.3$ & $65.0$ & $27.6$ & $73.6$ & $42.2$ & $64.2$ & $12.4$ & $77.9$ & $22.5$ & $76.6$ & $49.4$ & $71.3$ & $27.9$ & $65.5$ & $24.8$ & $69.3$ & $28.8$ \\
    DPIC \citep{NEURIPS2024_1d35af80} & $86.9$ & $68.9$ & $87.2$ & $66.9$ & $89.6$ & $72.9$ & $90.4$ & $73.6$ & $82.1$ & $49.2$ & $82.4$ & $56.6$ & $93.8$ & $79.2$ & $80.2$ & $49.3$ & $86.6$ & $64.6$ \\
    ImBD \citep{10.1609/aaai.v39i22.34525} & $88.0$ & $15.1$ & $91.4$ & $27.8$ & $85.0$ & $16.4$ & $93.4$ & $58.4$ & $83.7$ & $10.9$ & $89.4$ & $14.8$ & $88.1$ & $25.8$ & $81.4$ & $13.2$ & $87.5$ & $22.8$ \\
  \rowcolor{spadered!10}
    \textbf{SPADE (ours)} & $\mathbf{98.3}$ & $\mathbf{91.1}$ & $\mathbf{96.7}$ & $\mathbf{84.4}$ & $\mathbf{90.2}$ & $\mathbf{77.6}$ & $\mathbf{97.2}$ & $\mathbf{92.1}$ & $\mathbf{94.9}$ & $\mathbf{79.4}$ & $\mathbf{97.9}$ & $\mathbf{85.7}$ & $\mathbf{95.8}$ & $\mathbf{87.0}$ & $\mathbf{94.6}$ & $\mathbf{74.4}$ & $\mathbf{95.7}$ & $\mathbf{84.0}$ \\
    \hline
    \end{tabular}%
    }
    \caption{AUC and TPR on RAID. All supervised baselines are trained on MIRAGE-DIG and the reported number corresponds to the mean defined in \S\ref{sec:experimental-setup} over three seeds. \textbf{Bold} indicates the best method per column.}
    \label{tab:raid_dig}
\end{table*}

\paragraph{Evaluation dataset.} We evaluate on two datasets completely out-of-distribution (OOD) to serve as our primary benchmarks:
(1)~RAID \citep{dugan-etal-2024-raid}, comprising 8 domains, 11 models, and various text perturbations. We restrict RAID to 7 instruction-tuned model, since they are the most widely used, and use the subset sampled by \citet{ma-etal-2026-zero}; 
(2)~EvoBench \citep{yu-etal-2025-evobench}, comprising 5 domains and 6 model families; to keep the evaluation fully OOD, we remove XSum, which is also part of MIRAGE.
Details for each dataset are given in Appendix \ref{appendix:data}.

\paragraph{Metrics.} Following prior work, we use AUROC (AUC) \citep{pmlr-v202-mitchell23a, bao2024fastdetectgpt} and the true positive rate at a fixed false positive rate of 5\% (TPR) \citep{pmlr-v235-hans24a} as our main evaluation metrics. In addition, we report the macro F1-score for all experiments in the Appendix \ref{appendix:main-result}. On EvoBench, we compute each metric separately per domain on every generator family to avoid any imbalance and report the mean over families; On RAID, we average over all generators within each domain.

\subsection{Baselines}
\paragraph{(I) Zero-shot detectors} score a text with a scoring model:
(1)~Likelihood and Entropy, the mean token log-likelihood and entropy;
(2)~DNA-GPT \citep{ICLR2024_d4ce6738}, the log-likelihood gap between the original continuation of a truncated text and $K$ regenerated ones;
(3)~Fast-DetectGPT \citep{bao2024fastdetectgpt}, the conditional probability curvature;
(4)~Binoculars \citep{pmlr-v235-hans24a}, the perplexity normalized by the cross-perplexity between two models;
(5)~NTS \citep{ma-etal-2026-zero}, the normalized sensitivity of the token distribution to temperature.

\paragraph{(II) Supervised detectors} are trained on labeled data:
(1)~RoBERTa, a fine-tuned encoder classifier;
(2)~RAIDAR \citep{ICLR2024_1888b9df}, a multi-layer perceptron (MLP) on n-gram and edit-distance features between the text and its LLM rewrites;
(3)~DPIC \citep{NEURIPS2024_1d35af80}, a MLP on the embeddings of the text and of a response regenerated from its reconstructed prompt;
(4)~ImBD \citep{10.1609/aaai.v39i22.34525}, Fast-DetectGPT score under a fine-tuned scoring model with LoRA to prefer AI-generated style.

\subsection{Implementation} 
Our experiments use Qwen3-8B \citep{yang2025qwen3technicalreport} as the main backbone, and we further test across different backbones (LLaMA-3.1-8B \citep{grattafiori2024llama3herdmodels} and Falcon-7B \cite{almazrouei2023falconseriesopenlanguage}) in \S\ref{sec:robustness}. Each experiment is repeated over three seeds, and we report the mean and standard deviation across them. To ensure a fair comparison, all supervised methods, including SPADE, use the same three splits for training and validation. We details in Appendix \ref{appendix:implementation} on the hyperparameter choice, initialization, selection metrics, and implementations of all baselines.

\begin{figure*}[t]
  \centering
  \includegraphics[width=\textwidth]{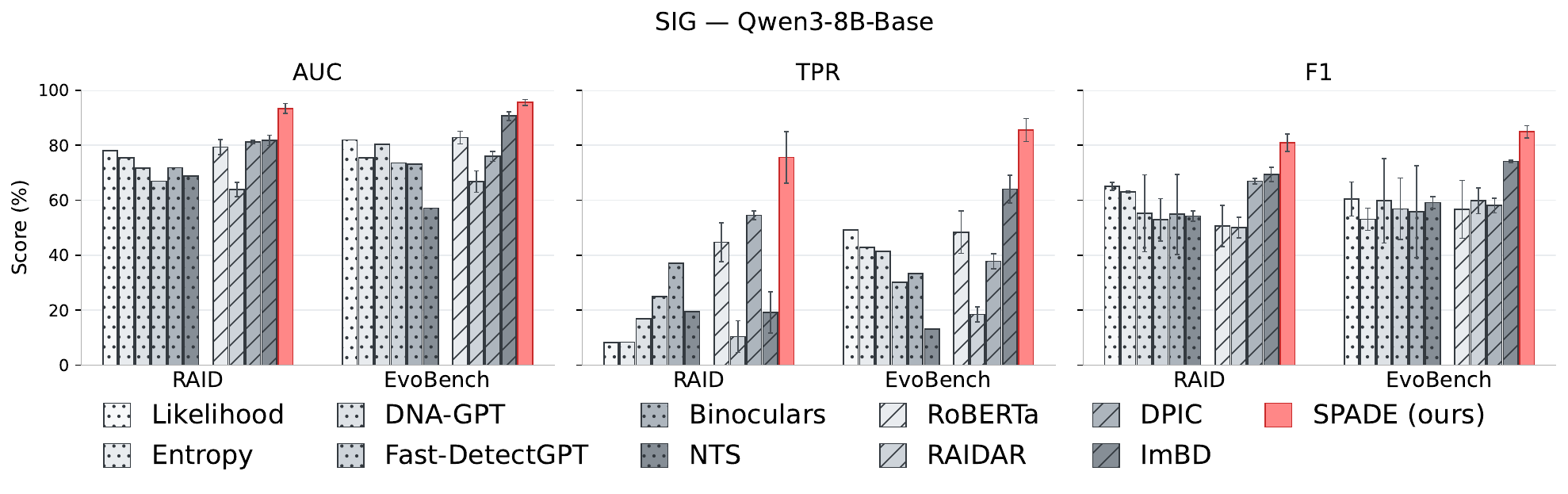}
  \caption{AUC, TPR, and F1 on EvoBench and RAID. All supervised baselines are trained on per family splits from MIRAGE-SIG (over three seeds). The reported number corresponds to mean$_{\text{std}}$ over the 9 families of the mean defined in \S\ref{sec:experimental-setup} over three seeds per family using Qwen3-8B. Hatched bars are supervised baselines and Dotted bars are for zero-shot baselines.}
  \label{fig:figure_sig_qwen}
\end{figure*}

\section{Experiments and Analysis}
\label{sec:experiments}

\subsection{Main Results} 
\label{sec:main-result}
\paragraph{Cross-Dataset Generalization.} Table \ref{tab:evobench_dig} and \ref{tab:raid_dig} present the overall performance of SPADE with other baselines on EvoBench and RAID, respectively. Crucially, all baselines struggle in low-FPR regime: zero-shot detectors have TPR lower than $50\%$, supervised detectors below $70\%$. Overall, supervised detectors perform better than zero-shot ones, with an average gain of $\sim9.7$ AUC on EvoBench and RAID, especially in hard domain where the texts are short (e.g., \textit{harmful} in EvoBench, see \S\ref{sec:text_length}).
Supervised detectors trained on the same setup as SPADE transfer only partially to cross-dataset (e.g., ImBD reaches $68.7$ TPR on EvoBench but only $22.8$ on RAID). In contrast, SPADE outperforms all supervised methods by $4.8$--$25.1$ AUC on EvoBench and $8.2$--$26.4$ AUC on RAID, and maintains a strong TPR score ($84$--$92.9\%$), highlighting its better transferability in cross-dataset. We find consistent results using LLaMA where SPADE outperforms other baselines (details in \S\ref{sec:robustness} per backbone). With Falcon, SPADE is competitive on RAID and wins on EvoBench.



\paragraph{Cross-Generator Generalization.}
We train SPADE on texts from one LLM family in a MIRAGE-SIG and testing it on texts generated by all family models in EvoBench and RAID.

As shown in Figure~\ref{fig:figure_sig_qwen}, SPADE achieves the best AUC, TPR, and F1 on both benchmarks, outperforming all baselines. Specifically, SPADE outperforms the supervised baselines by $4.9$--$28.8$ AUC, $21.5$--$67.1$ TPR, and $10.7$--$26.8$ F1 on EvoBench, and by $11.6$--$29.4$ AUC, $21.1$--$65.2$ TPR, and $11.5$--$30.9$ F1 on RAID (see Appendix \ref{appendix:generalization-using-one-source-model} for the exact LLM-family wise scores and mean scores). This shows that SPADE transfers better than the other supervised baselines across datasets even when trained on a single family with only $500$ text pairs (see Appendix \ref{appendix:mirage}).

\subsection{Backbone and Perturbation Robustness}
\label{sec:robustness}
\begin{figure}[t]
    \centering
    \includegraphics[width=\columnwidth]{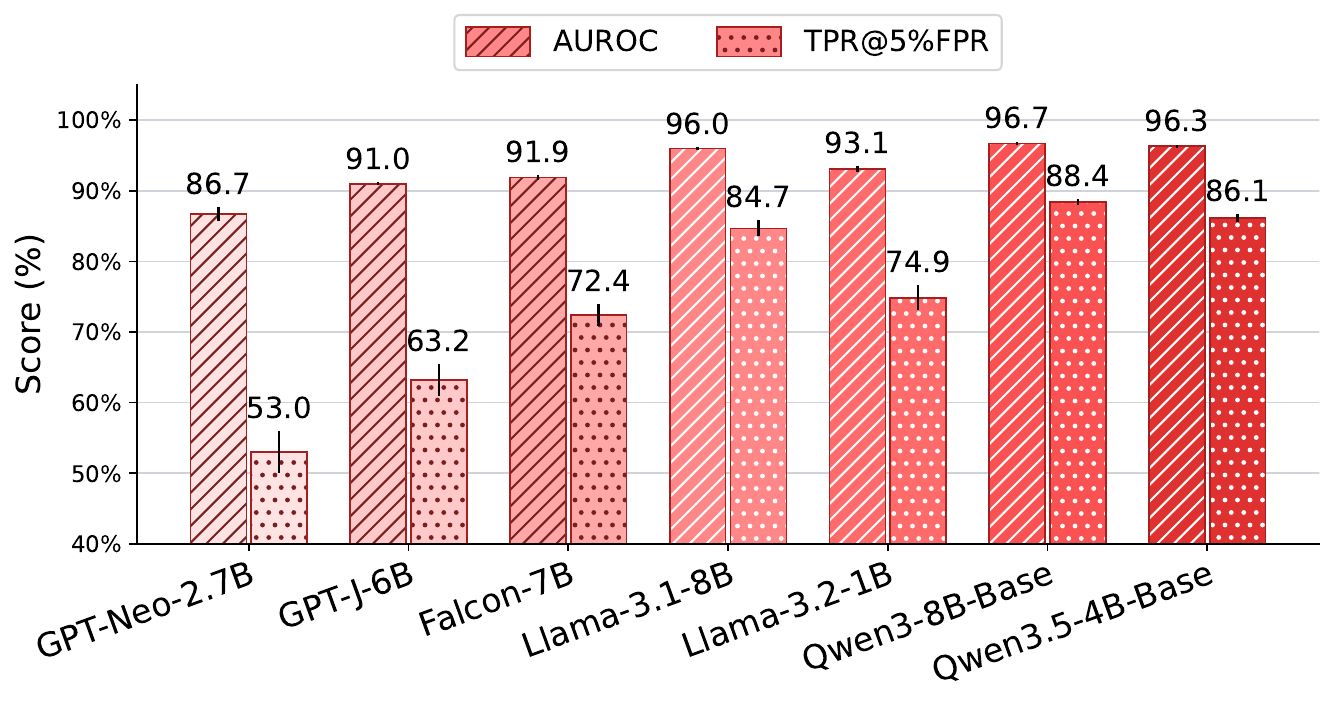}
    \caption{Impact of the backbone on SPADE's performance, averaged over EvoBench and RAID.}
    \label{fig:spade_on_other_backbone}
\end{figure}

\paragraph{SPADE's sensitivity to backbone choice.} Figure \ref{fig:spade_on_other_backbone} presents the mean performance of SPADE on the two benchmarks with various LLM backbones. We observe substantial variation across backbones, with more recent model families generally yielding stronger performance. Interestingly, model scale alone does not explain the differences: LLaMA-3.2-1B \citeyearpar{grattafiori2024llama3herdmodels} outperforms ($+2.5$--$11.7$ TPR) older but larger models (e.g., Falcon-7B \citeyearpar{almazrouei2023falconseriesopenlanguage} and GPT-J-6B \citeyearpar{gpt-j}), and Qwen3.5-4B \citeyearpar{qwenteam2026qwen35omnitechnicalreport} on par with Qwen3-8B \citeyearpar{yang2025qwen3technicalreport}. This suggests that the quality of the backbone's pre-training is the main driver of SPADE's effectiveness. Overall, we expect SPADE to benefit from future advances in backbone quality without requiring any changes to the method itself.

\paragraph{SPADE vs. others across backbones.} The sensitivity to backbone choice is not unique to SPADE, but common to all detectors. Table \ref{tab:main_other_backbone} compares SPADE to other baselines using two additional backbones (LLaMA-3.1-8B and Falcon-7B). A very surprising finding is that complex zero-shot methods (e.g., Fast-DetectGPT, Binoculars) perform substantially better with these two older backbones than with Qwen3-8B. We hypothesize that their degraded performance with Qwen3-8B may stem from differences in model architecture and training, as Qwen3-8B is a more recent hybrid reasoning and instruction-tuned model, potentially altering the likelihood patterns on which these zero-shot methods rely. Regeneration-based methods (e.g, DNA-GPT, RAIDAR, DPIC) remain relatively stable across backbones except for RAIDAR on Falcon-7B-Instruct, where it collapses close to random. With LLaMA-3.1-8B, SPADE remains the best method overall in the main result (see \S\ref{sec:main-result}). With Falcon-7B, Binoculars and NTS are more competitive where SPADE remains the best in AUC ($+0.4$) but lose in TPR ($-0.6$).

\begin{table}[t]
      \centering
      \resizebox{\columnwidth}{!}{%
      \small
      \begin{tabular}{l cc cc cc}
        \hline
        & \multicolumn{2}{c}{\textbf{EvoBench}} & \multicolumn{2}{c}{\textbf{RAID}} & \multicolumn{2}{c}{\textbf{Avg.}} \\
        \cmidrule(lr){2-3} \cmidrule(lr){4-5} \cmidrule(lr){6-7}
        \textbf{Detector} & AUC & TPR & AUC & TPR & AUC & TPR \\
    \hline
        \rowcolor{gray!20}
      \multicolumn{7}{c}{\textbf{LLaMA-3.1-8B}} \\
          Likelihood & $75.4$ & $36.9$ & $74.9$ & $7.5$ & $75.1$ & $22.2$ \\
          Entropy & $63.4$ & $26.7$ & $70.9$ & $6.8$ & $67.2$ & $16.8$ \\
          DNA-GPT & $75.2$ & $32.2$ & $70.1$ & $20.4$ & $72.7$ & $26.3$ \\
          Fast-DetectGPT & $86.4$ & $55.6$ & $86.9$ & $62.1$ & $86.6$ & $58.9$ \\
          Binoculars & $86.0$ & $56.9$ & $89.5$ & $70.1$ & $87.7$ & $63.5$ \\
          NTS & $60.2$ & $20.2$ & $75.3$ & $32.1$ & $67.7$ & $26.2$ \\
        \hline
          RAIDAR & $64.9_{0.5}$ & $18.8_{1.9}$ & $73.1_{0.6}$ & $37.6_{1.3}$ & $69.0$ & $28.2$ \\
          DPIC & $77.9_{0.8}$ & $45.3_{1.4}$ & $86.7_{0.4}$ & $65.3_{1.1}$ & $82.3$ & $55.3$ \\
          ImBD & $88.6_{1.0}$ & $55.9_{2.5}$ & $93.7_{2.6}$ & $67.4_{21.8}$ & $91.1$ & $61.6$ \\
          \rowcolor{spadered!10}
          \textbf{SPADE (ours)} & $\mathbf{97.4}_{0.0}$ & $\mathbf{89.5}_{0.3}$ & $\mathbf{94.5}_{0.4}$ & $\mathbf{79.9}_{2.6}$ & $\mathbf{96.0}$ & $\mathbf{84.7}$ \\
        \hline
          \rowcolor{gray!20}
          \multicolumn{7}{c}{\textbf{Falcon-7B}} \\
          Likelihood & $72.0$ & $29.1$ & $75.1$ & $7.9$ & $73.6$ & $18.5$ \\
          Entropy & $54.3$ & $9.8$ & $64.8$ & $4.4$ & $59.5$ & $7.1$ \\
          DNA-GPT & $74.0$ & $35.1$ & $79.2$ & $36.0$ & $76.6$ & $35.6$ \\
          Fast-DetectGPT & $82.1$ & $51.3$ & $85.2$ & $38.0$ & $83.7$ & $44.7$ \\
          Binoculars & $82.3$ & $52.3$ & $89.3$ & $\mathbf{67.5}$ & $85.8$ & $59.9$ \\
          NTS & $71.3$ & $29.3$ & $88.9$ & $65.2$ & $80.1$ & $47.2$ \\
        \hline
          RAIDAR & $50.0_{0.7}$ & $8.3_{0.7}$ & $54.7_{1.2}$ & $12.7_{0.2}$ & $52.4$ & $10.5$ \\
          DPIC & $78.3_{0.4}$ & $45.0_{1.1}$ & $86.9_{0.6}$ & $63.4_{1.2}$ & $82.6$ & $54.2$ \\
          ImBD & $86.9_{2.2}$ & $52.2_{5.4}$ & $87.4_{3.0}$ & $40.5_{15.6}$ & $87.1$ & $46.3$ \\
          \rowcolor{spadered!10}
          \textbf{SPADE (ours)} & $\mathbf{94.1}_{0.9}$ & $\mathbf{78.0}_{1.2}$ & $\mathbf{89.7}_{0.2}$ & $66.9_{2.0}$ & $\mathbf{91.9}$ & $\mathbf{72.4}$ \\
        \hline
      \end{tabular}%
      }
      \caption{Performance of SPADE and the baselines with LLaMA-3.1-8B and Falcon-7B. \textbf{Bold} indicates the best method within each backbone.}
      \label{tab:main_other_backbone}
  \end{table}

\begin{figure*}[t]
  \centering
  \includegraphics[width=\textwidth]{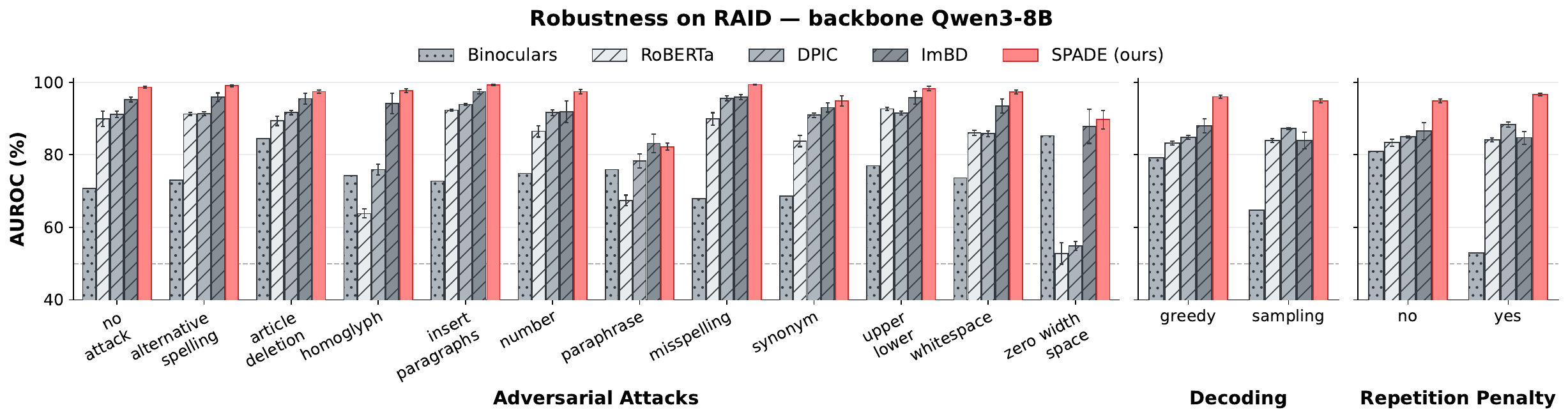}
  \caption{AUC of SPADE and four baselines on RAID, grouped by adversarial attacks. TPR results are consistent and can be found in Appendix \ref{appendix:robustness}. Hatched bars are supervised baselines and Dotted bar is for zero-shot baseline.}
  \label{fig:figure_attacks_qwen_auroc}
\end{figure*}

\paragraph{Perturbation attacks.}
\label{sec:attacks}
Existing detectors are known to be vulnerable to various text perturbations, which can substantially degrade their detection performance \citep{NEURIPS2023_575c4500, sadasivan2025aigeneratedtextreliablydetected}. We investigate whether SPADE remains robust when AI-generated texts are deliberately perturbed to evade detection. Figure~\ref{fig:figure_attacks_qwen_auroc} presents the AUC and TPR of SPADE and the four strongest baselines on RAID under several adversarial attacks. With Qwen3-8B-Base, SPADE achieves the best performance on every attack except paraphrasing ($82.2$ vs. $83.2$ AUC for ImBD; $40.6$ vs. $42.2$ TPR for DPIC). Over the 11 attacks, SPADE significantly outperforms every supervised baseline, including the strongest one, ImBD ($95.6$ vs. $93.1$ AUC, $p=0.005$; $84.4$ vs. $70.3$ TPR, $p=0.008$; Student's $t$-test over the three DIG seeds). With LLaMA-3.1-8B, SPADE is competitive with the strongest detector Binoculars and outperforms it on 5 attacks (e.g., alternative spelling, paragraph insertion, misspelling, synonym substitution and whitespace), but collapses on zero-width space attacks (which hurt all supervised methods) going from $92.1$ TPR without attack to $22.7$ (details per attack type is presented in Appendix \ref{appendix:robustness}).

\subsection{Ablation Study}
\label{sec:ablation_study}
We conduct ablation studies to independently assess the contributions of the hinge loss and log-likelihood ratio test to SPADE's performance. Results are given in Table 4.

\paragraph{Impact of the Hinge Loss.} We compare the regularized hinge loss to a simple cross-entropy loss. Without the contrastive term, the discriminative power of each test statistic drops on both benchmarks: $\text{PMI}(x, c_\text{ai})$'s AUC falls by $6.9$--$36.5$ points and its TPR by $13.4$--$22.2$ points, and $\text{PMI}(x, c_h)$'s AUC falls by $12.3$--$16.3$ points and its TPR by $35.7$--$45.9$ points. Soft prompts learned with a simple cross-entropy perform worse likely because they converge toward a generative optimum rather than a discriminative optimum. Overall, on the LLR, AUC and TPR improves by $2.0$--$16.0$ and $7.7$--$37.4$.

\paragraph{Impact of the Log-likelihood Ratio Test.} We compare the log-likelihood ratio (LLR) test, which contrasts the two learned contexts, with scoring a text using a single soft prompt, $\text{PMI}(x, c_\text{ai})$ or $\text{PMI}(x, c_h)$.  The LLR improves over the best single prompt by $7.0$ AUC and $11.4$ TPR points on EvoBench, and by $5.1$ AUC and $4.2$ TPR points on RAID. These results validate learning two soft prompts rather than one: modeling both classes and contrasting them yields a more accurate and more robust detector.
\begin{table}[t]
    \centering
    \small
    \setlength{\tabcolsep}{4pt}
    \resizebox{\columnwidth}{!}{%
    \begin{tabular}{cc cc cc cc}
    \hline
    \textbf{Hinge} & \textbf{Score} & \multicolumn{2}{c}{\textbf{EvoBench}} & \multicolumn{2}{c}{\textbf{RAID}} & \multicolumn{2}{c}{\textbf{Avg.}} \\
    \cmidrule(lr){3-4} \cmidrule(lr){5-6} \cmidrule(lr){7-8}
     & & AUC & TPR & AUC & TPR & AUC & TPR \\
    \hline
    \rowcolor{spadered!10}
    \checkmark & LLR (ours)    & \textbf{97.7} & \textbf{92.9} & \textbf{95.7} & \textbf{84.0} & \textbf{96.7} & \textbf{88.5} \\
    \checkmark & $\text{PMI}(x, c_\text{h})$  & 90.7 & 81.5 & 90.5 & 79.8 & 90.6 & 80.7 \\
    \checkmark & $\text{PMI}(x, c_\text{ai})$ & 86.3 & 57.9 & 90.6 & 35.4 & 88.5 & 46.7 \\
    \hline
    \ding{55}  & LLR    & 95.7 & 85.2 & 79.7 & 46.6 & 87.7 & 65.9 \\
    \ding{55}  & $\text{PMI}(x, c_\text{h})$  & 78.4 & 45.8 & 74.2 & 33.9 & 76.3 & 39.9 \\
    \ding{55}  & $\text{PMI}(x, c_\text{ai})$ & 79.4 & 44.5 & 54.1 & 13.2 & 66.8 & 28.9 \\
    \hline
    \end{tabular}%
    }
    \caption{Impact of the hinge loss and the LLR in SPADE's performance using Qwen3-8B. Hinge: regularized hinge loss (\checkmark) or plain cross-entropy loss (\ding{55}). We report the average of the two benchmarks.}
    \label{tab:ablation_loss_llr}
\end{table}

\begin{figure}[t]
    \centering
    \includegraphics[width=\columnwidth]{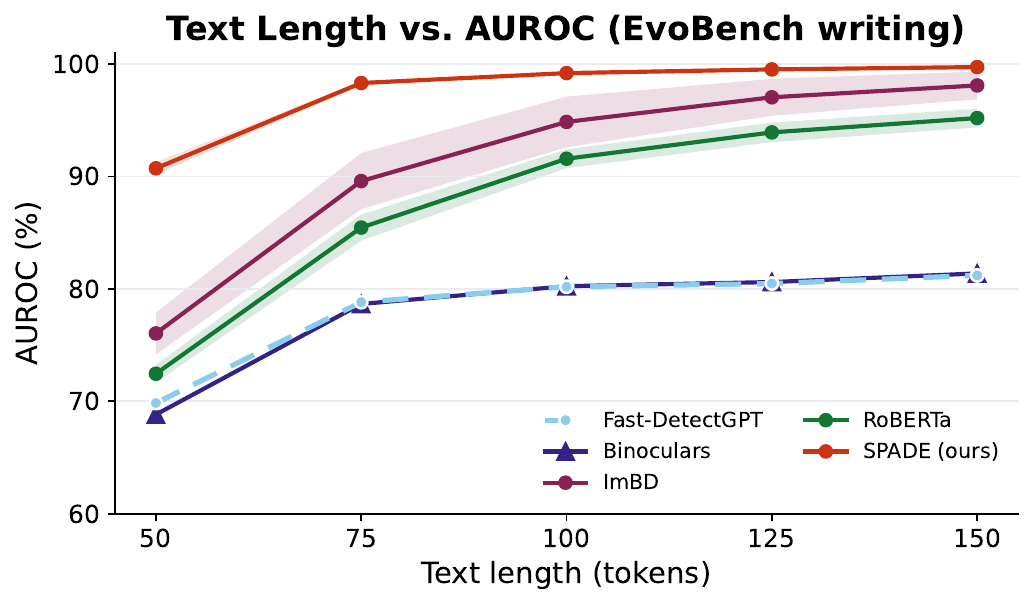}
    \caption{Impact of the text length on four existing baselines and SPADE on EvoBench (Writing) with Qwen3.}
    \label{fig:text_length}
\end{figure}

\subsection{Further Investigations}
\label{sec:further_investigations}

\paragraph{Impact of Text Length}
\label{sec:text_length}
Existing detectors performance are known to drop significantly when dealing with short text \citep{NEURIPS2024_b61bdf7e}. We test SPADE's robustness by truncating the Writing texts in EvoBench to specific lengths. Figure \ref{fig:text_length} presents the AUC of SPADE and four different baselines. Interestingly, supervised methods are more robust than zero-shot methods on short text. Zero-shot methods compute a probability-based score that is not optimized for detection and therefore require a substantial longer text to accumulate a reliable score. In contrast, SPADE is optimized to maximize the gap between the two classes, which makes it more robust than the baselines: it reaches $98.2$ AUC at 75 tokens, versus $88.6$ for ImBD and $78.7$ for Binoculars and Fast-DetectGPT.

\paragraph{Inference Speed}
\label{sec:inference-speed}
Existing regeneration-based detectors require regenerating one or several versions of the input text autoregressively limiting their inference speed \citep{hao-etal-2025-learning}.
For practical purposes, we compare SPADE's inference speed with the baselines in Table~\ref{tab:inference_time}. Our method requires a single forward pass on a batch of two, so it is slower than detectors requiring one forward pass on a single sequence, and faster than Binoculars, which requires two models. Notably, SPADE is much faster than regeneration-based methods ($\sim85.1$ $\times$ speed-up) while transferring better across datasets.

\begin{table}[t]
    \centering
    \resizebox{\columnwidth}{!}{%
    \begin{tabular}{l|c|c|c|c}
        \hline
        Detector & s / doc & s / 100 words & vs.\ SPADE & Rank \\
        \hline
        RoBERTa               & $0.008$  & $0.006_{0.010}$ & $0.06\times$ & \rankcell{2}{90} \\
        Likelihood            & $0.056$  & $0.038_{0.006}$ & $0.39\times$ & \rankcell{5}{60} \\
        Entropy               & $0.058$  & $0.039_{0.006}$ & $0.40\times$ & \rankcell{8}{30} \\
        NTS                   & $0.061$  & $0.042_{0.007}$ & $0.43\times$ & \rankcell{11}{0} \\
        ImBD                  & $0.066$  & $0.046_{0.008}$ & $0.47\times$ & \rankcell{4}{70} \\
        Fast-DetectGPT        & $0.118$  & $0.081_{0.013}$ & $0.83\times$ & \rankcell{9}{20} \\
        \textbf{SPADE (ours)} & $0.141$  & $0.098_{0.014}$ & $1\times$    & \rankcell{\textbf{1}}{100} \\
        Binoculars            & $0.262$  & $0.182_{0.028}$ & $1.87\times$ & \rankcell{6}{50} \\
        DNA-GPT               & $10.612$ & $7.472_{1.592}$ & $76.6\times$ & \rankcell{7}{40} \\
        DPIC                  & $11.416$ & $8.028_{5.318}$ & $82.3\times$ & \rankcell{3}{80} \\
        RAIDAR                & $13.647$ & $9.397_{1.999}$ & $96.4\times$ & \rankcell{10}{10} \\
        \hline
    \end{tabular}%
    }
    \caption{Inference latency per document, with Qwen3-8B on NVIDIA RTX A6000. Times are averaged over 180 MIRAGE-DIG test documents with 20 documents used for warm up. s / 100 words: mean$_{\text{std}}$ of the per-document time normalized by length. Ranked by overall AUC and TPR on EvoBench and RAID.}
    \label{tab:inference_time}
\end{table}

\paragraph{Cross-Lingual}
\label{sec:cross-lingual}
Previous research \citep{wang-etal-2024-m4} propose to assess detector performance on languages other than English. To assess our approach in cross-lingual scenario, we follow DetectRL-X (data details in Appendix \ref{appendix:data}) protocol \citep{wu-etal-2026-detectrl}, by training and tuning on a single language and testing on others (details in Appendix \ref{appendix:cross-lingual}). Encouragingly, SPADE shows no collapse on high-complexity languages defined by DetectRL-X (e.g., Arabic, Russian, Chinese) where it reaches $94.97$ $F_1^B$ on average, against $95.56$ for medium-complexity ones (e.g., German, French, Spanish, Portuguese). Similarly, transferring from Indo-European to Semitic or Sino-Tibetan languages costs SPADE nothing ($95.68$ in-language to $95.54$ out-language $F_1^B$), whereas \citet{wu-etal-2026-detectrl} report an average loss of $19.15$ points. 
However, SPADE shows its largest performance drop when transferring from English to German ($64.45$ $F_1^B$), despite performing well in the reverse direction ($99.51$ $F_1^B$). We leave further investigation of this asymmetry to future work.

\section{Conclusion}
We presented SPADE, a novel detection framework that learns class-specific soft prompts to distinguish AI-generated from human-written texts. Detection is a log-likelihood ratio test between the $\text{PMIs}$ conditioned by the soft prompts. SPADE achieves strong performance in cross-datasets and cross-generator compared to both zero-shot and supervised baselines, while remaining inference-efficient. Building on our strong performance in cross-generator scenario, we intend to investigate if SPADE can be adapted to model attribution task. Additionally, since the LLR decomposes into per-token contributions, we plan to investigate at the token level which characteristics is captured.

\section*{Limitations}
\label{sec:limitations}
Our findings indicate that using different backbone LLMs can significantly change the performance of all detectors. A more systematic comparison across all baselines would better clarify the impact of the backbone choice on each detector, in particular contrasting SPADE with the other supervised baselines.

A second limitation is that, like most state-of-the-art detectors, SPADE’s performance drops on AI-generated texts that was paraphrased to evade existing detectors (see Section \ref{sec:attacks}). Since SPADE is supervised, it can be trained on datasets containing perturbed texts, which we leave to future work.

Finally, our evaluation primarily focuses on texts generated by instruction-tuned models. As shown in Table \ref{tab:base_vs_instruct} in Appendix~\ref{appendix:base_instruct}, SPADE performs less effectively on texts from base models. While we leave further investigation to future work, this drop may stem from the exclusive use of instruction-tuned models in training or the weaker influence of instructions on base-model generation.


\section*{Ethical Considerations}
\label{sec:ethical_consideration}

\paragraph{Datasets.}
We evaluate on publicly available benchmarks for machine-generated text detection (RAID, EvoBench, DetectRL-X and MIRAGE) and use them only for non-commercial research, in line with their intended purpose and licenses (see Appendix~\ref{appendix:license}).

\paragraph{Societal Impact.}
Although SPADE transfers better than the baselines and is practical to deploy thanks to its fast inference and small storage requirements, the prediction itself is not bulletproof. Performance may degrade in specific scenarios, especially regarding calibration on in-domain data. Human judgment remains necessary, especially in high-stakes domains.

\section*{Acknowledgments}
This work has been financially supported by ChapsVision and the ANRT through CIFRE grant n°2025/1102. We thank GENCI for granting the access to the HPC resources of IDRIS under the allocation 2026-AD011017568 where most of our computation was done on the Jean-Zay supercomputers.


\bibliography{custom}

\appendix
\label{sec:appendix}

\section{Details of Data}
\label{appendix:data}
The source dataset per domain across all datasets: MIRAGE \citep{fu2025detectanyllm}, EvoBench \citep{yu-etal-2025-evobench}, RAID \citep{dugan-etal-2024-raid}, and DetectRL-X \citep{wu-etal-2026-detectrl}) are all different, except for XSum, which appears in both MIRAGE and EvoBench.

\subsection{MIRAGE}
\label{appendix:mirage}
We split MIRAGE-DIG at the document level with the ratio 70/15/15. MIRAGE-SIG is split at the document-level using the 2,384 shared human-written texts to avoid any leakage. We split by count: 500 training / 942 validation / 942 test pairs. Both SIG and DIG share the same source dataset. All source datasets are chosen as prior to 2021 to avoid any possible AI-generated texts corruption. The selected domains: academic, mail, website, news, comment. 
\begin{itemize}
    \item \textbf{Academic} contains BigPatent \citep{sharma-etal-2019-bigpatent}, NeurIPS, ArXiv, PubMed-Abstracts \citep{jin-etal-2019-pubmedqa} as datasets.  
    \item \textbf{eMail} contains data from Enron-Emails \citep{10.1007/978-3-540-30115-8_22}
    \item \textbf{Website} contains data from OpenWebText \citep{Gokaslan2019OpenWeb}
    \item \textbf{News} contains data from CNNDailyMails\citep{see-etal-2017-get, NIPS2015_afdec700}, XSum \citep{narayan-etal-2018-dont}, XLSum \citep{hasan-etal-2021-xl} 
    \item \textbf{Comment}: Amazon-Review \citep{10.1145/2507157.2507163, NIPS2015_250cf8b5}
\end{itemize}

MIRAGE uses 9 families of instruct LLMs (GPT, Claude, DeepSeek, Gemini, Qwen, LLaMA, Grok, Moonshot, and Doubao) with three types of prompt -- continuation, polishing to a specific writing style, and rephrasing.

\subsection{EvoBench}
Detection is inconsistent between models and versions of the models. EvoBench provides human-written texts from five domains, where each human-written text is paired with an AI-generated text from several black-box and white-box LLMs, covering several versions of these LLMs. The source datasets are the following:
\begin{itemize}
    \item \textbf{Harmful} contains data from \citep{kula2026multilingualmodelscheckworthysocial}.
    \item \textbf{PeerRead} contains data from \citep{kang-etal-2018-dataset}
    \item \textbf{PubMed} contains data from \citep{jin-etal-2019-pubmedqa}
    \item \textbf{Writing} contains data from \citep{fan-etal-2018-hierarchical}.
    \item \textbf{News} contains data from XSum \citep{narayan-etal-2018-dont}.
\end{itemize}

EvoBench and MIRAGE both draws human-written text from PubMed \citep{jin-etal-2019-pubmedqa}. However, they take two disjoint part of the dataset: EvoBench uses the Question-Answering dataset; MIRAGE uses the Abstract dataset.

\subsection{RAID}
\paragraph{Domains} RAID contains human-written and AI-generated texts spanning eight different domains: Abstracts, Books \citep{bamman2013newalignmentmethodsdiscriminative}, News \citep{10.1145/1143844.1143892}, Poetry\footnote{\url{https://www.kaggle.com/datasets/michaelarman/poemsdataset}}, Recipes \citep{bien-etal-2020-recipenlg}, Reddit posts \citep{volske-etal-2017-tl}, Movie reviews \citep{maas-etal-2011-learning}, and Wikipedia \citep{aaditya_bhat_2023}.

\paragraph{Source Models} The machine-generated texts come from 11 models: GPT-2 \citep{radford2019language}, GPT-3 \citep{NEURIPS2022_b1efde53}, ChatGPT, GPT-4 \citep{openai2024gpt4technicalreport}, MPT and MPT-Chat \citep{MosaicML2023Introducing}, Mistral and Mistral-Chat \citep{jiang2023mistral7b}, Cohere and Cohere-Chat, and LLaMA-2-70B-Chat \citep{touvron2023llama2openfoundation}. We focus solely on the instruction-tuned models and remove the base models Mistral, Cohere, GPT-2, and MPT from the evaluation.

\subsection{DetectRL-X}
DetectRL-X provides human-written texts and AI-generated texts on 6 different domains, using both white-box and black-box LLMs, and across 8 languages. The source datasets are the following:
\begin{itemize}
        \item \textbf{Academic Writing}: from Directory of Open Access Journals (DOAJ).
        \item \textbf{News}: mostly from BBC.
        \item \textbf{Novel}: corresponds to the creative fiction texts extracted from Common Crawl.
        \item \textbf{SEO}: seach engine optimization content from public and private data from the authors's organization.
        \item \textbf{Wiki}: encyclopedia entries from Wikipedia across various topics and languages.
        \item \textbf{WebText}: corresponds to general web content extracted from Common Crawl.
\end{itemize}

\subsection{License}
\label{appendix:license}
We use the datasets solely for non-commercial academic evaluation consistent with their intended use and licenses.
\begin{table}[ht]
    \centering
    \small
    \begin{tabular}{l l}
      \hline
      \textbf{Dataset} & \textbf{License} \\
      \hline
      MIRAGE \citep{fu2025detectanyllm}     & Pi-Lab License 1.0 \\
      EvoBench \citep{yu-etal-2025-evobench}  & MIT \\
      RAID \citep{dugan-etal-2024-raid}       & MIT \\
      DetectRL-X \citep{wu-etal-2026-detectrl} & Apache 2.0 \\
      \hline
    \end{tabular}
    \caption{Licenses of the evaluation datasets.}
    \label{tab:licenses}
\end{table}

\section{Base models vs. Instruct models}
\label{appendix:base_instruct}
A limitation of SPADE is its drop in performance when detecting text generated by base models. To quantify it, we report on RAID the AUROC of SPADE per generator family, for the instruct and the base version of each model.
\begin{table}[ht]
\centering
\small
\begin{tabular}{l cc c}
  \hline
  \textbf{Family} & \textbf{Instruct} & \textbf{Base} & $\Delta$ \\
  \hline
  Cohere  & $\mathbf{88.4}$ & $76.9$ & $-11.4$ \\
  Mistral & $\mathbf{98.0}$ & $76.7$ & $-21.3$ \\
  MPT     & $\mathbf{97.0}$ & $79.0$ & $-18.0$ \\
  \hline
  Mean    & $\mathbf{94.4}$ & $77.6$ & $-16.9$ \\
  \hline
\end{tabular}
\caption{AUC of SPADE-DIG on RAID per family, for the instruct and the base version of each generator, with Qwen3-8B-Base as backbone (mean over three seeds). $\Delta$ = Base $-$ Instruct. \textbf{Bold} indicates the higher AUC of the two model types.}
\label{tab:base_vs_instruct}
\end{table}

\section{Implementation Details}
\label{appendix:implementation}
All models are run with PyTorch 2.10 and Hugging Face Transformers 5.17 in bfloat16, with FlashAttention-2 (flash-attn 2.8.1) for training. Every detector scores at most the first 512 tokens of a text (backbone tokenizer, no special tokens). AUROC, ROC curves and the thresholds (Youden-J, TPR at 5\% FPR) are computed with scikit-learn 1.9.1; pAUC@5\% FPR is the normalized area under the ROC curve for FPR $\le 0.05$. F1 is the F1 of the AI class at the Youden-J threshold fitted on the MIRAGE validation split.

\subsection{Backbone type choices}
For zero-shot methods, we follow the original implementations with different backbones, keeping each paper’s base/instruct choice where possible. We report in the following table these choices (Table \ref{tab:baseline_backbone_choice}).

\begin{table}[ht]
\centering
\small
    \begin{tabular}{l cc}
      \hline
      \textbf{Detector} & \textbf{Base} & \textbf{Instruct} \\
      \hline
      Likelihood     & \checkmark & \ding{55} \\
      Entropy        & \checkmark & \ding{55} \\
      DNA-GPT        & \checkmark & \ding{55} \\
      Fast-DetectGPT & \checkmark & \checkmark \\
      Binoculars     & \checkmark & \checkmark \\
      NTS            & \checkmark & \ding{55} \\
      \hline
      RAIDAR           & \ding{55}  & \checkmark \\
      DPIC           & \ding{55}  & \checkmark \\
      ImBD           & \checkmark & \ding{55} \\
      \hline
    \end{tabular}
\caption{Backbone type used by each baseline.}
\label{tab:baseline_backbone_choice}
\end{table}

\subsection{Hyperparameter Selection}
\label{appendix:hyperparameter}
For hyperparameter selection, we trained for 3 epochs with a batch size of 16 on MIRAGE-DIG, and for 20 epochs on MIRAGE-SIG. We select checkpoints and hyperparameters with pAUC@5\% FPR rather than AUC or TPR, because AUC and TPR approach 1 after only a few epochs, which makes them unsuitable for selection. The only tuned hyperparameter is the learning rate, which differs across backbones. Learning rates are selected on MIRAGE-DIG validation (seed 0, pAUC@5\% FPR) from \{1e-5, 3e-5, 1e-4, 3e-4\}, extended to 1e-3 and 3e-3 when the best value lay at the edge of the grid (Falcon-7B only). The selected learning rates are 3e-4 for Qwen3-8B-Base and LLaMA-3.1-8B, and 3e-3 for Falcon-7B. MIRAGE-SIG uses the same learning rate per backbone. Following \citet{lester-etal-2021-power}, we use a soft prompt of length $64$, initialized with the embeddings of random vocabulary tokens.

\subsection{Zero-shot methods}
\paragraph{Binoculars} \citep{pmlr-v235-hans24a} is a zero-shot method based on the ratio between the perplexity and the cross-perplexity that uses two scoring models: an observer and a performer. The cross-perplexity acts as a normalizer measuring whether the tokens are surprising relative to another surrogate model. We use three different model pairs (Falcon-7b / Falcon-7b-Instruct, LLaMA-3.1-8B / LLaMA-3.1-8B-Instruct, and Qwen3-8B-Base / Qwen3-8B).

\paragraph{Fast-DetectGPT} \citep{bao2024fastdetectgpt} following DetectGPT \citep{pmlr-v202-mitchell23a} is a zero-shot method using the conditional probability curvature where AI-generated text lives on a local maximum of that curvature. The initial paper uses GPT-J-6B as the sampling and GPT-Neo-2.7B as the scoring models, and the GitHub repository proposes to use LLaMA-3-8B/LLaMA-3-8B-Instruct which is supposed to outperform the previously chosen models. We use three different model pairs (Falcon-7b / Falcon-7b-Instruct, LLaMA-3.1-8B / LLaMA-3.1-8B-Instruct, and Qwen3-8B-Base / Qwen3-8B).

\paragraph{NTS} \citep{ma-etal-2026-zero} is a zero-shot method using a difference between high-temperature and low-temperature scoring on the input text and uses temperature sensitivity as its detection signal. We use three different model pairs (Falcon-7B, LLaMA-3.1-8B, and Qwen3-8B-Base) as our scoring model.

\subsection{Regeneration-based methods}
We generate all regenerated texts of all datasets using vLLM 0.19.1 for efficiency. We carefully choose the three comparison backbones as (Qwen3-8B without thinking, LLaMA-3.1-8B-Instruct, and Falcon-7B-Instruct) for RAIDAR and DPIC on a single seed $0$. As for DNA-GPT, we replace the backbone by their respective base pair (Qwen3-8B-Base, LLaMA-3.1-8B, and Falcon-7B).

We then obtain three splits (train, validation, test) over three different seeds $0, 27, 78$ using the same document-level splits.

\textbf{DNA-GPT} \citep{ICLR2024_d4ce6738}. We use the the original protocol ($K = 10$, temperature 0.7). We keep the default values for all other parameters. Since we are computing the score using our own scoring model, we run DNA-GPT in the white-box setting.

\textbf{RAIDAR} \citep{ICLR2024_1888b9df}. We use the same $7$ 7 prompts as in the original paper with the specified temperature $0.7$ like in their paper. We let the other value per default of the backbone model.

\textbf{DPIC} \citep{NEURIPS2024_1d35af80}. We use the same prompt and the same specified temperature $0.2$ like in their paper and vary only the backbone model. We use the same embedding model "Alibaba-NLP/gte-Qwen2-7B-instruct".

\subsection{Supervised methods}
We train on the same splits as SPADE so that only the methodology differs.

\textbf{RoBERTa}. We follow the standard training setup for RoBERTa, with a learning rate of 2e-5, 5 epochs and a batch size of 8, and choose the best epoch on the validation set with the same metric, pAUC@5\% FPR.

\textbf{ImBD} \citep{10.1609/aaai.v39i22.34525}. Following their implementation, we optimize it as we do SPADE: same split, learning-rate grid search, and the same validation metric, pAUC@5\% FPR. The selected learning rates are 1e-4 for Qwen3-8B-Base and Falcon-7B, and 3e-5 for LLaMA-3.1-8B. ImBD on Falcon-7B degenerates at higher learning rates (validation pAUC@5\% FPR drops from 0.88 at 1e-4 to 0.37 at 3e-4) so we did not extend the grid.

\section{More details on the Main Results}
\label{appendix:main-result}
\subsection{Overall Performances}
\label{appendix:overall-performance}

We report the per-domain results of Section \ref{sec:main-result} with two additional backbone models (LLaMA-3.1-8B and Falcon-7B) for completeness. With LLaMA-3.1-8B, SPADE remains the state-of-the-art in both benchmark EvoBench (Table \ref{tab:evobench_dig_llama}) and RAID (Table \ref{tab:raid_dig_llama}). With Falcon-7B, SPADE is competitive with Binoculars and NTS on RAID (Table \ref{tab:raid_dig_falcon}) but remains the best overall on EvoBench (Table \ref{tab:evobench_dig_falcon}). This result is consistent with \S\ref{sec:robustness}, where SPADE's performance drops on less capable and older backbones.

\begin{table*}[t]
  \centering
  \resizebox{\textwidth}{!}{%
  \begin{tabular}{l ccc ccc ccc ccc ccc}
    \hline
    & \multicolumn{3}{c}{\textbf{harmful}} & \multicolumn{3}{c}{\textbf{peerread}} & \multicolumn{3}{c}{\textbf{pubmed}} & \multicolumn{3}{c}{\textbf{writing}} & \multicolumn{3}{c}{\textbf{Avg.}} \\
    \cmidrule(lr){2-4} \cmidrule(lr){5-7} \cmidrule(lr){8-10} \cmidrule(lr){11-13} \cmidrule(lr){14-16}
    \textbf{Detector} & AUC & TPR & F1 & AUC & TPR & F1 & AUC & TPR & F1 & AUC & TPR & F1 & AUC & TPR & F1 \\
    \hline
    \rowcolor{gray!20}
    \multicolumn{16}{c}{\textbf{Zero-Shot Detectors}} \\
    Likelihood & $57.1$ & $8.45$ & $28.7_{1.2}$ & $82.6$ & $37.6$ & $47.7_{2.1}$ & $90.8$ & $62.1$ & $82.9_{0.2}$ & $97.4$ & $89.1$ & $90.5_{0.5}$ & $82.0$ & $49.3$ & $62.4_{0.9}$ \\
    Entropy & $55.6$ & $6.16$ & $35.2_{3.8}$ & $59.1$ & $10.5$ & $12.9_{2.6}$ & $90.7$ & $68.3$ & $75.7_{1.1}$ & $96.7$ & $86.0$ & $89.1_{1.3}$ & $75.5$ & $42.8$ & $53.2_{1.6}$ \\
    DNA-GPT \citeyearpar{ICLR2024_d4ce6738} & $54.7$ & $5.49$ & $36.2_{7.0}$ & $89.0$ & $52.2$ & $73.7_{7.9}$ & $87.5$ & $48.7$ & $78.3_{1.6}$ & $90.4$ & $59.3$ & $75.1_{8.4}$ & $80.4$ & $41.4$ & $65.9_{6.2}$ \\
    Fast-DetectGPT \citeyearpar{bao2024fastdetectgpt} & $56.0$ & $6.32$ & $30.6_{2.5}$ & $86.6$ & $50.0$ & $78.9_{0.2}$ & $71.3$ & $22.3$ & $48.6_{3.6}$ & $80.7$ & $41.8$ & $72.2_{1.2}$ & $73.6$ & $30.1$ & $57.5_{1.9}$ \\
    Binoculars \citeyearpar{pmlr-v235-hans24a} & $54.4$ & $6.96$ & $35.4_{1.4}$ & $86.6$ & $51.5$ & $79.2_{0.4}$ & $69.1$ & $24.0$ & $54.4_{1.6}$ & $82.8$ & $51.0$ & $72.6_{0.8}$ & $73.2$ & $33.4$ & $60.4_{0.9}$ \\
    NTS \citeyearpar{ma-etal-2026-zero} & $48.9$ & $6.54$ & $58.2_{0.9}$ & $73.3$ & $28.0$ & $68.9_{0.7}$ & $46.2$ & $1.41$ & $56.0_{1.6}$ & $59.8$ & $16.8$ & $55.6_{1.2}$ & $57.1$ & $13.2$ & $59.7_{0.7}$ \\
    \hline
    \rowcolor{gray!20}
    \multicolumn{16}{c}{\textbf{Supervised Detectors}} \\
    RoBERTa & $95.3_{1.7}$ & $75.9_{8.2}$ & $82.7_{2.6}$ & $89.4_{0.8}$ & $63.5_{4.8}$ & $58.3_{6.2}$ & $74.1_{1.3}$ & $27.4_{5.2}$ & $57.1_{7.7}$ & $96.1_{1.2}$ & $86.4_{3.9}$ & $90.2_{1.7}$ & $88.7_{0.4}$ & $63.3_{2.4}$ & $72.1_{3.3}$ \\
    RAIDAR \citeyearpar{ICLR2024_1888b9df} & $76.6_{2.3}$ & $26.4_{3.2}$ & $69.2_{3.0}$ & $67.3_{0.6}$ & $20.0_{1.6}$ & $56.8_{1.5}$ & $64.5_{3.2}$ & $10.7_{3.8}$ & $49.2_{6.7}$ & $81.8_{0.8}$ & $35.5_{1.0}$ & $76.0_{0.5}$ & $72.6_{1.2}$ & $23.2_{1.2}$ & $62.8_{2.1}$ \\
    DPIC \citeyearpar{NEURIPS2024_1d35af80} & $80.5_{0.7}$ & $46.1_{2.9}$ & $66.3_{2.3}$ & $76.1_{1.9}$ & $41.9_{2.9}$ & $55.1_{5.9}$ & $66.0_{1.4}$ & $26.2_{1.7}$ & $33.5_{0.0}$ & $92.0_{0.1}$ & $72.8_{0.8}$ & $84.4_{0.2}$ & $78.6_{0.8}$ & $46.8_{1.8}$ & $59.8_{1.5}$ \\
    ImBD \citeyearpar{10.1609/aaai.v39i22.34525} & $92.3_{1.9}$ & $63.7_{10.0}$ & $69.2_{0.8}$ & $93.2_{1.2}$ & $67.0_{5.2}$ & $83.1_{0.4}$ & $87.3_{4.0}$ & $49.4_{7.9}$ & $75.2_{4.5}$ & $98.7_{0.5}$ & $94.5_{4.5}$ & $92.4_{5.6}$ & $92.9_{1.2}$ & $68.7_{4.0}$ & $80.0_{0.2}$ \\
    \rowcolor{spadered!10}
    \textbf{SPADE (ours)} & $\mathbf{96.7}_{0.4}$ & $\mathbf{88.4}_{1.5}$ & $\mathbf{87.3}_{1.1}$ & $\mathbf{95.8}_{0.9}$ & $\mathbf{89.1}_{1.8}$ & $\mathbf{91.0}_{0.8}$ & $\mathbf{98.3}_{0.3}$ & $\mathbf{95.1}_{0.5}$ & $\mathbf{85.2}_{5.2}$ & $\mathbf{99.8}_{0.0}$ & $\mathbf{98.9}_{0.2}$ & $\mathbf{98.3}_{0.5}$ & $\mathbf{97.7}_{0.3}$ & $\mathbf{92.9}_{0.7}$ & $\mathbf{90.5}_{1.7}$ \\
    \hline
  \end{tabular}%
  }
  \caption{AUC, TPR, and F1 on EvoBench (XSum excluded) with \textbf{Qwen3-8B-Base} as scoring backbone for the zero-shot detectors and SPADE. All supervised baselines are trained on MIRAGE-DIG and the reported number corresponds to mean$_{\text{std}}$ over three seeds; zero-shot AUC and TPR are single runs. F1 is computed at the Youden-J threshold fitted on each run's MIRAGE validation split. When a criterion ranks human texts above AI texts on the validation split, no threshold exists and its F1 is 0. \textbf{Bold} indicates the best method per column.}
  \label{tab:evobench_dig_qwen}
\end{table*}

\begin{table*}[t]
  \centering
  \resizebox{\textwidth}{!}{%
  \begin{tabular}{l ccc ccc ccc ccc ccc ccc ccc ccc ccc}
    \hline
    & \multicolumn{3}{c}{\textbf{Abstracts}} & \multicolumn{3}{c}{\textbf{Books}} & \multicolumn{3}{c}{\textbf{News}} & \multicolumn{3}{c}{\textbf{Poetry}} & \multicolumn{3}{c}{\textbf{Recipes}} & \multicolumn{3}{c}{\textbf{Reddit}} & \multicolumn{3}{c}{\textbf{Reviews}} & \multicolumn{3}{c}{\textbf{Wiki}} & \multicolumn{3}{c}{\textbf{Avg.}} \\
    \cmidrule(lr){2-4} \cmidrule(lr){5-7} \cmidrule(lr){8-10} \cmidrule(lr){11-13} \cmidrule(lr){14-16} \cmidrule(lr){17-19} \cmidrule(lr){20-22} \cmidrule(lr){23-25} \cmidrule(lr){26-28}
    \textbf{Detector} & AUC & TPR & F1 & AUC & TPR & F1 & AUC & TPR & F1 & AUC & TPR & F1 & AUC & TPR & F1 & AUC & TPR & F1 & AUC & TPR & F1 & AUC & TPR & F1 & AUC & TPR & F1 \\
    \hline
    \rowcolor{gray!20}
    \multicolumn{28}{c}{\textbf{Zero-Shot Detectors}} \\
    Likelihood & $75.5$ & $10.0$ & $61.7$ & $82.9$ & $11.2$ & $75.6$ & $75.6$ & $6.67$ & $62.4$ & $75.3$ & $4.00$ & $65.4$ & $77.1$ & $14.2$ & $52.4$ & $78.1$ & $4.70$ & $72.0$ & $81.3$ & $9.59$ & $73.8$ & $78.6$ & $4.86$ & $58.5$ & $78.1$ & $8.16$ & $65.2$ \\
    Entropy & $68.6$ & $10.0$ & $56.5$ & $80.8$ & $11.8$ & $74.1$ & $71.2$ & $6.00$ & $58.8$ & $75.3$ & $4.00$ & $66.4$ & $77.7$ & $16.3$ & $51.6$ & $76.1$ & $4.70$ & $71.3$ & $78.6$ & $9.59$ & $70.4$ & $75.2$ & $4.86$ & $56.4$ & $75.4$ & $8.41$ & $63.2$ \\
    DNA-GPT \citeyearpar{ICLR2024_d4ce6738} & $71.0$ & $20.7$ & $59.8$ & $77.5$ & $24.9$ & $67.8$ & $60.1$ & $10.7$ & $53.3$ & $68.3$ & $10.7$ & $55.6$ & $67.4$ & $14.2$ & $54.0$ & $76.2$ & $15.4$ & $62.6$ & $77.8$ & $24.0$ & $66.2$ & $75.3$ & $14.6$ & $63.7$ & $71.7$ & $16.9$ & $60.4$ \\
    Fast-DetectGPT \citeyearpar{bao2024fastdetectgpt} & $76.6$ & $32.3$ & $62.3$ & $66.4$ & $18.9$ & $58.2$ & $73.0$ & $18.7$ & $60.9$ & $59.4$ & $19.0$ & $50.5$ & $49.6$ & $15.6$ & $39.1$ & $72.2$ & $25.0$ & $61.1$ & $66.5$ & $28.1$ & $55.6$ & $72.5$ & $41.7$ & $58.1$ & $67.0$ & $24.9$ & $55.7$ \\
    Binoculars \citeyearpar{pmlr-v235-hans24a} & $80.7$ & $50.7$ & $64.3$ & $75.1$ & $40.8$ & $65.3$ & $76.1$ & $38.7$ & $64.1$ & $64.9$ & $27.3$ & $56.2$ & $53.8$ & $24.1$ & $44.1$ & $76.7$ & $36.2$ & $65.8$ & $72.3$ & $34.9$ & $61.9$ & $75.6$ & $43.8$ & $61.6$ & $71.9$ & $37.1$ & $60.4$ \\
    NTS \citeyearpar{ma-etal-2026-zero} & $76.3$ & $33.3$ & $64.6$ & $67.8$ & $20.1$ & $54.6$ & $69.1$ & $16.0$ & $58.9$ & $63.6$ & $15.3$ & $52.4$ & $74.4$ & $21.3$ & $55.4$ & $72.9$ & $20.1$ & $62.8$ & $63.1$ & $10.3$ & $52.1$ & $63.9$ & $18.8$ & $50.0$ & $68.9$ & $19.4$ & $56.4$ \\
    \hline
    \rowcolor{gray!20}
    \multicolumn{28}{c}{\textbf{Supervised Detectors}} \\
    RoBERTa & $87.9$ & $45.1$ & $54.9$ & $83.8$ & $46.2$ & $66.2$ & $79.4$ & $43.2$ & $57.8$ & $89.4$ & $65.6$ & $78.1$ & $78.7$ & $36.4$ & $43.5$ & $86.7$ & $50.2$ & $62.2$ & $90.5$ & $64.2$ & $75.3$ & $80.0$ & $41.8$ & $52.8$ & $84.6$ & $49.1$ & $61.3$ \\
    RAIDAR \citeyearpar{ICLR2024_1888b9df} & $59.9$ & $23.3$ & $50.6$ & $65.0$ & $27.6$ & $55.3$ & $73.6$ & $42.2$ & $62.1$ & $64.2$ & $12.4$ & $56.4$ & $77.9$ & $22.5$ & $39.3$ & $76.6$ & $49.4$ & $54.3$ & $71.3$ & $27.9$ & $53.4$ & $65.5$ & $24.8$ & $45.1$ & $69.3$ & $28.8$ & $52.1$ \\
    DPIC \citeyearpar{NEURIPS2024_1d35af80} & $86.9$ & $68.9$ & $78.2$ & $87.2$ & $66.9$ & $78.7$ & $89.6$ & $72.9$ & $79.9$ & $90.4$ & $73.6$ & $73.0$ & $82.1$ & $49.2$ & $71.3$ & $82.4$ & $56.6$ & $69.6$ & $93.8$ & $79.2$ & $84.4$ & $80.2$ & $49.3$ & $64.4$ & $86.6$ & $64.6$ & $74.9$ \\
    ImBD \citeyearpar{10.1609/aaai.v39i22.34525} & $88.0$ & $15.1$ & $78.8$ & $91.4$ & $27.8$ & $84.2$ & $85.0$ & $16.4$ & $71.1$ & $93.4$ & $58.4$ & $68.9$ & $83.7$ & $10.9$ & $76.6$ & $89.4$ & $14.8$ & $81.9$ & $88.1$ & $25.8$ & $78.9$ & $81.4$ & $13.2$ & $74.0$ & $87.5$ & $22.8$ & $76.8$ \\
    \rowcolor{spadered!10}
    \textbf{SPADE (ours)} & $\mathbf{98.3}$ & $\mathbf{91.1}$ & $\mathbf{92.0}$ & $\mathbf{96.7}$ & $\mathbf{84.4}$ & $\mathbf{87.4}$ & $\mathbf{90.2}$ & $\mathbf{77.6}$ & $\mathbf{82.0}$ & $\mathbf{97.2}$ & $\mathbf{92.1}$ & $\mathbf{83.4}$ & $\mathbf{94.9}$ & $\mathbf{79.4}$ & $\mathbf{82.9}$ & $\mathbf{97.9}$ & $\mathbf{85.7}$ & $\mathbf{89.0}$ & $\mathbf{95.8}$ & $\mathbf{87.0}$ & $\mathbf{86.5}$ & $\mathbf{94.6}$ & $\mathbf{74.4}$ & $\mathbf{81.9}$ & $\mathbf{95.7}$ & $\mathbf{84.0}$ & $\mathbf{85.6}$ \\
    \hline
  \end{tabular}%
  }
  \caption{AUC, TPR, and F1 on RAID with \textbf{Qwen3-8B-Base} as scoring backbone for the zero-shot detectors and SPADE. All supervised baselines are trained on MIRAGE-DIG and the reported number corresponds to the mean over three seeds; zero-shot AUC and TPR are single runs. F1 is computed at the Youden-J threshold fitted on each run's MIRAGE validation split. When a criterion ranks human texts above AI texts on the validation split, no threshold exists and its F1 is 0. \textbf{Bold} indicates the best method per column.}
  \label{tab:raid_dig_qwen}
\end{table*}

\begin{table*}[t]
  \centering
  \resizebox{\textwidth}{!}{%
  \begin{tabular}{l ccc ccc ccc ccc ccc}
    \hline
    & \multicolumn{3}{c}{\textbf{harmful}} & \multicolumn{3}{c}{\textbf{peerread}} & \multicolumn{3}{c}{\textbf{pubmed}} & \multicolumn{3}{c}{\textbf{writing}} & \multicolumn{3}{c}{\textbf{Avg.}} \\
    \cmidrule(lr){2-4} \cmidrule(lr){5-7} \cmidrule(lr){8-10} \cmidrule(lr){11-13} \cmidrule(lr){14-16}
    \textbf{Detector} & AUC & TPR & F1 & AUC & TPR & F1 & AUC & TPR & F1 & AUC & TPR & F1 & AUC & TPR & F1 \\
    \hline
    \rowcolor{gray!20}
    \multicolumn{16}{c}{\textbf{Zero-Shot Detectors}} \\
    Likelihood & $46.9$ & $2.60$ & $25.9_{5.5}$ & $76.9$ & $30.2$ & $44.1_{7.4}$ & $82.8$ & $39.5$ & $75.1_{0.7}$ & $94.8$ & $75.3$ & $86.0_{2.2}$ & $75.4$ & $36.9$ & $57.7_{3.9}$ \\
    Entropy & $42.1$ & $2.48$ & $0.00_{0.0}$ & $45.4$ & $4.93$ & $0.00_{0.0}$ & $76.7$ & $32.4$ & $0.00_{0.0}$ & $89.4$ & $66.9$ & $0.00_{0.0}$ & $63.4$ & $26.7$ & $0.00_{0.0}$ \\
    DNA-GPT \citeyearpar{ICLR2024_d4ce6738} & $49.3$ & $5.30$ & $37.1_{5.0}$ & $82.0$ & $39.3$ & $69.4_{4.7}$ & $82.2$ & $31.5$ & $74.7_{1.4}$ & $87.5$ & $52.7$ & $77.5_{2.4}$ & $75.2$ & $32.2$ & $64.7_{3.3}$ \\
    Fast-DetectGPT \citeyearpar{bao2024fastdetectgpt} & $67.0$ & $15.5$ & $39.4_{2.0}$ & $92.6$ & $71.8$ & $82.8_{0.6}$ & $87.4$ & $42.1$ & $76.9_{1.0}$ & $98.5$ & $93.1$ & $88.1_{0.9}$ & $86.4$ & $55.6$ & $71.8_{0.7}$ \\
    Binoculars \citeyearpar{pmlr-v235-hans24a} & $65.7$ & $15.3$ & $51.8_{0.9}$ & $92.4$ & $73.9$ & $85.6_{0.1}$ & $86.9$ & $44.1$ & $80.4_{0.4}$ & $98.8$ & $94.5$ & $91.0_{0.9}$ & $86.0$ & $56.9$ & $77.2_{0.1}$ \\
    NTS \citeyearpar{ma-etal-2026-zero} & $49.3$ & $3.75$ & $56.0_{0.5}$ & $72.4$ & $22.5$ & $68.7_{0.1}$ & $37.6$ & $0.47$ & $47.9_{1.0}$ & $81.5$ & $54.2$ & $72.8_{0.2}$ & $60.2$ & $20.2$ & $61.3_{0.4}$ \\
    \hline
    \rowcolor{gray!20}
    \multicolumn{16}{c}{\textbf{Supervised Detectors}} \\
    RoBERTa & $\mathbf{95.3}_{1.7}$ & $75.9_{8.2}$ & $82.7_{2.6}$ & $89.4_{0.8}$ & $63.5_{4.8}$ & $58.3_{6.2}$ & $74.1_{1.3}$ & $27.4_{5.2}$ & $57.1_{7.7}$ & $96.1_{1.2}$ & $86.4_{3.9}$ & $90.2_{1.7}$ & $88.7_{0.4}$ & $63.3_{2.4}$ & $72.1_{3.3}$ \\
    RAIDAR \citeyearpar{ICLR2024_1888b9df} & $61.5_{1.7}$ & $14.8_{3.8}$ & $64.3_{1.6}$ & $67.7_{1.4}$ & $24.0_{6.5}$ & $64.3_{0.4}$ & $51.9_{2.2}$ & $5.89_{1.8}$ & $58.9_{1.7}$ & $78.6_{0.5}$ & $30.6_{1.0}$ & $72.3_{0.5}$ & $64.9_{0.5}$ & $18.8_{1.9}$ & $65.0_{0.8}$ \\
    DPIC \citeyearpar{NEURIPS2024_1d35af80} & $79.7_{1.5}$ & $46.1_{2.9}$ & $67.5_{1.2}$ & $74.0_{1.6}$ & $36.4_{1.9}$ & $49.6_{1.3}$ & $66.4_{0.4}$ & $26.0_{1.5}$ & $27.4_{2.7}$ & $91.7_{0.2}$ & $72.6_{1.3}$ & $83.3_{1.1}$ & $77.9_{0.8}$ & $45.3_{1.4}$ & $56.9_{1.2}$ \\
    ImBD \citeyearpar{10.1609/aaai.v39i22.34525} & $91.6_{1.4}$ & $59.7_{4.4}$ & $73.8_{4.3}$ & $82.5_{2.9}$ & $40.2_{12.4}$ & $73.8_{1.1}$ & $83.7_{1.1}$ & $38.5_{5.7}$ & $75.4_{1.5}$ & $96.5_{0.9}$ & $85.0_{4.2}$ & $86.3_{1.9}$ & $88.6_{1.0}$ & $55.9_{2.5}$ & $77.3_{2.1}$ \\
    \rowcolor{spadered!10}
    \textbf{SPADE (ours)} & $94.8_{0.7}$ & $\mathbf{76.9}_{2.9}$ & $\mathbf{85.0}_{3.0}$ & $\mathbf{97.7}_{0.2}$ & $\mathbf{90.5}_{0.6}$ & $\mathbf{91.7}_{0.9}$ & $\mathbf{98.0}_{0.1}$ & $\mathbf{94.2}_{1.3}$ & $\mathbf{87.6}_{7.5}$ & $\mathbf{99.0}_{0.4}$ & $\mathbf{96.3}_{1.5}$ & $\mathbf{96.0}_{1.4}$ & $\mathbf{97.4}_{0.0}$ & $\mathbf{89.5}_{0.3}$ & $\mathbf{90.1}_{2.0}$ \\
    \hline
  \end{tabular}%
  }
  \caption{AUC, TPR, and F1 on EvoBench (XSum excluded) with \textbf{LLaMA-3.1-8B} as scoring backbone for the zero-shot detectors and SPADE. All supervised baselines are trained on MIRAGE-DIG and the reported number corresponds to mean$_{\text{std}}$ over three seeds; zero-shot AUC and TPR are single runs. F1 is computed at the Youden-J threshold fitted on each run's MIRAGE validation split. When a criterion ranks human texts above AI texts on the validation split, no threshold exists and its F1 is 0. \textbf{Bold} indicates the best method per column.}
  \label{tab:evobench_dig_llama}
\end{table*}

\begin{table*}[t]
  \centering
  \resizebox{\textwidth}{!}{%
  \begin{tabular}{l ccc ccc ccc ccc ccc ccc ccc ccc ccc}
    \hline
    & \multicolumn{3}{c}{\textbf{Abstracts}} & \multicolumn{3}{c}{\textbf{Books}} & \multicolumn{3}{c}{\textbf{News}} & \multicolumn{3}{c}{\textbf{Poetry}} & \multicolumn{3}{c}{\textbf{Recipes}} & \multicolumn{3}{c}{\textbf{Reddit}} & \multicolumn{3}{c}{\textbf{Reviews}} & \multicolumn{3}{c}{\textbf{Wiki}} & \multicolumn{3}{c}{\textbf{Avg.}} \\
    \cmidrule(lr){2-4} \cmidrule(lr){5-7} \cmidrule(lr){8-10} \cmidrule(lr){11-13} \cmidrule(lr){14-16} \cmidrule(lr){17-19} \cmidrule(lr){20-22} \cmidrule(lr){23-25} \cmidrule(lr){26-28}
    \textbf{Detector} & AUC & TPR & F1 & AUC & TPR & F1 & AUC & TPR & F1 & AUC & TPR & F1 & AUC & TPR & F1 & AUC & TPR & F1 & AUC & TPR & F1 & AUC & TPR & F1 & AUC & TPR & F1 \\
    \hline
    \rowcolor{gray!20}
    \multicolumn{28}{c}{\textbf{Zero-Shot Detectors}} \\
    Likelihood & $73.7$ & $10.0$ & $59.1$ & $79.3$ & $11.8$ & $70.0$ & $71.8$ & $6.00$ & $56.8$ & $70.3$ & $4.00$ & $63.0$ & $72.9$ & $8.51$ & $51.1$ & $76.0$ & $5.37$ & $67.6$ & $79.7$ & $9.59$ & $68.3$ & $75.1$ & $4.86$ & $54.9$ & $74.9$ & $7.52$ & $61.3$ \\
    Entropy & $61.8$ & $10.0$ & $0.00$ & $76.4$ & $11.8$ & $0.00$ & $66.0$ & $5.33$ & $0.00$ & $68.9$ & $2.67$ & $0.00$ & $73.0$ & $6.38$ & $0.00$ & $72.5$ & $4.70$ & $0.00$ & $76.9$ & $8.90$ & $0.00$ & $71.9$ & $4.86$ & $0.00$ & $70.9$ & $6.84$ & $0.00$ \\
    DNA-GPT \citeyearpar{ICLR2024_d4ce6738} & $73.1$ & $33.3$ & $56.6$ & $76.7$ & $33.7$ & $65.5$ & $58.0$ & $17.3$ & $51.7$ & $67.1$ & $8.00$ & $56.9$ & $66.2$ & $12.1$ & $52.6$ & $74.7$ & $22.8$ & $63.2$ & $73.9$ & $20.5$ & $61.2$ & $70.9$ & $15.3$ & $58.2$ & $70.1$ & $20.4$ & $58.2$ \\
    Fast-DetectGPT \citeyearpar{bao2024fastdetectgpt} & $91.1$ & $73.3$ & $76.9$ & $86.3$ & $60.9$ & $70.1$ & $84.9$ & $57.3$ & $68.1$ & $87.5$ & $66.7$ & $66.7$ & $82.3$ & $56.0$ & $62.9$ & $89.1$ & $64.4$ & $74.6$ & $87.6$ & $60.3$ & $67.4$ & $86.0$ & $57.6$ & $70.7$ & $86.9$ & $62.1$ & $69.7$ \\
    Binoculars \citeyearpar{pmlr-v235-hans24a} & $92.9$ & $82.7$ & $78.3$ & $90.0$ & $66.9$ & $77.7$ & $87.2$ & $68.7$ & $72.5$ & $91.6$ & $77.3$ & $\mathbf{78.1}$ & $85.2$ & $70.9$ & $66.6$ & $91.8$ & $71.1$ & $78.8$ & $91.2$ & $65.8$ & $75.5$ & $86.0$ & $57.6$ & $72.1$ & $89.5$ & $70.1$ & $74.9$ \\
    NTS \citeyearpar{ma-etal-2026-zero} & $89.2$ & $56.0$ & $75.8$ & $76.4$ & $36.7$ & $65.0$ & $77.8$ & $31.3$ & $65.8$ & $67.9$ & $23.3$ & $57.1$ & $70.2$ & $30.5$ & $56.0$ & $76.0$ & $34.2$ & $64.1$ & $73.9$ & $21.2$ & $62.5$ & $70.6$ & $23.6$ & $57.4$ & $75.3$ & $32.1$ & $63.0$ \\
    \hline
    \rowcolor{gray!20}
    \multicolumn{28}{c}{\textbf{Supervised Detectors}} \\
    RoBERTa & $87.9$ & $45.1$ & $54.9$ & $83.8$ & $46.2$ & $66.2$ & $79.4$ & $43.2$ & $57.8$ & $89.4$ & $65.6$ & $\mathbf{78.1}$ & $78.7$ & $36.4$ & $43.5$ & $86.7$ & $50.2$ & $62.2$ & $90.5$ & $64.2$ & $75.3$ & $80.0$ & $41.8$ & $52.8$ & $84.6$ & $49.1$ & $61.3$ \\
    RAIDAR \citeyearpar{ICLR2024_1888b9df} & $66.7$ & $25.1$ & $56.3$ & $69.2$ & $41.4$ & $61.2$ & $76.7$ & $42.4$ & $66.2$ & $68.2$ & $17.1$ & $55.5$ & $79.1$ & $57.4$ & $71.9$ & $79.4$ & $52.1$ & $58.3$ & $75.8$ & $35.6$ & $64.9$ & $70.1$ & $29.6$ & $56.1$ & $73.1$ & $37.6$ & $61.3$ \\
    DPIC \citeyearpar{NEURIPS2024_1d35af80} & $87.7$ & $69.6$ & $77.9$ & $88.6$ & $69.4$ & $78.9$ & $86.9$ & $69.3$ & $78.0$ & $89.1$ & $73.0$ & $69.4$ & $84.5$ & $54.4$ & $72.5$ & $82.2$ & $55.7$ & $69.1$ & $94.1$ & $79.2$ & $\mathbf{82.6}$ & $80.4$ & $51.9$ & $64.4$ & $86.7$ & $65.3$ & $74.1$ \\
    ImBD \citeyearpar{10.1609/aaai.v39i22.34525} & $92.8$ & $58.4$ & $82.7$ & $\mathbf{96.6}$ & $80.1$ & $\mathbf{86.3}$ & $\mathbf{93.5}$ & $63.1$ & $\mathbf{82.1}$ & $\mathbf{96.7}$ & $\mathbf{86.7}$ & $68.0$ & $90.7$ & $52.2$ & $74.5$ & $95.7$ & $74.9$ & $\mathbf{89.5}$ & $\mathbf{95.0}$ & $70.8$ & $82.3$ & $88.3$ & $53.0$ & $76.1$ & $93.7$ & $67.4$ & $80.2$ \\
    \rowcolor{spadered!10}
    \textbf{SPADE (ours)} & $\mathbf{98.0}$ & $\mathbf{88.7}$ & $\mathbf{88.3}$ & $95.0$ & $\mathbf{83.2}$ & $84.5$ & $91.7$ & $\mathbf{74.7}$ & $80.2$ & $94.9$ & $76.1$ & $76.3$ & $\mathbf{94.2}$ & $\mathbf{79.5}$ & $\mathbf{80.2}$ & $\mathbf{96.1}$ & $\mathbf{83.7}$ & $86.9$ & $94.7$ & $\mathbf{83.7}$ & $80.9$ & $\mathbf{91.8}$ & $\mathbf{69.4}$ & $\mathbf{76.5}$ & $\mathbf{94.5}$ & $\mathbf{79.9}$ & $\mathbf{81.7}$ \\
    \hline
  \end{tabular}%
  }
  \caption{AUC, TPR, and F1 on RAID with \textbf{LLaMA-3.1-8B} as scoring backbone for the zero-shot detectors and SPADE. All supervised baselines are trained on MIRAGE-DIG and the reported number corresponds to the mean over three seeds; zero-shot AUC and TPR are single runs. F1 is computed at the Youden-J threshold fitted on each run's MIRAGE validation split. When a criterion ranks human texts above AI texts on the validation split, no threshold exists and its F1 is 0. \textbf{Bold} indicates the best method per column.}
  \label{tab:raid_dig_llama}
\end{table*}

\begin{table*}[t]
  \centering
  \resizebox{\textwidth}{!}{%
  \begin{tabular}{l ccc ccc ccc ccc ccc}
    \hline
    & \multicolumn{3}{c}{\textbf{harmful}} & \multicolumn{3}{c}{\textbf{peerread}} & \multicolumn{3}{c}{\textbf{pubmed}} & \multicolumn{3}{c}{\textbf{writing}} & \multicolumn{3}{c}{\textbf{Avg.}} \\
    \cmidrule(lr){2-4} \cmidrule(lr){5-7} \cmidrule(lr){8-10} \cmidrule(lr){11-13} \cmidrule(lr){14-16}
    \textbf{Detector} & AUC & TPR & F1 & AUC & TPR & F1 & AUC & TPR & F1 & AUC & TPR & F1 & AUC & TPR & F1 \\
    \hline
    \rowcolor{gray!20}
    \multicolumn{16}{c}{\textbf{Zero-Shot Detectors}} \\
    Likelihood & $43.7$ & $2.31$ & $0.00_{0.0}$ & $72.6$ & $24.7$ & $0.00_{0.0}$ & $82.7$ & $33.2$ & $0.00_{0.0}$ & $89.0$ & $56.3$ & $0.00_{0.0}$ & $72.0$ & $29.1$ & $0.00_{0.0}$ \\
    Entropy & $41.8$ & $1.95$ & $0.00_{0.0}$ & $43.1$ & $4.38$ & $0.00_{0.0}$ & $67.9$ & $15.9$ & $0.00_{0.0}$ & $64.3$ & $17.2$ & $0.00_{0.0}$ & $54.3$ & $9.85$ & $0.00_{0.0}$ \\
    DNA-GPT \citeyearpar{ICLR2024_d4ce6738} & $44.8$ & $2.96$ & $37.7_{2.6}$ & $79.9$ & $35.6$ & $65.0_{2.5}$ & $81.3$ & $41.3$ & $73.6_{0.3}$ & $90.1$ & $60.4$ & $80.4_{0.4}$ & $74.0$ & $35.1$ & $64.2_{1.4}$ \\
    Fast-DetectGPT \citeyearpar{bao2024fastdetectgpt} & $54.5$ & $8.23$ & $25.1_{1.0}$ & $83.9$ & $39.9$ & $67.0_{1.0}$ & $91.5$ & $62.1$ & $74.2_{1.1}$ & $98.5$ & $95.0$ & $91.0_{0.5}$ & $82.1$ & $51.3$ & $64.3_{0.6}$ \\
    Binoculars \citeyearpar{pmlr-v235-hans24a} & $54.9$ & $8.41$ & $42.6_{0.4}$ & $84.2$ & $45.7$ & $74.8_{0.1}$ & $91.2$ & $59.6$ & $81.9_{0.1}$ & $\mathbf{98.7}$ & $\mathbf{95.6}$ & $\mathbf{91.1}_{0.0}$ & $82.3$ & $52.3$ & $72.6_{0.1}$ \\
    NTS \citeyearpar{ma-etal-2026-zero} & $54.7$ & $6.99$ & $54.5_{1.3}$ & $70.1$ & $14.0$ & $67.4_{0.7}$ & $63.5$ & $8.28$ & $60.3_{0.5}$ & $96.7$ & $87.9$ & $88.1_{0.6}$ & $71.3$ & $29.3$ & $67.6_{0.5}$ \\
    \hline
    \rowcolor{gray!20}
    \multicolumn{16}{c}{\textbf{Supervised Detectors}} \\
    RoBERTa & $\mathbf{95.3}_{1.7}$ & $\mathbf{75.9}_{8.2}$ & $82.7_{2.6}$ & $89.4_{0.8}$ & $63.5_{4.8}$ & $58.3_{6.2}$ & $74.1_{1.3}$ & $27.4_{5.2}$ & $57.1_{7.7}$ & $96.1_{1.2}$ & $86.4_{3.9}$ & $90.2_{1.7}$ & $88.7_{0.4}$ & $63.3_{2.4}$ & $72.1_{3.3}$ \\
    RAIDAR \citeyearpar{ICLR2024_1888b9df} & $49.6_{1.9}$ & $5.45_{0.6}$ & $45.0_{6.6}$ & $53.8_{0.9}$ & $11.0_{2.7}$ & $49.6_{0.8}$ & $37.8_{2.9}$ & $2.96_{0.1}$ & $21.0_{5.9}$ & $59.0_{0.8}$ & $13.7_{0.7}$ & $49.5_{1.4}$ & $50.0_{0.7}$ & $8.26_{0.7}$ & $41.3_{3.6}$ \\
    DPIC \citeyearpar{NEURIPS2024_1d35af80} & $80.5_{0.5}$ & $44.8_{0.9}$ & $70.2_{2.1}$ & $75.0_{0.3}$ & $35.6_{0.8}$ & $42.4_{4.3}$ & $65.4_{1.1}$ & $26.2_{3.9}$ & $24.9_{3.4}$ & $92.1_{0.2}$ & $73.6_{0.8}$ & $84.1_{0.1}$ & $78.3_{0.4}$ & $45.0_{1.1}$ & $55.4_{2.2}$ \\
    ImBD \citeyearpar{10.1609/aaai.v39i22.34525} & $87.9_{2.2}$ & $49.0_{9.0}$ & $73.7_{5.5}$ & $83.0_{5.3}$ & $37.8_{12.9}$ & $72.9_{2.4}$ & $80.5_{4.8}$ & $36.2_{13.3}$ & $70.9_{0.6}$ & $96.3_{0.3}$ & $85.7_{3.1}$ & $80.7_{4.7}$ & $86.9_{2.2}$ & $52.2_{5.4}$ & $74.5_{1.8}$ \\
    \rowcolor{spadered!10}
    \textbf{SPADE (ours)} & $93.9_{0.5}$ & $74.7_{1.0}$ & $\mathbf{85.0}_{2.0}$ & $\mathbf{92.8}_{0.4}$ & $\mathbf{74.2}_{5.8}$ & $\mathbf{85.1}_{0.9}$ & $\mathbf{92.3}_{3.4}$ & $\mathbf{76.5}_{9.3}$ & $\mathbf{82.3}_{9.0}$ & $97.3_{0.8}$ & $86.5_{4.0}$ & $90.7_{2.3}$ & $\mathbf{94.1}_{0.9}$ & $\mathbf{78.0}_{1.2}$ & $\mathbf{85.8}_{3.1}$ \\
    \hline
  \end{tabular}%
  }
  \caption{AUC, TPR, and F1 on EvoBench (XSum excluded) with \textbf{Falcon-7B} as scoring backbone for the zero-shot detectors and SPADE. All supervised baselines are trained on MIRAGE-DIG and the reported number corresponds to mean$_{\text{std}}$ over three seeds; zero-shot AUC and TPR are single runs. F1 is computed at the Youden-J threshold fitted on each run's MIRAGE validation split. When a criterion ranks human texts above AI texts on the validation split, no threshold exists and its F1 is 0. \textbf{Bold} indicates the best method per column.}
  \label{tab:evobench_dig_falcon}
\end{table*}

\begin{table*}[t]
  \centering
  \resizebox{\textwidth}{!}{%
  \begin{tabular}{l ccc ccc ccc ccc ccc ccc ccc ccc ccc}
    \hline
    & \multicolumn{3}{c}{\textbf{Abstracts}} & \multicolumn{3}{c}{\textbf{Books}} & \multicolumn{3}{c}{\textbf{News}} & \multicolumn{3}{c}{\textbf{Poetry}} & \multicolumn{3}{c}{\textbf{Recipes}} & \multicolumn{3}{c}{\textbf{Reddit}} & \multicolumn{3}{c}{\textbf{Reviews}} & \multicolumn{3}{c}{\textbf{Wiki}} & \multicolumn{3}{c}{\textbf{Avg.}} \\
    \cmidrule(lr){2-4} \cmidrule(lr){5-7} \cmidrule(lr){8-10} \cmidrule(lr){11-13} \cmidrule(lr){14-16} \cmidrule(lr){17-19} \cmidrule(lr){20-22} \cmidrule(lr){23-25} \cmidrule(lr){26-28}
    \textbf{Detector} & AUC & TPR & F1 & AUC & TPR & F1 & AUC & TPR & F1 & AUC & TPR & F1 & AUC & TPR & F1 & AUC & TPR & F1 & AUC & TPR & F1 & AUC & TPR & F1 & AUC & TPR & F1 \\
    \hline
    \rowcolor{gray!20}
    \multicolumn{28}{c}{\textbf{Zero-Shot Detectors}} \\
    Likelihood & $72.2$ & $8.00$ & $0.00$ & $78.5$ & $10.7$ & $0.00$ & $71.7$ & $4.00$ & $0.00$ & $72.1$ & $1.33$ & $0.00$ & $74.8$ & $24.8$ & $0.00$ & $76.3$ & $4.03$ & $0.00$ & $78.0$ & $6.16$ & $0.00$ & $77.5$ & $4.17$ & $0.00$ & $75.1$ & $7.90$ & $0.00$ \\
    Entropy & $51.2$ & $7.33$ & $0.00$ & $69.3$ & $7.69$ & $0.00$ & $57.0$ & $4.00$ & $0.00$ & $61.2$ & $2.00$ & $0.00$ & $68.8$ & $5.67$ & $0.00$ & $70.9$ & $4.03$ & $0.00$ & $71.3$ & $0.68$ & $0.00$ & $68.6$ & $4.17$ & $0.00$ & $64.8$ & $4.45$ & $0.00$ \\
    DNA-GPT \citeyearpar{ICLR2024_d4ce6738} & $82.2$ & $50.0$ & $63.8$ & $83.1$ & $39.6$ & $70.2$ & $71.6$ & $35.0$ & $54.4$ & $77.0$ & $16.0$ & $67.1$ & $78.2$ & $54.6$ & $53.4$ & $79.7$ & $34.9$ & $64.0$ & $81.8$ & $21.9$ & $66.3$ & $80.4$ & $36.1$ & $62.7$ & $79.2$ & $36.0$ & $62.7$ \\
    Fast-DetectGPT \citeyearpar{bao2024fastdetectgpt} & $87.6$ & $42.7$ & $74.8$ & $88.6$ & $48.1$ & $75.1$ & $83.0$ & $40.7$ & $68.9$ & $84.9$ & $38.3$ & $68.2$ & $83.6$ & $51.8$ & $71.0$ & $82.2$ & $11.1$ & $72.7$ & $85.6$ & $26.0$ & $70.9$ & $85.9$ & $45.5$ & $72.6$ & $85.2$ & $38.0$ & $71.8$ \\
    Binoculars \citeyearpar{pmlr-v235-hans24a} & $90.2$ & $\mathbf{72.0}$ & $76.4$ & $92.4$ & $\mathbf{69.8}$ & $\mathbf{81.1}$ & $87.5$ & $64.0$ & $75.6$ & $90.3$ & $71.3$ & $76.7$ & $87.3$ & $\mathbf{75.0}$ & $\mathbf{74.7}$ & $87.5$ & $54.4$ & $74.9$ & $91.4$ & $67.1$ & $77.5$ & $88.1$ & $\mathbf{66.0}$ & $\mathbf{73.7}$ & $89.3$ & $\mathbf{67.5}$ & $\mathbf{76.3}$ \\
    NTS \citeyearpar{ma-etal-2026-zero} & $\mathbf{92.4}$ & $\mathbf{72.0}$ & $75.9$ & $92.1$ & $67.5$ & $77.8$ & $86.4$ & $62.0$ & $72.3$ & $89.9$ & $\mathbf{72.7}$ & $72.2$ & $\mathbf{87.9}$ & $67.4$ & $73.6$ & $86.3$ & $61.7$ & $71.3$ & $87.5$ & $56.2$ & $71.2$ & $\mathbf{88.9}$ & $61.8$ & $72.8$ & $88.9$ & $65.2$ & $73.4$ \\
    \hline
    \rowcolor{gray!20}
    \multicolumn{28}{c}{\textbf{Supervised Detectors}} \\
    RoBERTa & $87.9$ & $45.1$ & $54.9$ & $83.8$ & $46.2$ & $66.2$ & $79.4$ & $43.2$ & $57.8$ & $89.4$ & $65.6$ & $\mathbf{78.1}$ & $78.7$ & $36.4$ & $43.5$ & $86.7$ & $50.2$ & $62.2$ & $90.5$ & $64.2$ & $75.3$ & $80.0$ & $41.8$ & $52.8$ & $84.6$ & $49.1$ & $61.3$ \\
    RAIDAR \citeyearpar{ICLR2024_1888b9df} & $52.8$ & $14.2$ & $41.5$ & $52.2$ & $8.68$ & $46.3$ & $59.9$ & $18.4$ & $45.9$ & $56.5$ & $7.56$ & $38.2$ & $42.5$ & $11.8$ & $22.4$ & $73.8$ & $22.6$ & $45.4$ & $60.5$ & $10.5$ & $42.4$ & $39.5$ & $7.87$ & $26.1$ & $54.7$ & $12.7$ & $38.5$ \\
    DPIC \citeyearpar{NEURIPS2024_1d35af80} & $87.3$ & $69.8$ & $\mathbf{77.8}$ & $89.4$ & $67.7$ & $77.4$ & $88.3$ & $70.2$ & $\mathbf{77.9}$ & $88.5$ & $70.7$ & $70.7$ & $83.6$ & $56.0$ & $69.8$ & $83.3$ & $55.3$ & $69.7$ & $\mathbf{93.5}$ & $\mathbf{73.1}$ & $\mathbf{80.5}$ & $81.1$ & $44.4$ & $62.2$ & $86.9$ & $63.4$ & $73.3$ \\
    ImBD \citeyearpar{10.1609/aaai.v39i22.34525} & $83.7$ & $20.7$ & $71.8$ & $\mathbf{92.8}$ & $58.6$ & $75.1$ & $85.7$ & $34.9$ & $75.4$ & $\mathbf{91.7}$ & $\mathbf{72.7}$ & $63.9$ & $79.3$ & $7.80$ & $62.0$ & $91.9$ & $41.2$ & $81.3$ & $89.7$ & $51.8$ & $72.2$ & $84.1$ & $36.1$ & $69.2$ & $87.4$ & $40.5$ & $71.4$ \\
    \rowcolor{spadered!10}
    \textbf{SPADE (ours)} & $87.5$ & $63.3$ & $73.7$ & $90.9$ & $62.7$ & $76.6$ & $\mathbf{89.7}$ & $\mathbf{71.1}$ & $76.7$ & $88.7$ & $65.3$ & $75.9$ & $86.5$ & $64.5$ & $74.2$ & $\mathbf{95.3}$ & $\mathbf{78.5}$ & $\mathbf{81.8}$ & $91.5$ & $69.2$ & $77.8$ & $87.5$ & $60.4$ & $71.7$ & $\mathbf{89.7}$ & $66.9$ & $76.1$ \\
    \hline
  \end{tabular}%
  }
  \caption{AUC, TPR, and F1 on RAID with \textbf{Falcon-7B} as scoring backbone for the zero-shot detectors and SPADE. All supervised baselines are trained on MIRAGE-DIG and the reported number corresponds to the mean over three seeds; zero-shot AUC and TPR are single runs. F1 is computed at the Youden-J threshold fitted on each run's MIRAGE validation split. When a criterion ranks human texts above AI texts on the validation split, no threshold exists and its F1 is 0. \textbf{Bold} indicates the best method per column.}
  \label{tab:raid_dig_falcon}
\end{table*}

\subsection{Generalization using one source model}
\label{appendix:generalization-using-one-source-model}
We report the results of supervised detectors across the nine families of Section \ref{sec:main-result} for the three backbone models, and also per family over the three seeds. Although some training families are present in the benchmarks, the three tables \ref{tab:per_family_qwen3_8b_base}, \ref{tab:per_family_llama31_8b}, and \ref{tab:per_family_falcon7b} show similar behavior for all supervised methods where some families transfer less well (e.g., Doubao, DeepSeek, and Qwen) than others that are not in-domain (Grok, Moonshot).

\begin{figure*}[t]
  \centering
  \includegraphics[width=\textwidth]{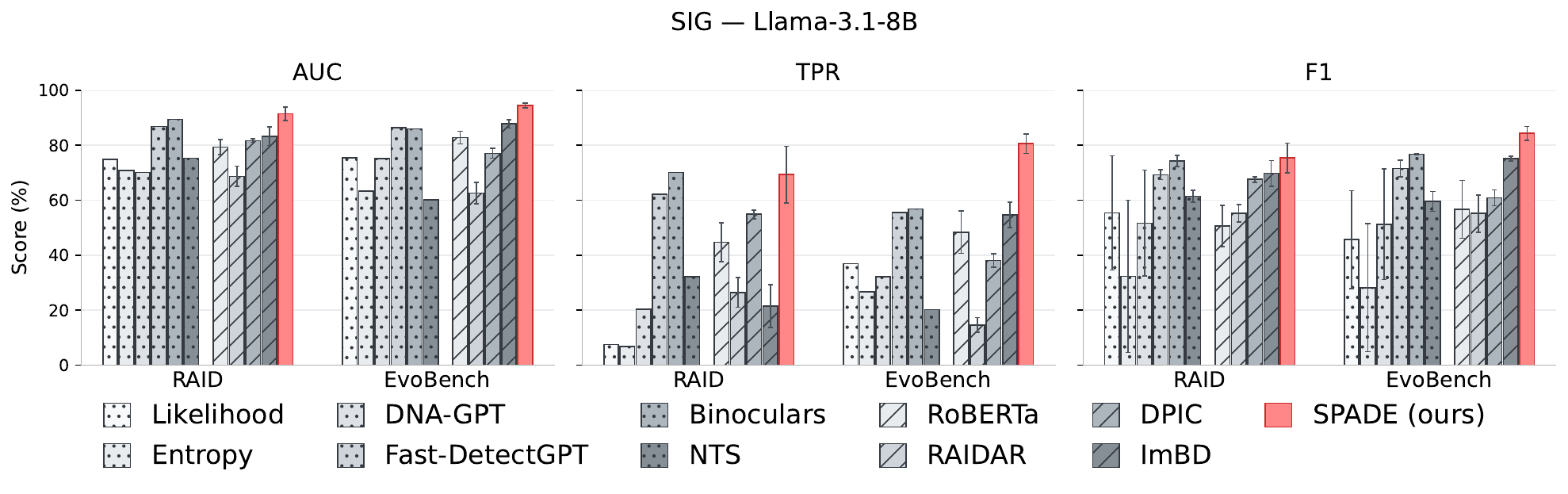}
  \caption{AUC, TPR, and F1 on EvoBench (XSum excluded) and RAID. All supervised baselines are trained on per family splits from MIRAGE-SIG (over three seeds). The reported number corresponds to mean$_{\text{std}}$ over the 9 families of the mean over three seeds per family using LLaMA-3.1-8B}
  \label{fig:figure_sig_llama}
\end{figure*}

\begin{figure*}[t]
  \centering
  \includegraphics[width=\textwidth]{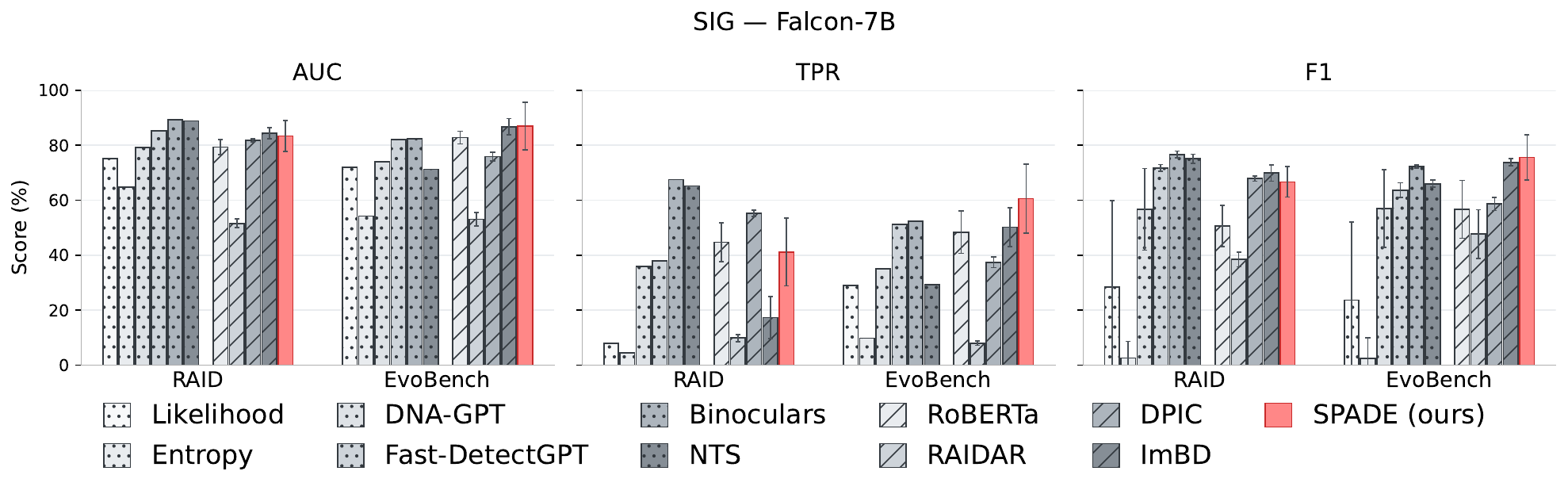}
  \caption{AUC, TPR, and F1 on EvoBench (XSum excluded) and RAID. All supervised baselines are trained on per family splits from MIRAGE-SIG (over three seeds). The reported number corresponds to mean$_{\text{std}}$ over the 9 families of the mean over three seeds per family using Falcon-7B}
  \label{fig:figure_sig_falcon}
\end{figure*}

\begin{table*}[t]
  \centering
  \resizebox{\textwidth}{!}{%
  \begin{tabular}{l ccc ccc ccc ccc ccc}
    \hline
    \textbf{Train} $\downarrow$ \ \textbf{Method} $\rightarrow$ & \multicolumn{3}{c}{RoBERTa} & \multicolumn{3}{c}{RAIDAR} & \multicolumn{3}{c}{DPIC} & \multicolumn{3}{c}{ImBD} & \multicolumn{3}{c}{\textbf{SPADE (ours)}} \\
    \cmidrule(lr){2-4} \cmidrule(lr){5-7} \cmidrule(lr){8-10} \cmidrule(lr){11-13} \cmidrule(lr){14-16}
    \textit{EvoBench} & AUC & TPR & F1 & AUC & TPR & F1 & AUC & TPR & F1 & AUC & TPR & F1 & AUC & TPR & F1 \\
    \hline
    \rowcolor{gray!20} \multicolumn{16}{l}{\textit{IN --- training family present in EvoBench}} \\
    Qwen & $83.2_{1.9}$ & $49.8_{4.7}$ & $66.7_{2.8}$ & $70.6_{2.0}$ & $19.9_{3.0}$ & $68.0_{1.6}$ & $76.8_{2.2}$ & $40.2_{2.7}$ & $58.7_{5.0}$ & $91.2_{1.1}$ & $66.3_{3.2}$ & $74.3_{1.2}$ & $95.3_{0.8}$ & $83.7_{5.3}$ & $83.2_{5.4}$ \\
    Claude & $81.4_{1.8}$ & $50.8_{1.7}$ & $52.1_{2.4}$ & $66.9_{1.7}$ & $20.6_{1.1}$ & $54.5_{1.7}$ & $74.8_{1.4}$ & $36.3_{1.7}$ & $53.1_{6.9}$ & $88.2_{1.9}$ & $59.2_{6.7}$ & $74.1_{0.6}$ & $94.6_{1.3}$ & $84.2_{2.9}$ & $85.2_{3.6}$ \\
    Gemini & $83.8_{0.9}$ & $50.6_{3.2}$ & $55.5_{2.9}$ & $64.1_{1.7}$ & $19.8_{1.5}$ & $52.6_{11.7}$ & $75.6_{1.3}$ & $36.6_{2.4}$ & $58.8_{3.9}$ & $91.9_{0.9}$ & $69.6_{1.5}$ & $74.2_{0.3}$ & $96.3_{0.2}$ & $87.6_{1.1}$ & $87.2_{3.2}$ \\
    GPT & $83.1_{2.0}$ & $51.0_{1.0}$ & $61.4_{6.3}$ & $67.0_{0.9}$ & $17.7_{1.3}$ & $56.8_{3.8}$ & $77.7_{0.5}$ & $38.9_{0.8}$ & $61.8_{5.2}$ & $91.8_{0.4}$ & $65.9_{1.1}$ & $74.2_{0.5}$ & $96.3_{0.5}$ & $88.6_{0.4}$ & $84.3_{1.4}$ \\
    LLaMA & $85.4_{1.3}$ & $55.2_{1.8}$ & $62.6_{3.5}$ & $67.3_{2.0}$ & $16.3_{2.1}$ & $59.5_{9.1}$ & $77.4_{0.5}$ & $39.1_{0.4}$ & $58.5_{2.9}$ & $93.1_{0.4}$ & $71.5_{1.7}$ & $74.7_{0.4}$ & $97.0_{0.4}$ & $89.4_{1.5}$ & $89.8_{0.7}$ \\
    \rowcolor{gray!20} \multicolumn{16}{l}{\textit{OUT --- training family absent from EvoBench}} \\
    Doubao & $77.8_{3.5}$ & $29.8_{4.3}$ & $33.1_{1.7}$ & $58.6_{3.4}$ & $12.7_{4.3}$ & $60.6_{4.1}$ & $72.0_{0.4}$ & $31.6_{1.2}$ & $54.6_{1.3}$ & $89.0_{1.7}$ & $55.8_{4.9}$ & $74.0_{0.7}$ & $93.6_{0.1}$ & $76.3_{3.4}$ & $85.3_{2.1}$ \\
    DeepSeek & $81.2_{1.1}$ & $43.0_{2.6}$ & $51.1_{7.9}$ & $72.4_{1.9}$ & $22.6_{2.3}$ & $63.4_{2.5}$ & $76.7_{1.5}$ & $38.3_{1.5}$ & $58.7_{5.9}$ & $89.2_{1.1}$ & $60.6_{1.8}$ & $74.8_{0.6}$ & $94.9_{0.4}$ & $84.0_{3.5}$ & $83.2_{3.1}$ \\
    Moonshot & $84.8_{1.3}$ & $53.6_{0.8}$ & $65.4_{0.3}$ & $68.1_{3.1}$ & $18.3_{3.2}$ & $62.8_{7.3}$ & $76.5_{1.1}$ & $39.3_{0.8}$ & $58.1_{7.4}$ & $91.3_{1.6}$ & $64.8_{6.2}$ & $73.4_{1.2}$ & $96.5_{0.6}$ & $88.1_{2.7}$ & $82.8_{1.0}$ \\
    Grok & $84.3_{0.5}$ & $51.9_{1.1}$ & $62.5_{3.9}$ & $65.9_{2.2}$ & $19.0_{4.0}$ & $59.8_{3.3}$ & $77.0_{0.4}$ & $40.3_{4.3}$ & $60.3_{3.3}$ & $90.7_{0.9}$ & $63.1_{2.8}$ & $74.4_{0.4}$ & $96.0_{0.7}$ & $88.5_{1.7}$ & $83.2_{3.7}$ \\
    \hline
    \textit{Mean} & $82.8_{2.3}$ & $48.4_{7.7}$ & $56.7_{10.5}$ & $66.8_{3.9}$ & $18.5_{2.8}$ & $59.8_{4.7}$ & $76.0_{1.8}$ & $37.8_{2.7}$ & $58.1_{2.7}$ & $90.7_{1.6}$ & $64.1_{5.0}$ & $74.2_{0.4}$ & $\mathbf{95.6}_{1.1}$ & $\mathbf{85.6}_{4.2}$ & $\mathbf{84.9}_{2.3}$ \\
    \hline\hline
    \textbf{Train} $\downarrow$ \ \textbf{Method} $\rightarrow$ & \multicolumn{3}{c}{RoBERTa} & \multicolumn{3}{c}{RAIDAR} & \multicolumn{3}{c}{DPIC} & \multicolumn{3}{c}{ImBD} & \multicolumn{3}{c}{\textbf{SPADE (ours)}} \\
    \cmidrule(lr){2-4} \cmidrule(lr){5-7} \cmidrule(lr){8-10} \cmidrule(lr){11-13} \cmidrule(lr){14-16}
    \textit{RAID} & AUC & TPR & F1 & AUC & TPR & F1 & AUC & TPR & F1 & AUC & TPR & F1 & AUC & TPR & F1 \\
    \hline
    \rowcolor{gray!20} \multicolumn{16}{l}{\textit{IN --- training family present in RAID}} \\
    GPT & $78.2_{2.3}$ & $40.3_{3.3}$ & $54.7_{6.8}$ & $66.2_{2.9}$ & $14.0_{8.5}$ & $50.9_{2.1}$ & $81.1_{1.4}$ & $55.7_{1.9}$ & $67.4_{2.0}$ & $82.0_{0.8}$ & $17.5_{1.8}$ & $71.6_{0.5}$ & $94.2_{0.9}$ & $80.9_{2.4}$ & $83.7_{1.2}$ \\
    LLaMA & $84.4_{1.1}$ & $56.9_{3.9}$ & $59.0_{2.8}$ & $63.5_{2.6}$ & $16.0_{7.2}$ & $52.5_{3.1}$ & $81.8_{0.2}$ & $56.1_{2.0}$ & $68.2_{1.4}$ & $82.0_{1.3}$ & $14.8_{3.8}$ & $72.0_{1.5}$ & $95.5_{1.3}$ & $84.1_{3.5}$ & $81.2_{5.8}$ \\
    \rowcolor{gray!20} \multicolumn{16}{l}{\textit{OUT --- training family absent from RAID}} \\
    Doubao & $78.8_{1.2}$ & $43.9_{10.4}$ & $43.7_{3.2}$ & $58.2_{1.7}$ & $4.44_{0.1}$ & $50.8_{1.1}$ & $80.5_{0.5}$ & $51.8_{1.7}$ & $66.6_{1.0}$ & $81.0_{0.4}$ & $11.0_{0.5}$ & $70.7_{1.7}$ & $89.7_{2.4}$ & $55.4_{9.3}$ & $73.7_{2.2}$ \\
    Grok & $81.5_{1.5}$ & $45.6_{3.6}$ & $54.5_{6.4}$ & $64.3_{0.4}$ & $4.75_{0.3}$ & $54.0_{0.7}$ & $80.9_{0.8}$ & $54.0_{0.4}$ & $67.6_{0.0}$ & $80.6_{0.3}$ & $15.4_{0.5}$ & $68.7_{2.5}$ & $91.7_{1.3}$ & $67.6_{8.3}$ & $78.0_{0.7}$ \\
    DeepSeek & $74.9_{2.4}$ & $31.0_{4.5}$ & $42.4_{11.4}$ & $66.4_{1.3}$ & $11.8_{7.6}$ & $46.1_{4.0}$ & $81.5_{0.6}$ & $54.7_{0.8}$ & $67.3_{0.8}$ & $78.6_{5.6}$ & $16.1_{7.5}$ & $63.0_{7.8}$ & $92.8_{0.4}$ & $71.0_{1.4}$ & $80.9_{3.1}$ \\
    Claude & $77.5_{0.3}$ & $44.3_{1.4}$ & $40.8_{2.4}$ & $62.5_{3.8}$ & $18.5_{2.5}$ & $41.9_{1.3}$ & $80.0_{0.6}$ & $52.6_{0.8}$ & $65.1_{2.0}$ & $85.3_{4.8}$ & $31.9_{17.9}$ & $69.8_{3.9}$ & $92.9_{1.5}$ & $78.4_{2.5}$ & $83.0_{1.6}$ \\
    Gemini & $77.5_{1.7}$ & $42.6_{4.5}$ & $45.5_{1.6}$ & $63.7_{1.5}$ & $4.47_{0.7}$ & $49.7_{1.1}$ & $81.8_{0.2}$ & $54.9_{0.7}$ & $66.1_{0.9}$ & $82.3_{2.3}$ & $19.6_{10.0}$ & $69.6_{3.6}$ & $93.9_{0.6}$ & $80.5_{3.9}$ & $82.5_{2.3}$ \\
    Qwen & $80.1_{1.3}$ & $47.4_{1.5}$ & $57.1_{3.5}$ & $66.1_{2.2}$ & $15.2_{10.2}$ & $51.3_{2.0}$ & $81.2_{0.5}$ & $53.5_{0.9}$ & $66.8_{0.7}$ & $80.4_{1.6}$ & $14.2_{1.9}$ & $69.6_{2.3}$ & $94.7_{0.6}$ & $80.5_{2.3}$ & $83.4_{1.8}$ \\
    Moonshot & $81.8_{2.1}$ & $50.1_{3.2}$ & $59.0_{2.2}$ & $65.1_{0.8}$ & $4.84_{2.4}$ & $53.4_{2.6}$ & $81.7_{0.9}$ & $56.8_{1.1}$ & $67.7_{0.9}$ & $83.6_{2.6}$ & $31.5_{12.2}$ & $69.9_{7.5}$ & $94.9_{0.9}$ & $81.7_{2.8}$ & $82.3_{2.1}$ \\
    \hline
    \textit{Mean} & $79.4_{2.8}$ & $44.7_{7.1}$ & $50.7_{7.5}$ & $64.0_{2.6}$ & $10.4_{5.8}$ & $50.1_{3.8}$ & $81.2_{0.6}$ & $54.5_{1.6}$ & $67.0_{1.0}$ & $81.8_{1.9}$ & $19.1_{7.5}$ & $69.5_{2.6}$ & $\mathbf{93.4}_{1.8}$ & $\mathbf{75.6}_{9.3}$ & $\mathbf{81.0}_{3.2}$ \\
    \hline
  \end{tabular}%
  }
  \caption{AUC, TPR, and F1 on EvoBench (top) and RAID (bottom) per training family, with \textbf{Qwen3-8B-Base} as backbone (RoBERTa has none). Each detector is trained on a single MIRAGE generator family and tested on the full benchmark. Per family: mean$_{\text{std}}$ across three seeds; \textit{Mean}: mean$_{\text{std}}$ across the nine training families. Rows are grouped by whether the training family is present in the benchmark (EvoBench: GPT, Claude, Gemini, LLaMA, Qwen; RAID: GPT, LLaMA) or absent from it, and sorted within each group by SPADE's TPR, lowest first. F1 is computed at the Youden-J threshold fitted on each run's MIRAGE validation split; when a criterion ranks human texts above AI texts on the validation split, no threshold exists and its F1 is 0. \textbf{Bold} indicates the best method on the cross-family mean.}
  \label{tab:per_family_qwen3_8b_base}
\end{table*}

\begin{table*}[t]
  \centering
  \resizebox{\textwidth}{!}{%
  \begin{tabular}{l ccc ccc ccc ccc ccc}
    \hline
    \textbf{Train} $\downarrow$ \ \textbf{Method} $\rightarrow$ & \multicolumn{3}{c}{RoBERTa} & \multicolumn{3}{c}{RAIDAR} & \multicolumn{3}{c}{DPIC} & \multicolumn{3}{c}{ImBD} & \multicolumn{3}{c}{\textbf{SPADE (ours)}} \\
    \cmidrule(lr){2-4} \cmidrule(lr){5-7} \cmidrule(lr){8-10} \cmidrule(lr){11-13} \cmidrule(lr){14-16}
    \textit{EvoBench} & AUC & TPR & F1 & AUC & TPR & F1 & AUC & TPR & F1 & AUC & TPR & F1 & AUC & TPR & F1 \\
    \hline
    \rowcolor{gray!20} \multicolumn{16}{l}{\textit{IN --- training family present in EvoBench}} \\
    GPT & $83.1_{2.0}$ & $51.0_{1.0}$ & $61.4_{6.3}$ & $65.4_{1.3}$ & $17.8_{0.1}$ & $57.2_{1.3}$ & $78.6_{0.9}$ & $39.5_{1.0}$ & $64.0_{1.5}$ & $86.7_{0.3}$ & $52.1_{3.0}$ & $74.6_{0.4}$ & $94.3_{1.0}$ & $78.5_{5.0}$ & $84.0_{1.1}$ \\
    Claude & $81.4_{1.8}$ & $50.8_{1.7}$ & $52.1_{2.4}$ & $64.1_{2.5}$ & $15.5_{1.7}$ & $51.4_{3.3}$ & $76.1_{0.6}$ & $37.5_{1.2}$ & $58.4_{2.4}$ & $87.5_{0.9}$ & $53.3_{2.6}$ & $76.0_{0.2}$ & $94.0_{0.5}$ & $80.6_{1.6}$ & $81.1_{3.5}$ \\
    Qwen & $83.2_{1.9}$ & $49.8_{4.7}$ & $66.7_{2.8}$ & $67.3_{1.4}$ & $15.9_{1.6}$ & $62.6_{3.6}$ & $78.2_{0.3}$ & $39.6_{0.5}$ & $61.4_{1.1}$ & $88.0_{1.0}$ & $57.1_{1.9}$ & $75.1_{0.4}$ & $94.5_{1.9}$ & $80.8_{7.5}$ & $83.3_{2.4}$ \\
    Gemini & $83.8_{0.9}$ & $50.6_{3.2}$ & $55.5_{2.9}$ & $55.8_{2.3}$ & $11.4_{1.3}$ & $40.8_{3.5}$ & $76.1_{0.9}$ & $35.3_{2.5}$ & $60.3_{5.5}$ & $89.4_{0.5}$ & $55.3_{1.3}$ & $76.3_{0.5}$ & $95.1_{0.2}$ & $81.7_{0.9}$ & $86.5_{1.3}$ \\
    LLaMA & $85.4_{1.3}$ & $55.2_{1.8}$ & $62.6_{3.5}$ & $63.7_{1.1}$ & $15.5_{1.4}$ & $56.6_{4.3}$ & $78.3_{0.9}$ & $38.3_{1.4}$ & $63.2_{4.1}$ & $88.6_{2.7}$ & $59.1_{7.2}$ & $75.1_{0.2}$ & $95.8_{0.2}$ & $85.9_{1.9}$ & $88.3_{0.5}$ \\
    \rowcolor{gray!20} \multicolumn{16}{l}{\textit{OUT --- training family absent from EvoBench}} \\
    Doubao & $77.8_{3.5}$ & $29.8_{4.3}$ & $33.1_{1.7}$ & $57.2_{0.4}$ & $8.88_{1.0}$ & $57.7_{1.5}$ & $73.1_{0.8}$ & $33.0_{2.1}$ & $55.9_{2.2}$ & $84.6_{0.9}$ & $44.1_{4.0}$ & $73.3_{0.9}$ & $93.1_{0.8}$ & $74.0_{4.5}$ & $84.3_{1.5}$ \\
    DeepSeek & $81.2_{1.1}$ & $43.0_{2.6}$ & $51.1_{7.9}$ & $65.6_{1.8}$ & $14.9_{1.0}$ & $56.7_{5.6}$ & $77.2_{0.8}$ & $38.8_{0.8}$ & $58.3_{4.3}$ & $87.5_{1.1}$ & $54.2_{1.1}$ & $75.3_{1.0}$ & $93.8_{1.1}$ & $78.2_{2.7}$ & $83.5_{4.7}$ \\
    Moonshot & $84.8_{1.3}$ & $53.6_{0.8}$ & $65.4_{0.3}$ & $62.9_{3.1}$ & $15.8_{3.7}$ & $62.8_{2.3}$ & $78.3_{0.9}$ & $40.7_{1.4}$ & $61.4_{3.7}$ & $89.0_{2.2}$ & $58.9_{5.8}$ & $75.5_{0.8}$ & $94.6_{1.5}$ & $80.9_{3.7}$ & $81.3_{2.3}$ \\
    Grok & $84.3_{0.5}$ & $51.9_{1.1}$ & $62.5_{3.9}$ & $61.7_{0.8}$ & $15.4_{1.4}$ & $50.6_{9.3}$ & $78.1_{0.2}$ & $40.3_{1.6}$ & $64.7_{4.0}$ & $89.0_{0.5}$ & $58.1_{2.2}$ & $75.6_{1.1}$ & $95.5_{0.9}$ & $84.6_{4.6}$ & $87.4_{0.7}$ \\
    \hline
    \textit{Mean} & $82.8_{2.3}$ & $48.4_{7.7}$ & $56.7_{10.5}$ & $62.6_{3.9}$ & $14.6_{2.7}$ & $55.2_{6.8}$ & $77.1_{1.8}$ & $38.1_{2.5}$ & $60.9_{2.9}$ & $87.8_{1.5}$ & $54.7_{4.7}$ & $75.2_{0.9}$ & $\mathbf{94.5}_{0.9}$ & $\mathbf{80.6}_{3.5}$ & $\mathbf{84.4}_{2.5}$ \\
    \hline\hline
    \textbf{Train} $\downarrow$ \ \textbf{Method} $\rightarrow$ & \multicolumn{3}{c}{RoBERTa} & \multicolumn{3}{c}{RAIDAR} & \multicolumn{3}{c}{DPIC} & \multicolumn{3}{c}{ImBD} & \multicolumn{3}{c}{\textbf{SPADE (ours)}} \\
    \cmidrule(lr){2-4} \cmidrule(lr){5-7} \cmidrule(lr){8-10} \cmidrule(lr){11-13} \cmidrule(lr){14-16}
    \textit{RAID} & AUC & TPR & F1 & AUC & TPR & F1 & AUC & TPR & F1 & AUC & TPR & F1 & AUC & TPR & F1 \\
    \hline
    \rowcolor{gray!20} \multicolumn{16}{l}{\textit{IN --- training family present in RAID}} \\
    GPT & $78.2_{2.3}$ & $40.3_{3.3}$ & $54.7_{6.8}$ & $69.5_{1.4}$ & $29.6_{2.6}$ & $56.1_{1.4}$ & $81.9_{0.7}$ & $55.8_{1.0}$ & $67.7_{0.9}$ & $82.5_{2.1}$ & $12.3_{2.2}$ & $69.5_{3.9}$ & $92.2_{0.6}$ & $69.7_{6.1}$ & $76.4_{0.5}$ \\
    LLaMA & $84.4_{1.1}$ & $56.9_{3.9}$ & $59.0_{2.8}$ & $70.4_{0.3}$ & $35.3_{1.7}$ & $58.6_{1.9}$ & $82.2_{0.5}$ & $55.8_{0.9}$ & $69.1_{0.9}$ & $87.5_{1.8}$ & $31.0_{2.4}$ & $74.8_{2.1}$ & $94.6_{1.1}$ & $81.9_{3.6}$ & $79.5_{3.2}$ \\
    \rowcolor{gray!20} \multicolumn{16}{l}{\textit{OUT --- training family absent from RAID}} \\
    Claude & $77.5_{0.3}$ & $44.3_{1.4}$ & $40.8_{2.4}$ & $68.8_{2.9}$ & $30.0_{5.7}$ & $51.6_{3.8}$ & $80.7_{0.5}$ & $54.6_{0.8}$ & $66.9_{1.4}$ & $83.3_{1.7}$ & $15.7_{3.4}$ & $70.6_{3.0}$ & $87.3_{3.8}$ & $52.4_{12.1}$ & $66.2_{8.4}$ \\
    Doubao & $78.8_{1.2}$ & $43.9_{10.4}$ & $43.7_{3.2}$ & $59.5_{1.2}$ & $16.2_{2.2}$ & $50.0_{3.2}$ & $81.1_{1.1}$ & $51.5_{2.2}$ & $66.6_{1.9}$ & $79.6_{0.8}$ & $17.3_{0.8}$ & $64.6_{1.4}$ & $87.5_{1.5}$ & $52.5_{3.7}$ & $66.3_{3.2}$ \\
    DeepSeek & $74.9_{2.4}$ & $31.0_{4.5}$ & $42.4_{11.4}$ & $68.7_{1.4}$ & $26.2_{9.1}$ & $52.8_{2.5}$ & $82.5_{0.7}$ & $55.4_{0.9}$ & $67.7_{0.4}$ & $79.7_{1.7}$ & $14.9_{3.9}$ & $65.1_{1.7}$ & $90.6_{1.6}$ & $69.9_{7.0}$ & $78.4_{4.0}$ \\
    Gemini & $77.5_{1.7}$ & $42.6_{4.5}$ & $45.5_{1.6}$ & $68.9_{1.5}$ & $23.3_{6.2}$ & $55.9_{2.5}$ & $81.7_{0.3}$ & $54.9_{1.0}$ & $65.8_{0.5}$ & $82.4_{1.6}$ & $22.6_{6.0}$ & $67.3_{2.3}$ & $92.3_{1.7}$ & $72.4_{10.4}$ & $76.4_{3.7}$ \\
    Qwen & $80.1_{1.3}$ & $47.4_{1.5}$ & $57.1_{3.5}$ & $71.4_{1.2}$ & $28.8_{4.9}$ & $57.5_{1.4}$ & $82.1_{0.3}$ & $55.2_{1.0}$ & $68.0_{0.8}$ & $88.2_{1.2}$ & $30.1_{6.0}$ & $77.0_{1.5}$ & $92.4_{2.6}$ & $72.4_{11.2}$ & $76.7_{3.8}$ \\
    Moonshot & $81.8_{2.1}$ & $50.1_{3.2}$ & $59.0_{2.2}$ & $72.2_{1.3}$ & $24.8_{9.3}$ & $59.5_{0.8}$ & $82.2_{0.7}$ & $57.4_{0.3}$ & $68.9_{1.1}$ & $86.8_{2.9}$ & $32.6_{7.3}$ & $74.4_{3.5}$ & $92.9_{1.1}$ & $75.3_{2.4}$ & $78.4_{3.4}$ \\
    Grok & $81.5_{1.5}$ & $45.6_{3.6}$ & $54.5_{6.4}$ & $68.7_{2.1}$ & $24.0_{7.1}$ & $55.3_{2.3}$ & $81.2_{0.1}$ & $53.9_{2.1}$ & $67.9_{0.3}$ & $79.8_{1.6}$ & $16.7_{2.9}$ & $65.1_{3.6}$ & $93.2_{1.9}$ & $77.9_{2.5}$ & $80.3_{0.2}$ \\
    \hline
    \textit{Mean} & $79.4_{2.8}$ & $44.7_{7.1}$ & $50.7_{7.5}$ & $68.7_{3.7}$ & $26.5_{5.4}$ & $55.3_{3.2}$ & $81.7_{0.6}$ & $54.9_{1.6}$ & $67.6_{1.0}$ & $83.3_{3.4}$ & $21.5_{7.8}$ & $69.8_{4.7}$ & $\mathbf{91.5}_{2.5}$ & $\mathbf{69.4}_{10.3}$ & $\mathbf{75.4}_{5.4}$ \\
    \hline
  \end{tabular}%
  }
  \caption{AUC, TPR, and F1 on EvoBench (top) and RAID (bottom) per training family, with \textbf{LLaMA-3.1-8B} as backbone (RoBERTa has none). Each detector is trained on a single MIRAGE generator family and tested on the full benchmark. Per family: mean$_{\text{std}}$ across three seeds; \textit{Mean}: mean$_{\text{std}}$ across the nine training families. Rows are grouped by whether the training family is present in the benchmark (EvoBench: GPT, Claude, Gemini, LLaMA, Qwen; RAID: GPT, LLaMA) or absent from it, and sorted within each group by SPADE's TPR, lowest first. F1 is computed at the Youden-J threshold fitted on each run's MIRAGE validation split; when a criterion ranks human texts above AI texts on the validation split, no threshold exists and its F1 is 0. \textbf{Bold} indicates the best method on the cross-family mean.}
  \label{tab:per_family_llama31_8b}
\end{table*}

\begin{table*}[t]
  \centering
  \resizebox{\textwidth}{!}{%
  \begin{tabular}{l ccc ccc ccc ccc ccc}
    \hline
    \textbf{Train} $\downarrow$ \ \textbf{Method} $\rightarrow$ & \multicolumn{3}{c}{RoBERTa} & \multicolumn{3}{c}{RAIDAR} & \multicolumn{3}{c}{DPIC} & \multicolumn{3}{c}{ImBD} & \multicolumn{3}{c}{\textbf{SPADE (ours)}} \\
    \cmidrule(lr){2-4} \cmidrule(lr){5-7} \cmidrule(lr){8-10} \cmidrule(lr){11-13} \cmidrule(lr){14-16}
    \textit{EvoBench} & AUC & TPR & F1 & AUC & TPR & F1 & AUC & TPR & F1 & AUC & TPR & F1 & AUC & TPR & F1 \\
    \hline
    \rowcolor{gray!20} \multicolumn{16}{l}{\textit{IN --- training family present in EvoBench}} \\
    Claude & $81.4_{1.8}$ & $50.8_{1.7}$ & $52.1_{2.4}$ & $56.0_{1.9}$ & $8.48_{1.0}$ & $52.6_{1.8}$ & $74.1_{0.7}$ & $36.1_{0.9}$ & $57.6_{0.5}$ & $88.6_{1.6}$ & $54.8_{4.4}$ & $75.7_{1.4}$ & $87.4_{2.3}$ & $55.8_{8.7}$ & $76.9_{5.4}$ \\
    GPT & $83.1_{2.0}$ & $51.0_{1.0}$ & $61.4_{6.3}$ & $50.1_{3.1}$ & $6.80_{0.9}$ & $31.4_{5.3}$ & $77.3_{0.5}$ & $38.6_{1.0}$ & $60.6_{2.9}$ & $85.8_{1.9}$ & $47.8_{5.3}$ & $72.4_{1.0}$ & $88.8_{2.4}$ & $56.7_{10.2}$ & $78.6_{3.0}$ \\
    Qwen & $83.2_{1.9}$ & $49.8_{4.7}$ & $66.7_{2.8}$ & $55.4_{3.6}$ & $8.54_{1.7}$ & $59.4_{3.3}$ & $77.0_{0.3}$ & $38.1_{0.9}$ & $59.0_{8.6}$ & $88.0_{0.5}$ & $53.0_{0.9}$ & $73.3_{1.9}$ & $89.7_{0.4}$ & $66.0_{4.1}$ & $81.6_{2.7}$ \\
    Gemini & $83.8_{0.9}$ & $50.6_{3.2}$ & $55.5_{2.9}$ & $50.7_{2.8}$ & $7.54_{2.1}$ & $36.0_{13.2}$ & $75.0_{1.1}$ & $35.4_{3.3}$ & $56.5_{6.7}$ & $87.2_{1.7}$ & $52.7_{3.7}$ & $74.1_{2.4}$ & $90.1_{4.1}$ & $67.9_{4.9}$ & $78.0_{5.0}$ \\
    LLaMA & $85.4_{1.3}$ & $55.2_{1.8}$ & $62.6_{3.5}$ & $54.9_{2.6}$ & $8.31_{0.3}$ & $54.3_{2.4}$ & $76.9_{0.9}$ & $38.4_{0.4}$ & $60.9_{2.4}$ & $88.6_{2.2}$ & $56.2_{5.8}$ & $74.5_{2.1}$ & $93.7_{0.8}$ & $74.3_{3.2}$ & $83.3_{1.1}$ \\
    \rowcolor{gray!20} \multicolumn{16}{l}{\textit{OUT --- training family absent from EvoBench}} \\
    Doubao & $77.8_{3.5}$ & $29.8_{4.3}$ & $33.1_{1.7}$ & $50.4_{2.5}$ & $7.44_{1.0}$ & $48.6_{3.1}$ & $73.0_{1.0}$ & $34.1_{1.0}$ & $55.2_{1.6}$ & $79.1_{2.5}$ & $32.3_{5.6}$ & $71.4_{4.6}$ & $67.2_{0.5}$ & $34.5_{4.5}$ & $54.6_{0.4}$ \\
    DeepSeek & $81.2_{1.1}$ & $43.0_{2.6}$ & $51.1_{7.9}$ & $56.1_{2.3}$ & $9.36_{0.9}$ & $49.6_{9.5}$ & $76.2_{0.9}$ & $37.0_{1.2}$ & $58.2_{3.4}$ & $86.9_{1.3}$ & $50.7_{4.5}$ & $74.3_{1.3}$ & $88.1_{1.5}$ & $57.4_{13.6}$ & $81.2_{1.4}$ \\
    Moonshot & $84.8_{1.3}$ & $53.6_{0.8}$ & $65.4_{0.3}$ & $51.5_{3.1}$ & $7.55_{2.1}$ & $51.4_{4.0}$ & $76.7_{0.4}$ & $40.0_{0.9}$ & $62.8_{1.5}$ & $88.2_{1.7}$ & $51.5_{6.3}$ & $74.6_{0.9}$ & $90.0_{3.8}$ & $67.3_{7.5}$ & $80.5_{0.2}$ \\
    Grok & $84.3_{0.5}$ & $51.9_{1.1}$ & $62.5_{3.9}$ & $53.0_{1.4}$ & $7.88_{0.5}$ & $45.8_{11.3}$ & $76.8_{0.9}$ & $39.1_{1.7}$ & $57.4_{2.4}$ & $88.5_{1.3}$ & $52.7_{3.3}$ & $73.9_{1.2}$ & $92.5_{2.2}$ & $72.0_{7.0}$ & $81.9_{2.5}$ \\
    \hline
    \textit{Mean} & $82.8_{2.3}$ & $48.4_{7.7}$ & $56.7_{10.5}$ & $53.1_{2.5}$ & $7.99_{0.8}$ & $47.7_{8.9}$ & $75.9_{1.5}$ & $37.4_{1.9}$ & $58.7_{2.4}$ & $86.8_{3.0}$ & $50.2_{7.1}$ & $73.8_{1.3}$ & $\mathbf{87.5}_{7.9}$ & $\mathbf{61.3}_{12.1}$ & $\mathbf{77.4}_{8.8}$ \\
    \hline\hline
    \textbf{Train} $\downarrow$ \ \textbf{Method} $\rightarrow$ & \multicolumn{3}{c}{RoBERTa} & \multicolumn{3}{c}{RAIDAR} & \multicolumn{3}{c}{DPIC} & \multicolumn{3}{c}{ImBD} & \multicolumn{3}{c}{\textbf{SPADE (ours)}} \\
    \cmidrule(lr){2-4} \cmidrule(lr){5-7} \cmidrule(lr){8-10} \cmidrule(lr){11-13} \cmidrule(lr){14-16}
    \textit{RAID} & AUC & TPR & F1 & AUC & TPR & F1 & AUC & TPR & F1 & AUC & TPR & F1 & AUC & TPR & F1 \\
    \hline
    \rowcolor{gray!20} \multicolumn{16}{l}{\textit{IN --- training family present in RAID}} \\
    GPT & $78.2_{2.3}$ & $40.3_{3.3}$ & $54.7_{6.8}$ & $49.9_{1.7}$ & $9.27_{1.0}$ & $35.0_{2.3}$ & $81.4_{0.3}$ & $54.2_{2.2}$ & $67.4_{1.1}$ & $82.1_{0.9}$ & $5.76_{0.8}$ & $68.8_{3.9}$ & $85.5_{3.6}$ & $45.0_{9.5}$ & $66.1_{6.8}$ \\
    LLaMA & $84.4_{1.1}$ & $56.9_{3.9}$ & $59.0_{2.8}$ & $50.5_{1.2}$ & $7.81_{2.0}$ & $40.7_{2.0}$ & $82.2_{0.7}$ & $57.3_{1.5}$ & $69.7_{0.4}$ & $86.8_{3.1}$ & $24.7_{20.2}$ & $72.7_{4.0}$ & $90.7_{0.7}$ & $66.6_{5.7}$ & $74.8_{1.1}$ \\
    \rowcolor{gray!20} \multicolumn{16}{l}{\textit{OUT --- training family absent from RAID}} \\
    Doubao & $78.8_{1.2}$ & $43.9_{10.4}$ & $43.7_{3.2}$ & $49.8_{3.2}$ & $8.39_{2.4}$ & $38.5_{3.5}$ & $81.0_{0.6}$ & $53.8_{0.4}$ & $66.6_{0.4}$ & $80.4_{1.7}$ & $17.2_{10.3}$ & $63.8_{6.5}$ & $74.1_{0.7}$ & $26.3_{0.6}$ & $56.7_{3.2}$ \\
    Gemini & $77.5_{1.7}$ & $42.6_{4.5}$ & $45.5_{1.6}$ & $52.0_{1.5}$ & $10.5_{1.6}$ & $38.8_{2.6}$ & $82.9_{0.8}$ & $56.2_{0.6}$ & $66.7_{0.9}$ & $85.4_{1.2}$ & $20.8_{6.9}$ & $68.0_{7.7}$ & $83.2_{4.6}$ & $36.4_{6.3}$ & $64.0_{14.1}$ \\
    DeepSeek & $74.9_{2.4}$ & $31.0_{4.5}$ & $42.4_{11.4}$ & $52.4_{2.3}$ & $10.7_{0.4}$ & $35.3_{3.1}$ & $82.1_{0.2}$ & $55.6_{2.0}$ & $68.0_{0.4}$ & $83.6_{1.5}$ & $6.31_{1.7}$ & $70.8_{0.4}$ & $79.5_{1.2}$ & $36.5_{14.8}$ & $63.4_{7.8}$ \\
    Qwen & $80.1_{1.3}$ & $47.4_{1.5}$ & $57.1_{3.5}$ & $54.7_{1.5}$ & $11.4_{1.5}$ & $42.7_{0.9}$ & $82.0_{0.5}$ & $55.3_{0.4}$ & $68.2_{0.3}$ & $85.3_{0.5}$ & $19.2_{11.4}$ & $73.0_{0.9}$ & $83.5_{3.0}$ & $49.6_{8.6}$ & $66.7_{4.4}$ \\
    Moonshot & $81.8_{2.1}$ & $50.1_{3.2}$ & $59.0_{2.2}$ & $53.2_{1.3}$ & $10.9_{2.2}$ & $39.5_{2.2}$ & $82.0_{0.8}$ & $55.5_{1.1}$ & $68.6_{1.2}$ & $85.8_{1.4}$ & $27.0_{9.3}$ & $72.1_{2.2}$ & $84.9_{6.2}$ & $52.7_{11.9}$ & $66.7_{5.9}$ \\
    Grok & $81.5_{1.5}$ & $45.6_{3.6}$ & $54.5_{6.4}$ & $50.6_{1.4}$ & $8.91_{1.4}$ & $40.6_{3.2}$ & $81.8_{0.5}$ & $55.7_{1.1}$ & $68.1_{0.7}$ & $83.7_{1.1}$ & $12.3_{7.5}$ & $69.4_{1.1}$ & $86.9_{3.8}$ & $54.8_{9.0}$ & $71.9_{2.9}$ \\
    Claude & $77.5_{0.3}$ & $44.3_{1.4}$ & $40.8_{2.4}$ & $51.6_{1.1}$ & $11.2_{1.8}$ & $35.7_{1.8}$ & $81.2_{0.4}$ & $54.2_{1.0}$ & $68.0_{0.5}$ & $86.2_{1.0}$ & $22.5_{24.5}$ & $70.9_{4.1}$ & $84.1_{0.9}$ & $55.0_{5.8}$ & $62.0_{7.7}$ \\
    \hline
    \textit{Mean} & $79.4_{2.8}$ & $44.7_{7.1}$ & $50.7_{7.5}$ & $51.6_{1.6}$ & $9.91_{1.3}$ & $38.5_{2.7}$ & $81.8_{0.6}$ & $\mathbf{55.3}_{1.1}$ & $67.9_{0.9}$ & $\mathbf{84.4}_{2.1}$ & $17.3_{7.7}$ & $\mathbf{69.9}_{2.9}$ & $83.6_{4.7}$ & $47.0_{12.3}$ & $65.8_{5.3}$ \\
    \hline
  \end{tabular}%
  }
  \caption{AUC, TPR, and F1 on EvoBench (top) and RAID (bottom) per training family, with \textbf{Falcon-7B} as backbone (RoBERTa has none). Each detector is trained on a single MIRAGE generator family and tested on the full benchmark. Per family: mean$_{\text{std}}$ across three seeds; \textit{Mean}: mean$_{\text{std}}$ across the nine training families. Rows are grouped by whether the training family is present in the benchmark (EvoBench: GPT, Claude, Gemini, LLaMA, Qwen; RAID: GPT, LLaMA) or absent from it, and sorted within each group by SPADE's TPR, lowest first. F1 is computed at the Youden-J threshold fitted on each run's MIRAGE validation split. \textbf{Bold} indicates the best method on the cross-family mean.}
  \label{tab:per_family_falcon7b}
\end{table*}

\subsection{Backbones and Perturbations robustness}
\label{appendix:robustness}
\begin{figure*}[t]
  \centering
  \includegraphics[width=\textwidth]{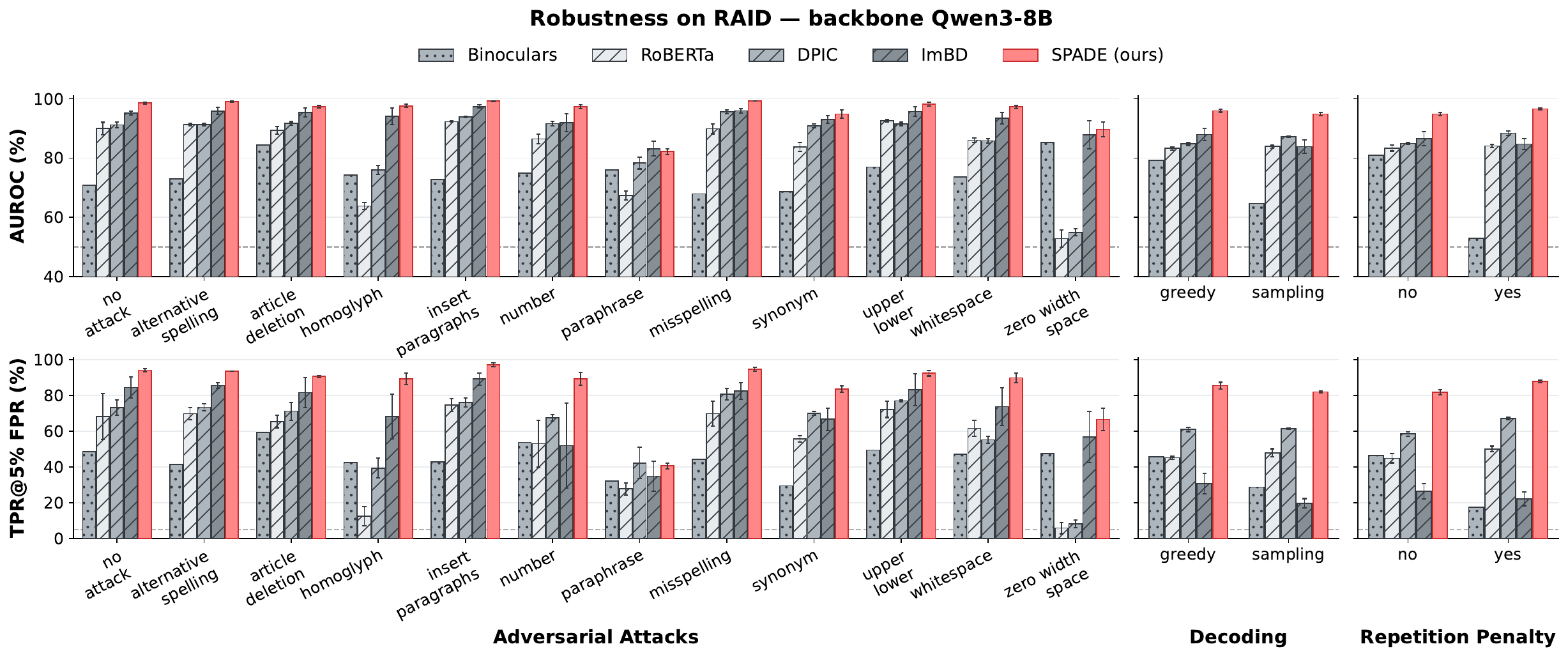}
  \caption{AUC and TPR of 5 baselines on RAID benchmark against SPADE grouped by adversarial attacks, decoding strategy, and with repetition penalty or not.}
  \label{fig:figure_attacks_qwen}
\end{figure*}

\begin{figure*}[t]
  \centering
  \includegraphics[width=\textwidth]{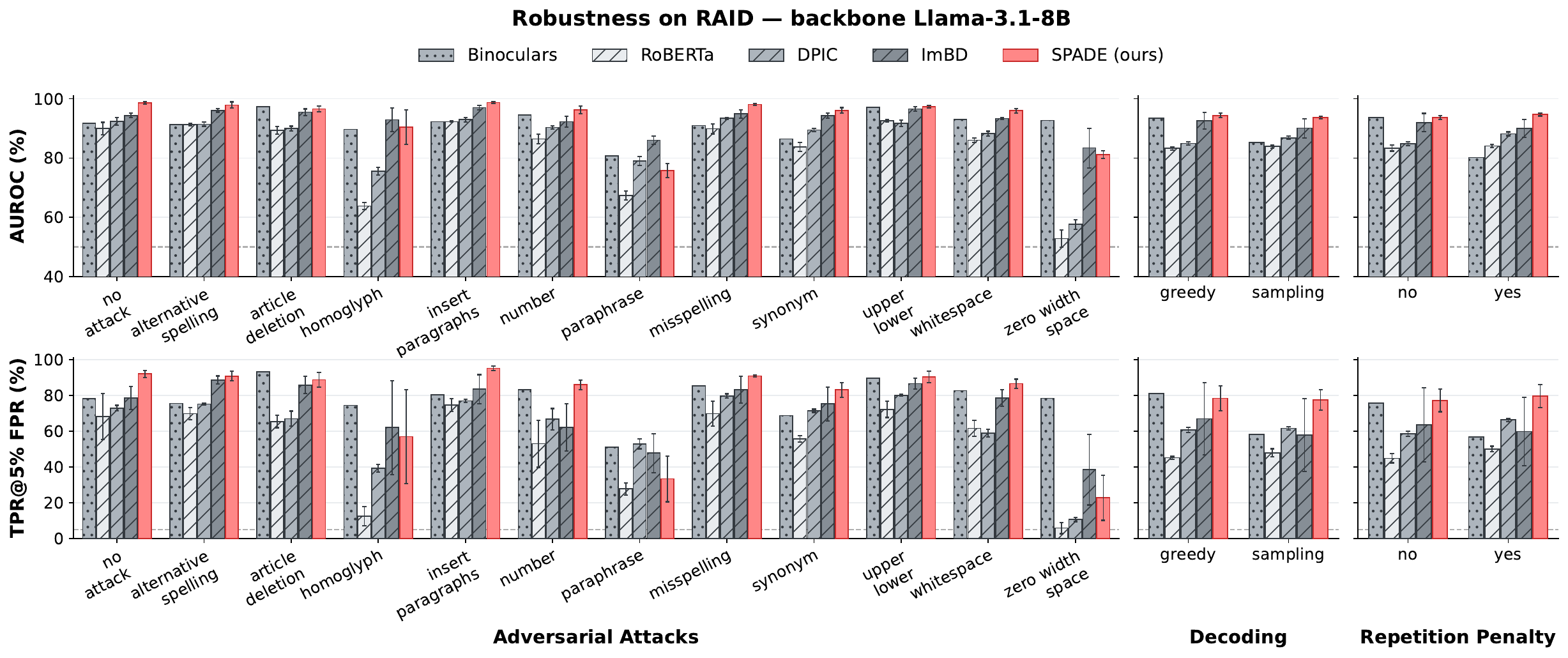}
  \caption{AUC and TPR of 5 baselines on RAID benchmark against SPADE grouped by adversarial attacks, decoding strategy, and with repetition penalty or not.}
  \label{fig:figure_attacks_llama}
\end{figure*}

\begin{figure*}[t]
  \centering
  \includegraphics[width=\textwidth]{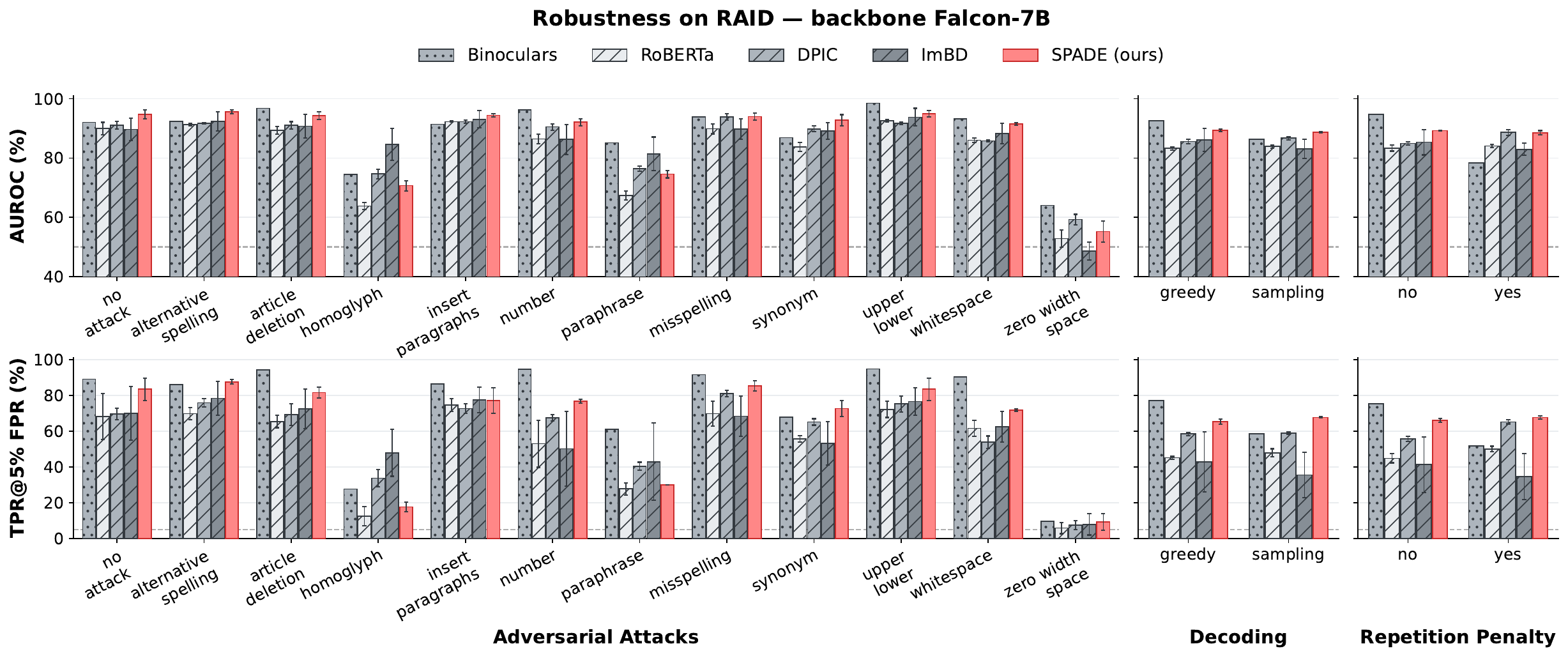}
  \caption{AUC and TPR of 5 baselines on RAID benchmark against SPADE grouped by adversarial attacks, decoding strategy, and with repetition penalty or not.}
  \label{fig:figure_attacks_falcon}
\end{figure*}

Tables \ref{tab:raid_attacks_qwen3_8b_base}, \ref{tab:raid_attacks_llama31_8b}, and \ref{tab:raid_attacks_falcon7b} report the per-attack AUC and TPR of all detectors on RAID for the three backbones, and Figures \ref{fig:figure_attacks_qwen}, \ref{fig:figure_attacks_llama}, and \ref{fig:figure_attacks_falcon} show the corresponding plots for Qwen3-8B-Base, LLaMA-3.1-8B, and Falcon-7B, including the decoding strategy and repetition penalty settings. Paraphrasing is hard for all detectors. SPADE's main weaknesses are the zero-width space attack (22.7 and 9.28 TPR with LLaMA-3.1-8B and Falcon-7B) and, with Falcon-7B, the homoglyph attack (17.7 TPR).

\begin{table*}[t]
  \centering
  \resizebox{\textwidth}{!}{%
  \begin{tabular}{l ccc ccc ccc ccc ccc ccc}
    \hline
    \textbf{Attack} $\downarrow$ \ \textbf{Method} $\rightarrow$ & \multicolumn{3}{c}{Likelihood} & \multicolumn{3}{c}{Entropy} & \multicolumn{3}{c}{DNA-GPT} & \multicolumn{3}{c}{Fast-DetectGPT} & \multicolumn{3}{c}{Binoculars} & \multicolumn{3}{c}{\textbf{SPADE}} \\
    \cmidrule(lr){2-4} \cmidrule(lr){5-7} \cmidrule(lr){8-10} \cmidrule(lr){11-13} \cmidrule(lr){14-16} \cmidrule(lr){17-19}
     & AUC & TPR & F1 & AUC & TPR & F1 & AUC & TPR & F1 & AUC & TPR & F1 & AUC & TPR & F1 & AUC & TPR & F1 \\
    \hline
    No attack & $87.6$ & $62.4$ & $70.2$ & $85.7$ & $49.5$ & $68.2$ & $79.4$ & $39.6$ & $65.7$ & $64.0$ & $30.7$ & $55.4$ & $70.8$ & $48.5$ & $62.4$ & $\mathbf{98.6}$ & $\mathbf{94.1}$ & $\mathbf{94.1}$ \\
    Alt. spelling & $84.7$ & $47.9$ & $67.6$ & $82.3$ & $44.7$ & $64.7$ & $74.7$ & $30.9$ & $62.9$ & $68.5$ & $36.2$ & $56.7$ & $73.0$ & $41.5$ & $61.4$ & $\mathbf{99.0}$ & $\mathbf{93.6}$ & $\mathbf{94.1}$ \\
    Article deletion & $87.7$ & $57.3$ & $75.6$ & $82.7$ & $37.9$ & $71.2$ & $80.9$ & $33.0$ & $68.7$ & $80.3$ & $50.5$ & $67.1$ & $84.5$ & $59.2$ & $74.7$ & $\mathbf{97.4}$ & $\mathbf{90.6}$ & $\mathbf{91.9}$ \\
    Homoglyph & $79.3$ & $42.6$ & $52.1$ & $77.6$ & $31.9$ & $52.1$ & $77.0$ & $41.5$ & $53.3$ & $63.5$ & $24.0$ & $51.5$ & $74.3$ & $42.6$ & $57.1$ & $\mathbf{97.6}$ & $\mathbf{89.3}$ & $\mathbf{89.5}$ \\
    Insert paragraphs & $86.7$ & $57.3$ & $69.6$ & $85.2$ & $50.4$ & $68.9$ & $79.9$ & $42.7$ & $68.1$ & $68.0$ & $30.8$ & $58.3$ & $72.7$ & $42.7$ & $62.7$ & $\mathbf{99.2}$ & $\mathbf{97.2}$ & $\mathbf{95.1}$ \\
    Number & $91.7$ & $71.6$ & $71.5$ & $89.1$ & $58.9$ & $68.8$ & $85.4$ & $51.6$ & $70.4$ & $71.7$ & $42.1$ & $59.2$ & $74.9$ & $53.7$ & $65.3$ & $\mathbf{97.4}$ & $\mathbf{89.3}$ & $\mathbf{90.0}$ \\
    Paraphrase & $67.9$ & $24.4$ & $52.3$ & $61.5$ & $18.9$ & $46.4$ & $63.7$ & $23.3$ & $49.4$ & $65.3$ & $15.7$ & $51.5$ & $75.9$ & $32.2$ & $54.2$ & $\mathbf{82.2}$ & $\mathbf{40.6}$ & $\mathbf{55.3}$ \\
    Misspelling & $88.5$ & $61.1$ & $67.2$ & $87.4$ & $53.7$ & $65.9$ & $79.1$ & $40.0$ & $62.2$ & $63.1$ & $32.6$ & $51.2$ & $67.9$ & $44.2$ & $58.0$ & $\mathbf{99.3}$ & $\mathbf{94.7}$ & $\mathbf{94.2}$ \\
    Synonym & $77.0$ & $34.8$ & $55.3$ & $72.0$ & $19.6$ & $55.9$ & $65.4$ & $16.1$ & $36.4$ & $69.7$ & $22.3$ & $36.1$ & $68.6$ & $29.5$ & $35.5$ & $\mathbf{94.8}$ & $\mathbf{83.6}$ & $\mathbf{86.2}$ \\
    Upper/lower & $86.9$ & $44.3$ & $73.8$ & $80.3$ & $40.2$ & $65.1$ & $81.3$ & $33.0$ & $69.7$ & $72.0$ & $37.1$ & $59.6$ & $77.0$ & $49.5$ & $67.1$ & $\mathbf{98.2}$ & $\mathbf{92.4}$ & $\mathbf{92.3}$ \\
    Whitespace & $84.9$ & $54.8$ & $72.4$ & $81.0$ & $42.3$ & $68.0$ & $76.4$ & $38.5$ & $65.6$ & $69.4$ & $37.5$ & $58.6$ & $73.6$ & $47.1$ & $64.5$ & $\mathbf{97.3}$ & $\mathbf{89.9}$ & $\mathbf{90.9}$ \\
    Zero-width space & $85.9$ & $49.5$ & $51.2$ & $78.6$ & $18.6$ & $51.2$ & $83.0$ & $23.7$ & $51.2$ & $80.3$ & $41.2$ & $\mathbf{65.1}$ & $85.2$ & $47.4$ & $64.0$ & $\mathbf{89.7}$ & $\mathbf{66.6}$ & $60.5$ \\
    \hline
    \textit{Mean} & $83.7$ & $49.6$ & $64.4$ & $79.8$ & $37.9$ & $61.7$ & $77.0$ & $34.0$ & $59.8$ & $70.2$ & $33.6$ & $55.9$ & $75.2$ & $44.5$ & $60.4$ & $\mathbf{95.6}$ & $\mathbf{84.4}$ & $\mathbf{85.5}$ \\
    \hline\hline
    \textbf{Attack} $\downarrow$ \ \textbf{Method} $\rightarrow$ & \multicolumn{3}{c}{NTS} & \multicolumn{3}{c}{RoBERTa} & \multicolumn{3}{c}{RAIDAR} & \multicolumn{3}{c}{DPIC} & \multicolumn{3}{c}{ImBD} & \multicolumn{3}{c}{\textbf{SPADE}} \\
    \cmidrule(lr){2-4} \cmidrule(lr){5-7} \cmidrule(lr){8-10} \cmidrule(lr){11-13} \cmidrule(lr){14-16} \cmidrule(lr){17-19}
     & AUC & TPR & F1 & AUC & TPR & F1 & AUC & TPR & F1 & AUC & TPR & F1 & AUC & TPR & F1 & AUC & TPR & F1 \\
    \hline
    No attack & $74.3$ & $36.6$ & $62.6$ & $90.0$ & $68.3$ & $74.9$ & $69.4$ & $24.4$ & $55.6$ & $91.2$ & $73.3$ & $82.7$ & $95.2$ & $84.5$ & $87.9$ & $\mathbf{98.6}$ & $\mathbf{94.1}$ & $\mathbf{94.1}$ \\
    Alt. spelling & $72.8$ & $28.7$ & $59.4$ & $91.3$ & $69.9$ & $76.5$ & $79.0$ & $28.0$ & $59.2$ & $91.3$ & $73.4$ & $83.3$ & $95.9$ & $85.5$ & $89.7$ & $\mathbf{99.0}$ & $\mathbf{93.6}$ & $\mathbf{94.1}$ \\
    Article deletion & $69.1$ & $22.3$ & $56.9$ & $89.4$ & $65.4$ & $70.7$ & $74.0$ & $29.8$ & $49.8$ & $91.7$ & $71.2$ & $80.6$ & $95.5$ & $81.6$ & $88.0$ & $\mathbf{97.4}$ & $\mathbf{90.6}$ & $\mathbf{91.9}$ \\
    Homoglyph & $73.4$ & $43.6$ & $53.6$ & $63.8$ & $12.5$ & $44.5$ & $40.7$ & $6.38$ & $4.61$ & $75.9$ & $39.4$ & $56.2$ & $94.1$ & $68.1$ & $59.4$ & $\mathbf{97.6}$ & $\mathbf{89.3}$ & $\mathbf{89.5}$ \\
    Insert paragraphs & $73.3$ & $23.1$ & $63.6$ & $92.3$ & $74.6$ & $72.6$ & $79.3$ & $29.3$ & $66.1$ & $93.9$ & $76.1$ & $82.0$ & $97.4$ & $89.2$ & $90.3$ & $\mathbf{99.2}$ & $\mathbf{97.2}$ & $\mathbf{95.1}$ \\
    Number & $76.0$ & $38.9$ & $63.9$ & $86.5$ & $53.0$ & $68.7$ & $68.1$ & $20.0$ & $48.9$ & $91.6$ & $67.4$ & $80.9$ & $91.9$ & $51.9$ & $80.5$ & $\mathbf{97.4}$ & $\mathbf{89.3}$ & $\mathbf{90.0}$ \\
    Paraphrase & $67.4$ & $22.2$ & $53.8$ & $67.4$ & $27.8$ & $43.6$ & $51.0$ & $5.56$ & $21.1$ & $78.3$ & $\mathbf{42.2}$ & $\mathbf{65.3}$ & $\mathbf{83.2}$ & $34.8$ & $64.5$ & $82.2$ & $40.6$ & $55.3$ \\
    Misspelling & $75.9$ & $38.9$ & $60.5$ & $89.9$ & $69.8$ & $76.7$ & $82.0$ & $42.8$ & $65.1$ & $95.6$ & $80.7$ & $85.1$ & $95.9$ & $82.5$ & $85.1$ & $\mathbf{99.3}$ & $\mathbf{94.7}$ & $\mathbf{94.2}$ \\
    Synonym & $37.6$ & $8.04$ & $36.8$ & $83.8$ & $55.7$ & $68.5$ & $76.1$ & $25.6$ & $63.1$ & $90.9$ & $69.9$ & $82.5$ & $93.0$ & $66.7$ & $78.5$ & $\mathbf{94.8}$ & $\mathbf{83.6}$ & $\mathbf{86.2}$ \\
    Upper/lower & $66.6$ & $22.7$ & $52.9$ & $92.7$ & $72.2$ & $72.3$ & $85.4$ & $45.4$ & $71.7$ & $91.5$ & $77.0$ & $83.3$ & $95.7$ & $83.2$ & $89.5$ & $\mathbf{98.2}$ & $\mathbf{92.4}$ & $\mathbf{92.3}$ \\
    Whitespace & $67.9$ & $19.2$ & $53.6$ & $86.0$ & $61.5$ & $57.5$ & $79.4$ & $27.2$ & $65.8$ & $85.8$ & $55.1$ & $71.4$ & $93.4$ & $73.7$ & $84.1$ & $\mathbf{97.3}$ & $\mathbf{89.9}$ & $\mathbf{90.9}$ \\
    Zero-width space & $84.5$ & $53.6$ & $\mathbf{66.9}$ & $52.8$ & $5.84$ & $0.68$ & $39.8$ & $4.81$ & $4.60$ & $54.9$ & $8.25$ & $16.5$ & $87.8$ & $56.7$ & $51.2$ & $\mathbf{89.7}$ & $\mathbf{66.6}$ & $60.5$ \\
    \hline
    \textit{Mean} & $69.5$ & $29.2$ & $56.5$ & $81.4$ & $51.7$ & $59.3$ & $68.6$ & $24.1$ & $47.3$ & $85.6$ & $60.1$ & $71.6$ & $93.1$ & $70.3$ & $78.2$ & $\mathbf{95.6}$ & $\mathbf{84.4}$ & $\mathbf{85.5}$ \\
    \hline
  \end{tabular}%
  }
  \caption{AUC, TPR, and F1 on RAID per adversarial attack, with \textbf{Qwen3-8B-Base} as backbone. Each attack compares the instruct generators' texts with the human texts carrying the same attack, pooled over the eight domains. \textit{Mean} averages the eleven attacks, excluding \textit{No attack}. F1 is computed at the Youden-J threshold fitted on each run's MIRAGE validation split. When a criterion ranks human texts above AI texts on the validation split, no threshold exists and its F1 is 0. \textbf{Bold} indicates the best method per row and metric within each panel.}
  \label{tab:raid_attacks_qwen3_8b_base}
\end{table*}

\begin{table*}[t]
  \centering
  \resizebox{\textwidth}{!}{%
  \begin{tabular}{l ccc ccc ccc ccc ccc ccc}
    \hline
    \textbf{Attack} $\downarrow$ \ \textbf{Method} $\rightarrow$ & \multicolumn{3}{c}{Likelihood} & \multicolumn{3}{c}{Entropy} & \multicolumn{3}{c}{DNA-GPT} & \multicolumn{3}{c}{Fast-DetectGPT} & \multicolumn{3}{c}{Binoculars} & \multicolumn{3}{c}{\textbf{SPADE}} \\
    \cmidrule(lr){2-4} \cmidrule(lr){5-7} \cmidrule(lr){8-10} \cmidrule(lr){11-13} \cmidrule(lr){14-16} \cmidrule(lr){17-19}
     & AUC & TPR & F1 & AUC & TPR & F1 & AUC & TPR & F1 & AUC & TPR & F1 & AUC & TPR & F1 & AUC & TPR & F1 \\
    \hline
    No attack & $82.9$ & $43.6$ & $64.9$ & $78.7$ & $29.7$ & $0.00$ & $73.9$ & $11.9$ & $62.2$ & $88.9$ & $71.3$ & $71.2$ & $91.8$ & $78.2$ & $77.8$ & $\mathbf{98.6}$ & $\mathbf{92.1}$ & $\mathbf{92.3}$ \\
    Alt. spelling & $80.1$ & $29.8$ & $61.8$ & $75.8$ & $20.2$ & $0.00$ & $72.0$ & $20.2$ & $59.8$ & $90.4$ & $71.5$ & $76.0$ & $91.3$ & $75.5$ & $80.5$ & $\mathbf{97.9}$ & $\mathbf{90.8}$ & $\mathbf{91.9}$ \\
    Article deletion & $84.0$ & $41.7$ & $68.7$ & $74.8$ & $26.2$ & $0.00$ & $80.1$ & $28.2$ & $66.5$ & $96.3$ & $91.3$ & $85.0$ & $\mathbf{97.3}$ & $\mathbf{93.2}$ & $\mathbf{91.5}$ & $96.6$ & $88.7$ & $86.9$ \\
    Homoglyph & $78.4$ & $51.1$ & $52.2$ & $75.3$ & $36.2$ & $0.00$ & $79.5$ & $37.2$ & $53.3$ & $83.3$ & $55.3$ & $55.4$ & $89.7$ & $\mathbf{74.5}$ & $64.2$ & $\mathbf{90.4}$ & $57.0$ & $\mathbf{72.5}$ \\
    Insert paragraphs & $83.3$ & $53.0$ & $66.0$ & $80.4$ & $38.5$ & $0.00$ & $76.2$ & $53.8$ & $60.1$ & $90.4$ & $76.9$ & $75.8$ & $92.3$ & $80.3$ & $83.3$ & $\mathbf{98.7}$ & $\mathbf{95.2}$ & $\mathbf{94.4}$ \\
    Number & $88.5$ & $61.1$ & $63.9$ & $84.1$ & $47.4$ & $0.00$ & $80.7$ & $17.9$ & $62.9$ & $92.4$ & $73.3$ & $76.1$ & $94.6$ & $83.2$ & $84.8$ & $\mathbf{96.2}$ & $\mathbf{86.0}$ & $\mathbf{86.2}$ \\
    Paraphrase & $65.2$ & $25.6$ & $49.5$ & $57.6$ & $12.2$ & $0.00$ & $58.1$ & $12.2$ & $47.4$ & $73.4$ & $25.8$ & $51.5$ & $\mathbf{80.8}$ & $\mathbf{51.1}$ & $54.1$ & $75.8$ & $33.3$ & $\mathbf{54.7}$ \\
    Misspelling & $85.8$ & $48.4$ & $61.2$ & $82.0$ & $41.1$ & $0.00$ & $77.3$ & $24.2$ & $58.4$ & $90.1$ & $81.2$ & $71.8$ & $91.0$ & $85.3$ & $79.1$ & $\mathbf{98.1}$ & $\mathbf{90.9}$ & $\mathbf{91.7}$ \\
    Synonym & $74.2$ & $22.3$ & $58.3$ & $65.7$ & $14.3$ & $0.00$ & $68.6$ & $20.5$ & $47.7$ & $86.5$ & $61.6$ & $70.4$ & $86.4$ & $68.8$ & $71.7$ & $\mathbf{96.1}$ & $\mathbf{83.1}$ & $\mathbf{88.0}$ \\
    Upper/lower & $83.8$ & $33.0$ & $66.8$ & $74.6$ & $22.7$ & $0.00$ & $79.5$ & $25.8$ & $65.9$ & $96.3$ & $86.6$ & $85.5$ & $97.1$ & $89.7$ & $89.9$ & $\mathbf{97.3}$ & $\mathbf{90.4}$ & $\mathbf{90.2}$ \\
    Whitespace & $80.4$ & $37.5$ & $66.7$ & $76.2$ & $30.8$ & $0.00$ & $73.2$ & $22.1$ & $61.3$ & $92.0$ & $78.8$ & $80.7$ & $93.0$ & $82.7$ & $85.7$ & $\mathbf{96.0}$ & $\mathbf{86.5}$ & $\mathbf{89.0}$ \\
    Zero-width space & $82.6$ & $41.2$ & $51.2$ & $75.0$ & $13.4$ & $0.00$ & $81.5$ & $39.2$ & $51.2$ & $\mathbf{92.9}$ & $\mathbf{82.7}$ & $55.0$ & $92.7$ & $78.4$ & $\mathbf{57.3}$ & $81.1$ & $22.7$ & $55.5$ \\
    \hline
    \textit{Mean} & $80.6$ & $40.4$ & $60.6$ & $74.7$ & $27.5$ & $0.00$ & $75.1$ & $27.4$ & $57.7$ & $89.5$ & $71.4$ & $71.2$ & $91.5$ & $\mathbf{78.4}$ & $76.6$ & $\mathbf{93.1}$ & $75.0$ & $\mathbf{81.9}$ \\
    \hline\hline
    \textbf{Attack} $\downarrow$ \ \textbf{Method} $\rightarrow$ & \multicolumn{3}{c}{NTS} & \multicolumn{3}{c}{RoBERTa} & \multicolumn{3}{c}{RAIDAR} & \multicolumn{3}{c}{DPIC} & \multicolumn{3}{c}{ImBD} & \multicolumn{3}{c}{\textbf{SPADE}} \\
    \cmidrule(lr){2-4} \cmidrule(lr){5-7} \cmidrule(lr){8-10} \cmidrule(lr){11-13} \cmidrule(lr){14-16} \cmidrule(lr){17-19}
     & AUC & TPR & F1 & AUC & TPR & F1 & AUC & TPR & F1 & AUC & TPR & F1 & AUC & TPR & F1 & AUC & TPR & F1 \\
    \hline
    No attack & $80.8$ & $55.4$ & $69.8$ & $90.0$ & $68.3$ & $74.9$ & $79.1$ & $35.3$ & $68.9$ & $92.3$ & $72.9$ & $83.0$ & $94.4$ & $78.5$ & $87.9$ & $\mathbf{98.6}$ & $\mathbf{92.1}$ & $\mathbf{92.3}$ \\
    Alt. spelling & $82.7$ & $46.8$ & $71.1$ & $91.3$ & $69.9$ & $76.5$ & $80.8$ & $31.2$ & $67.7$ & $91.4$ & $75.2$ & $81.5$ & $96.1$ & $88.7$ & $89.3$ & $\mathbf{97.9}$ & $\mathbf{90.8}$ & $\mathbf{91.9}$ \\
    Article deletion & $86.4$ & $59.2$ & $77.7$ & $89.4$ & $65.4$ & $70.7$ & $82.6$ & $51.1$ & $71.3$ & $89.9$ & $67.0$ & $79.9$ & $95.5$ & $85.8$ & $86.5$ & $\mathbf{96.6}$ & $\mathbf{88.7}$ & $\mathbf{86.9}$ \\
    Homoglyph & $72.0$ & $43.6$ & $59.1$ & $63.8$ & $12.5$ & $44.5$ & $50.9$ & $3.19$ & $16.9$ & $75.5$ & $39.4$ & $54.1$ & $\mathbf{92.9}$ & $\mathbf{62.1}$ & $68.6$ & $90.4$ & $57.0$ & $\mathbf{72.5}$ \\
    Insert paragraphs & $80.7$ & $41.9$ & $70.1$ & $92.3$ & $74.6$ & $72.6$ & $83.3$ & $45.0$ & $72.4$ & $93.0$ & $76.9$ & $81.4$ & $97.0$ & $83.5$ & $87.3$ & $\mathbf{98.7}$ & $\mathbf{95.2}$ & $\mathbf{94.4}$ \\
    Number & $80.4$ & $35.8$ & $67.2$ & $86.5$ & $53.0$ & $68.7$ & $81.1$ & $44.2$ & $66.4$ & $90.3$ & $66.7$ & $77.9$ & $92.3$ & $62.1$ & $79.8$ & $\mathbf{96.2}$ & $\mathbf{86.0}$ & $\mathbf{86.2}$ \\
    Paraphrase & $70.8$ & $36.7$ & $50.6$ & $67.4$ & $27.8$ & $43.6$ & $65.8$ & $11.1$ & $23.8$ & $79.0$ & $\mathbf{53.0}$ & $64.2$ & $\mathbf{86.0}$ & $47.8$ & $\mathbf{67.3}$ & $75.8$ & $33.3$ & $54.7$ \\
    Misspelling & $82.4$ & $52.6$ & $66.8$ & $89.9$ & $69.8$ & $76.7$ & $86.0$ & $55.8$ & $75.6$ & $93.4$ & $79.6$ & $82.1$ & $94.9$ & $83.2$ & $86.5$ & $\mathbf{98.1}$ & $\mathbf{90.9}$ & $\mathbf{91.7}$ \\
    Synonym & $40.8$ & $7.14$ & $37.7$ & $83.8$ & $55.7$ & $68.5$ & $76.5$ & $17.9$ & $59.4$ & $89.5$ & $71.4$ & $80.2$ & $94.3$ & $75.3$ & $85.0$ & $\mathbf{96.1}$ & $\mathbf{83.1}$ & $\mathbf{88.0}$ \\
    Upper/lower & $80.2$ & $57.7$ & $69.3$ & $92.7$ & $72.2$ & $72.3$ & $84.4$ & $40.9$ & $71.9$ & $91.7$ & $80.1$ & $84.2$ & $96.6$ & $86.6$ & $89.8$ & $\mathbf{97.3}$ & $\mathbf{90.4}$ & $\mathbf{90.2}$ \\
    Whitespace & $78.0$ & $28.8$ & $65.7$ & $86.0$ & $61.5$ & $57.5$ & $81.6$ & $32.1$ & $71.8$ & $88.3$ & $59.0$ & $73.6$ & $93.3$ & $78.5$ & $84.8$ & $\mathbf{96.0}$ & $\mathbf{86.5}$ & $\mathbf{89.0}$ \\
    Zero-width space & $\mathbf{87.4}$ & $\mathbf{61.9}$ & $\mathbf{63.3}$ & $52.8$ & $5.84$ & $0.68$ & $40.9$ & $5.15$ & $17.8$ & $57.6$ & $10.7$ & $17.0$ & $83.4$ & $38.5$ & $59.3$ & $81.1$ & $22.7$ & $55.5$ \\
    \hline
    \textit{Mean} & $76.5$ & $42.9$ & $63.5$ & $81.4$ & $51.7$ & $59.3$ & $74.0$ & $30.7$ & $55.9$ & $85.4$ & $61.7$ & $70.6$ & $92.9$ & $72.0$ & $80.4$ & $\mathbf{93.1}$ & $\mathbf{75.0}$ & $\mathbf{81.9}$ \\
    \hline
  \end{tabular}%
  }
  \caption{AUC, TPR, and F1 on RAID per adversarial attack, with \textbf{LLaMA-3.1-8B} as backbone. Each attack compares the instruct generators' texts with the human texts carrying the same attack, pooled over the eight domains. \textit{Mean} averages the eleven attacks, excluding \textit{No attack}. F1 is computed at the Youden-J threshold fitted on each run's MIRAGE validation split. When a criterion ranks human texts above AI texts on the validation split, no threshold exists and its F1 is 0. \textbf{Bold} indicates the best method per row and metric within each panel.}
  \label{tab:raid_attacks_llama31_8b}
\end{table*}

\begin{table*}[t]
  \centering
  \resizebox{\textwidth}{!}{%
  \begin{tabular}{l ccc ccc ccc ccc ccc ccc}
    \hline
    \textbf{Attack} $\downarrow$ \ \textbf{Method} $\rightarrow$ & \multicolumn{3}{c}{Likelihood} & \multicolumn{3}{c}{Entropy} & \multicolumn{3}{c}{DNA-GPT} & \multicolumn{3}{c}{Fast-DetectGPT} & \multicolumn{3}{c}{Binoculars} & \multicolumn{3}{c}{\textbf{SPADE}} \\
    \cmidrule(lr){2-4} \cmidrule(lr){5-7} \cmidrule(lr){8-10} \cmidrule(lr){11-13} \cmidrule(lr){14-16} \cmidrule(lr){17-19}
     & AUC & TPR & F1 & AUC & TPR & F1 & AUC & TPR & F1 & AUC & TPR & F1 & AUC & TPR & F1 & AUC & TPR & F1 \\
    \hline
    No attack & $81.4$ & $29.7$ & $0.00$ & $69.3$ & $16.8$ & $0.00$ & $83.2$ & $53.5$ & $65.2$ & $91.8$ & $88.1$ & $77.2$ & $92.0$ & $\mathbf{89.1}$ & $83.4$ & $\mathbf{94.8}$ & $83.5$ & $\mathbf{87.1}$ \\
    Alt. spelling & $79.4$ & $21.3$ & $0.00$ & $63.0$ & $12.8$ & $0.00$ & $81.5$ & $48.9$ & $64.4$ & $91.8$ & $86.2$ & $79.9$ & $92.4$ & $86.2$ & $84.7$ & $\mathbf{95.6}$ & $\mathbf{87.6}$ & $\mathbf{89.4}$ \\
    Article deletion & $84.4$ & $50.5$ & $0.00$ & $63.9$ & $7.77$ & $0.00$ & $89.7$ & $62.1$ & $70.8$ & $96.5$ & $92.2$ & $90.0$ & $\mathbf{96.8}$ & $\mathbf{94.2}$ & $\mathbf{95.0}$ & $94.2$ & $81.6$ & $86.8$ \\
    Homoglyph & $\mathbf{77.9}$ & $39.4$ & $0.00$ & $67.0$ & $12.8$ & $0.00$ & $77.2$ & $\mathbf{42.6}$ & $53.5$ & $67.8$ & $23.4$ & $52.8$ & $74.4$ & $27.7$ & $\mathbf{59.8}$ & $70.7$ & $17.7$ & $28.3$ \\
    Insert paragraphs & $82.8$ & $47.0$ & $0.00$ & $70.6$ & $15.4$ & $0.00$ & $81.6$ & $49.6$ & $63.8$ & $91.0$ & $\mathbf{86.3}$ & $78.7$ & $91.4$ & $\mathbf{86.3}$ & $83.4$ & $\mathbf{94.5}$ & $77.2$ & $\mathbf{88.3}$ \\
    Number & $90.5$ & $64.2$ & $0.00$ & $76.9$ & $17.9$ & $0.00$ & $89.6$ & $64.2$ & $66.5$ & $94.8$ & $85.3$ & $81.0$ & $\mathbf{96.2}$ & $\mathbf{94.7}$ & $\mathbf{87.0}$ & $92.1$ & $76.8$ & $82.5$ \\
    Paraphrase & $67.2$ & $25.6$ & $0.00$ & $55.0$ & $12.2$ & $0.00$ & $66.8$ & $24.4$ & $51.7$ & $75.7$ & $40.3$ & $53.4$ & $\mathbf{85.1}$ & $\mathbf{61.1}$ & $\mathbf{55.7}$ & $74.6$ & $30.0$ & $54.8$ \\
    Misspelling & $84.5$ & $38.9$ & $0.00$ & $71.0$ & $20.0$ & $0.00$ & $86.3$ & $65.3$ & $62.6$ & $93.1$ & $89.5$ & $78.9$ & $93.8$ & $\mathbf{91.6}$ & $\mathbf{87.4}$ & $\mathbf{94.0}$ & $85.3$ & $87.1$ \\
    Synonym & $76.4$ & $30.4$ & $0.00$ & $58.2$ & $9.82$ & $0.00$ & $80.6$ & $43.8$ & $66.1$ & $86.7$ & $64.3$ & $71.8$ & $86.8$ & $67.9$ & $76.3$ & $\mathbf{92.8}$ & $\mathbf{72.6}$ & $\mathbf{83.1}$ \\
    Upper/lower & $84.2$ & $39.2$ & $0.00$ & $64.9$ & $13.4$ & $0.00$ & $89.4$ & $59.8$ & $73.2$ & $97.8$ & $90.7$ & $90.2$ & $\mathbf{98.5}$ & $\mathbf{94.8}$ & $\mathbf{94.3}$ & $94.9$ & $83.5$ & $87.3$ \\
    Whitespace & $78.9$ & $41.3$ & $0.00$ & $64.0$ & $10.6$ & $0.00$ & $83.1$ & $62.8$ & $68.7$ & $92.8$ & $87.5$ & $87.6$ & $\mathbf{93.2}$ & $\mathbf{90.4}$ & $\mathbf{91.2}$ & $91.5$ & $71.8$ & $79.8$ \\
    Zero-width space & $62.6$ & $\mathbf{13.4}$ & $0.00$ & $52.6$ & $7.22$ & $0.00$ & $57.8$ & $8.25$ & $\mathbf{51.2}$ & $59.8$ & $9.28$ & $\mathbf{51.2}$ & $\mathbf{64.0}$ & $9.54$ & $\mathbf{51.2}$ & $55.2$ & $9.28$ & $22.1$ \\
    \hline
    \textit{Mean} & $79.0$ & $37.4$ & $0.00$ & $64.3$ & $12.7$ & $0.00$ & $80.3$ & $48.3$ & $63.0$ & $86.2$ & $68.6$ & $74.1$ & $\mathbf{88.4}$ & $\mathbf{73.1}$ & $\mathbf{78.7}$ & $86.4$ & $63.0$ & $71.8$ \\
    \hline\hline
    \textbf{Attack} $\downarrow$ \ \textbf{Method} $\rightarrow$ & \multicolumn{3}{c}{NTS} & \multicolumn{3}{c}{RoBERTa} & \multicolumn{3}{c}{RAIDAR} & \multicolumn{3}{c}{DPIC} & \multicolumn{3}{c}{ImBD} & \multicolumn{3}{c}{\textbf{SPADE}} \\
    \cmidrule(lr){2-4} \cmidrule(lr){5-7} \cmidrule(lr){8-10} \cmidrule(lr){11-13} \cmidrule(lr){14-16} \cmidrule(lr){17-19}
     & AUC & TPR & F1 & AUC & TPR & F1 & AUC & TPR & F1 & AUC & TPR & F1 & AUC & TPR & F1 & AUC & TPR & F1 \\
    \hline
    No attack & $93.2$ & $\mathbf{86.1}$ & $80.9$ & $90.0$ & $68.3$ & $74.9$ & $53.9$ & $15.5$ & $38.8$ & $91.1$ & $69.6$ & $82.1$ & $89.6$ & $70.0$ & $80.1$ & $\mathbf{94.8}$ & $83.5$ & $\mathbf{87.1}$ \\
    Alt. spelling & $94.1$ & $80.9$ & $85.0$ & $91.3$ & $69.9$ & $76.5$ & $57.0$ & $16.3$ & $40.6$ & $91.7$ & $75.9$ & $82.2$ & $92.4$ & $78.4$ & $81.0$ & $\mathbf{95.6}$ & $\mathbf{87.6}$ & $\mathbf{89.4}$ \\
    Article deletion & $\mathbf{96.8}$ & $\mathbf{91.3}$ & $\mathbf{89.2}$ & $89.4$ & $65.4$ & $70.7$ & $54.1$ & $9.06$ & $41.6$ & $91.0$ & $69.3$ & $80.5$ & $90.7$ & $72.5$ & $79.1$ & $94.2$ & $81.6$ & $86.8$ \\
    Homoglyph & $\mathbf{87.5}$ & $\mathbf{66.0}$ & $\mathbf{66.5}$ & $63.8$ & $12.5$ & $44.5$ & $45.0$ & $9.57$ & $24.4$ & $74.6$ & $33.7$ & $50.4$ & $84.6$ & $47.9$ & $44.1$ & $70.7$ & $17.7$ & $28.3$ \\
    Insert paragraphs & $94.3$ & $\mathbf{81.2}$ & $83.7$ & $92.3$ & $74.6$ & $72.6$ & $64.6$ & $21.7$ & $50.9$ & $92.2$ & $72.6$ & $79.9$ & $93.1$ & $77.5$ & $79.4$ & $\mathbf{94.5}$ & $77.2$ & $\mathbf{88.3}$ \\
    Number & $\mathbf{94.7}$ & $71.6$ & $\mathbf{82.6}$ & $86.5$ & $53.0$ & $68.7$ & $50.8$ & $5.96$ & $31.2$ & $90.5$ & $67.4$ & $77.3$ & $86.3$ & $50.2$ & $72.5$ & $92.1$ & $\mathbf{76.8}$ & $82.5$ \\
    Paraphrase & $81.3$ & $\mathbf{52.2}$ & $56.6$ & $67.4$ & $27.8$ & $43.6$ & $57.8$ & $7.41$ & $29.8$ & $76.4$ & $40.4$ & $63.0$ & $\mathbf{81.4}$ & $43.0$ & $\mathbf{65.6}$ & $74.6$ & $30.0$ & $54.8$ \\
    Misspelling & $\mathbf{94.1}$ & $\mathbf{87.4}$ & $84.3$ & $89.9$ & $69.8$ & $76.7$ & $60.8$ & $9.82$ & $40.4$ & $93.9$ & $81.1$ & $84.5$ & $89.7$ & $68.4$ & $75.7$ & $94.0$ & $85.3$ & $\mathbf{87.1}$ \\
    Synonym & $60.3$ & $22.3$ & $52.1$ & $83.8$ & $55.7$ & $68.5$ & $53.7$ & $9.82$ & $37.9$ & $89.8$ & $65.2$ & $78.6$ & $89.1$ & $53.3$ & $78.2$ & $\mathbf{92.8}$ & $\mathbf{72.6}$ & $\mathbf{83.1}$ \\
    Upper/lower & $\mathbf{96.2}$ & $\mathbf{89.7}$ & $\mathbf{90.2}$ & $92.7$ & $72.2$ & $72.3$ & $58.6$ & $12.7$ & $41.9$ & $91.7$ & $75.3$ & $82.7$ & $93.8$ & $76.6$ & $84.4$ & $94.9$ & $83.5$ & $87.3$ \\
    Whitespace & $\mathbf{95.5}$ & $\mathbf{79.8}$ & $\mathbf{85.3}$ & $86.0$ & $61.5$ & $57.5$ & $59.0$ & $17.6$ & $44.9$ & $85.8$ & $53.8$ & $71.4$ & $88.2$ & $62.5$ & $74.8$ & $91.5$ & $71.8$ & $79.8$ \\
    Zero-width space & $\mathbf{71.2}$ & $\mathbf{25.8}$ & $\mathbf{51.6}$ & $52.8$ & $5.84$ & $0.68$ & $46.4$ & $2.06$ & $40.9$ & $59.2$ & $7.56$ & $22.5$ & $48.5$ & $7.90$ & $51.2$ & $55.2$ & $9.28$ & $22.1$ \\
    \hline
    \textit{Mean} & $\mathbf{87.8}$ & $\mathbf{68.0}$ & $\mathbf{75.2}$ & $81.4$ & $51.7$ & $59.3$ & $55.3$ & $11.1$ & $38.6$ & $85.2$ & $58.4$ & $70.3$ & $85.3$ & $58.0$ & $71.5$ & $86.4$ & $63.0$ & $71.8$ \\
    \hline
  \end{tabular}%
  }
  \caption{AUC, TPR, and F1 on RAID per adversarial attack, with \textbf{Falcon-7B} as backbone. Each attack compares the instruct generators' texts with the human texts carrying the same attack, pooled over the eight domains. \textit{Mean} averages the eleven attacks, excluding \textit{No attack}. F1 is computed at the Youden-J threshold fitted on each run's MIRAGE validation split. When a criterion ranks human texts above AI texts on the validation split, no threshold exists and its F1 is 0. \textbf{Bold} indicates the best method per row and metric within each panel.}
  \label{tab:raid_attacks_falcon7b}
\end{table*}

\section{Cross-Lingual}
\label{appendix:cross-lingual}
We present in table \ref{tab:detectrlx_f1b_f1f} a more detailed table for the Cross-Lingual task. We additionally report F1 at the fixed threshold 0, the natural choice for a log-likelihood ratio test. Notably, the threshold at $0$ yields much higher F1 across all languages, all above 93\%.

\begin{table*}[t]
  \centering
  \resizebox{\textwidth}{!}{%
  \begin{tabular}{l ccc ccc ccc ccc}
    \hline
    \textbf{Train} $\downarrow$ \ \textbf{Test} $\rightarrow$
      & \multicolumn{3}{c}{English} & \multicolumn{3}{c}{Chinese}
      & \multicolumn{3}{c}{Spanish} & \multicolumn{3}{c}{Arabic} \\
    \cmidrule(lr){2-4} \cmidrule(lr){5-7} \cmidrule(lr){8-10} \cmidrule(lr){11-13}
      & F1B & F1F & F1@0 & F1B & F1F & F1@0 & F1B & F1F & F1@0 & F1B & F1F & F1@0 \\
    \hline
    English    & $\mathbf{99.90}$ & $\mathbf{99.90}$ & $\mathbf{99.90}$ & $95.52$ & $96.67$ & $98.87$ & $83.52$ & $92.37$ & $99.68$ & $71.53$ & $78.64$ & $95.99$ \\
    Chinese    & $98.76$ & $98.76$ & $99.60$ & $\mathbf{99.96}$ & $\mathbf{99.96}$ & $\mathbf{99.97}$ & $97.17$ & $97.17$ & $97.94$ & $93.59$ & $93.59$ & $97.62$ \\
    Spanish    & $98.75$ & $98.75$ & $99.62$ & $90.98$ & $90.98$ & $98.58$ & $\mathbf{99.97}$ & $\mathbf{99.97}$ & $\mathbf{100.00}$ & $93.76$ & $93.76$ & $96.06$ \\
    Arabic     & $99.26$ & $89.31$ & $99.32$ & $99.82$ & $98.36$ & $99.82$ & $98.44$ & $97.64$ & $98.49$ & $\mathbf{99.91}$ & $\mathbf{99.77}$ & $\mathbf{99.90}$ \\
    French     & $99.60$ & $99.60$ & $99.61$ & $99.91$ & $99.91$ & $99.92$ & $98.07$ & $98.07$ & $98.07$ & $97.89$ & $97.89$ & $98.34$ \\
    Russian    & $99.24$ & $52.59$ & $99.39$ & $99.96$ & $99.27$ & $99.96$ & $97.35$ & $96.63$ & $97.38$ & $99.77$ & $99.17$ & $99.78$ \\
    Portuguese & $99.52$ & $97.26$ & $99.55$ & $99.85$ & $96.50$ & $99.87$ & $98.85$ & $98.31$ & $98.82$ & $97.94$ & $99.14$ & $98.60$ \\
    German     & $99.51$ & $96.35$ & $99.52$ & $99.69$ & $97.96$ & $99.91$ & $98.55$ & $97.76$ & $98.52$ & $99.67$ & $98.92$ & $99.72$ \\
    \hline\hline
    \textbf{Train} $\downarrow$ \ \textbf{Test} $\rightarrow$
      & \multicolumn{3}{c}{French} & \multicolumn{3}{c}{Russian}
      & \multicolumn{3}{c}{Portuguese} & \multicolumn{3}{c}{German} \\
    \cmidrule(lr){2-4} \cmidrule(lr){5-7} \cmidrule(lr){8-10} \cmidrule(lr){11-13}
      & F1B & F1F & F1@0 & F1B & F1F & F1@0 & F1B & F1F & F1@0 & F1B & F1F & F1@0 \\
    \hline
    English    & $89.99$ & $93.67$ & $99.41$ & $80.43$ & $83.62$ & $94.77$ & $88.85$ & $93.59$ & $99.58$ & $64.45$ & $72.90$ & $93.91$ \\
    Chinese    & $94.96$ & $94.96$ & $99.06$ & $89.51$ & $89.51$ & $96.18$ & $97.92$ & $97.92$ & $99.72$ & $81.89$ & $81.89$ & $93.88$ \\
    Spanish    & $99.58$ & $99.58$ & $99.76$ & $91.89$ & $91.89$ & $96.26$ & $99.85$ & $99.85$ & $99.94$ & $98.12$ & $98.12$ & $99.04$ \\
    Arabic     & $99.45$ & $96.86$ & $99.44$ & $99.25$ & $96.31$ & $99.01$ & $99.87$ & $98.95$ & $99.88$ & $97.96$ & $97.10$ & $97.71$ \\
    French     & $\mathbf{99.96}$ & $\mathbf{99.96}$ & $\mathbf{99.97}$ & $98.28$ & $98.28$ & $98.62$ & $99.97$ & $99.97$ & $99.97$ & $97.94$ & $97.94$ & $98.74$ \\
    Russian    & $99.62$ & $99.02$ & $99.66$ & $\mathbf{99.87}$ & $\mathbf{99.73}$ & $\mathbf{99.85}$ & $99.91$ & $99.67$ & $99.92$ & $99.79$ & $98.92$ & $99.75$ \\
    Portuguese & $99.77$ & $99.83$ & $99.84$ & $95.68$ & $98.38$ & $97.01$ & $\mathbf{99.94}$ & $\mathbf{99.92}$ & $\mathbf{99.94}$ & $94.16$ & $98.12$ & $96.23$ \\
    German     & $99.78$ & $99.42$ & $99.78$ & $99.42$ & $98.98$ & $99.46$ & $99.82$ & $99.39$ & $99.82$ & $\mathbf{99.94}$ & $\mathbf{99.75}$ & $\mathbf{99.95}$ \\
    \hline
  \end{tabular}%
  }
  \caption{SPADE on DetectRL-X \citep{wu-etal-2026-detectrl} under its cross-language protocol. SPADE is trained and tuned on a single language train split, and tested on the others. F1B: macro-F1 at the Youden-J threshold fitted on the training split; F1F: macro-F1 at the threshold with FPR = 1\% on the training split; F1@0: macro-F1 at the fixed threshold $\tau = 0$.}
  \label{tab:detectrlx_f1b_f1f}
\end{table*}

\section{Training Necessity}
\label{appendix:training_time}
We report in Table~\ref{tab:training_time} the time and memory needed to train SPADE on MIRAGE-DIG. Training covers the 64,652 training texts (32,326 human/AI pairs) for 3 epochs, with a batch size of 16 and soft prompts of 64 tokens, on a single H100 NVL GPU (94 GB) with FlashAttention-2 and gradient checkpointing. We time one epoch per backbone; epoch times vary by less than 1\% across epochs and seeds, so the total is three times the epoch time. Only the soft prompts are optimized ($2 \times 64$ embeddings) and the backbone stays frozen.

\begin{table}[htbp]
    \centering
    \resizebox{\columnwidth}{!}{%
    \begin{tabular}{lccc}
    \toprule
    Backbone & per epoch & total (3 epochs) & peak VRAM \\
    \midrule
    Qwen3-8B-Base   & 2.1 h & 6.2 h & 33.6 GB \\
    LLaMA-3.1-8B    & 1.7 h & 5.2 h & 30.7 GB \\
    Falcon-7B       & 1.4 h & 4.2 h & 24.0 GB \\
    \bottomrule
    \end{tabular}%
    }
    \caption{Training time and peak GPU memory of SPADE on MIRAGE-DIG (single H100 NVL, bfloat16, FlashAttention-2, gradient checkpointing).}
    \label{tab:training_time}
\end{table}
\end{document}